\documentclass{article}
\usepackage[utf8]{inputenc}
\usepackage[T1]{fontenc}
\usepackage[protrusion=true,expansion=false]{microtype}
\usepackage{graphicx}
\usepackage{booktabs}
\usepackage{array}
\usepackage{multicol}\raggedcolumns
\usepackage{longtable}
\usepackage{hyperref}
\usepackage[preprint]{icml2026}
\usepackage{amsmath}
\usepackage{amssymb}
\usepackage{mathtools}
\usepackage{textcomp}
\usepackage{newunicodechar}
\usepackage{algorithm}
\newunicodechar{≥}{\ensuremath{\ge}}
\newunicodechar{≤}{\ensuremath{\le}}
\newunicodechar{×}{\ensuremath{\times}}
\newunicodechar{−}{\ensuremath{-}}
\newunicodechar{→}{\ensuremath{\to}}
\newunicodechar{≈}{\ensuremath{\approx}}
\newunicodechar{μ}{\ensuremath{\mu}}
\newunicodechar{Δ}{\ensuremath{\Delta}}
\newunicodechar{α}{\ensuremath{\alpha}}
\newunicodechar{θ}{\ensuremath{\theta}}
\newunicodechar{ℓ}{\ensuremath{\ell}}
\newunicodechar{·}{\ensuremath{\cdot}}
\newunicodechar{∞}{\ensuremath{\infty}}
\newunicodechar{ε}{\ensuremath{\varepsilon}}
\newunicodechar{γ}{\ensuremath{\gamma}}
\newunicodechar{⁻}{\ensuremath{^{-}}}
\newunicodechar{²}{\ensuremath{^{2}}}
\newunicodechar{³}{\ensuremath{^{3}}}
\newunicodechar{⁴}{\ensuremath{^{4}}}
\newunicodechar{¹}{\ensuremath{^{1}}}
\newunicodechar{⁶}{\ensuremath{^{6}}}
\newunicodechar{⁹}{\ensuremath{^{9}}}
\newunicodechar{⁷}{\ensuremath{^{7}}}
\newunicodechar{⁸}{\ensuremath{^{8}}}
\newunicodechar{⁰}{\ensuremath{^{0}}}
\newunicodechar{ğ}{\u{g}}

\icmltitlerunning{Regret without Optimal-Arm Feedback}
\renewcommand{\today}{November 7, 2026}
\hypersetup{pdfsubject={Preprint},hypertexnames=false,bookmarks=true}
\begin{document}
\twocolumn[
\icmltitle{Regret without Optimal-Arm Feedback under Reward-Dependent Delays}
\begin{icmlauthorlist}
\icmlauthor{Zhong Han}{aff}
\end{icmlauthorlist}
\icmlaffiliation{aff}{Department of Biomedical Engineering, College of Design and Engineering, National University of Singapore}
\icmlcorrespondingauthor{Zhong Han}{zhong.han.hz@u.nus.edu}
\vskip 0.3in
]
\printAffiliationsAndNotice{}
\begin{abstract}
We study stochastic bandits with reward-dependent delays, possibly infinite, whose reports omit the originating round. For each \(c\in[0,1)\), Balanced Historical Optimism (BHO) has a delay term controlled by the largest suboptimal-arm quantile at level \(1-c\Delta_i\), without requiring optimal-arm reports. Lower bounds give an inverse-square slack cost in the logarithmic statistical regime and linear minimax regret at \(c=1\). A family with growing \(K\) forces worst-case regret \(\Omega(G\log K)\) for every learner, where \(G=\max_i\Delta_i h_i\) is the largest gap-weighted suboptimal cutoff. Pending-triggered balance preserves BHO's weighted-cutoff bound for \(c<1\) and attains the historical-greedy exploration coefficient with deterministic rewards, a unique optimum and fixed bounded delays. Gap-weighted comparisons and experiments show when shared waiting helps and what balanced exploration costs.

\end{abstract}
\section{Introduction}
A verification service may confirm correct and incorrect predictions at different speeds and leave some outcomes unreported. Such reporting can bias averages of returned rewards. If reports retain only the model label, the learner also cannot match them to the originating decisions.

Lancewicki et al.~(2021; L21 in Table 1) ask whether regret can depend only on the largest suboptimal-arm delay quantile. Their optimistic--pessimistic successive elimination (OPSE) also pays an optimal-arm quantile. L21's Optimistic-UCB expression, Eq. (8), avoids it but sums suboptimal quantiles. The question is posed at \(1-\Delta_i\), while their upper bounds use \(1-\Delta_i/4\) and their lower bound \(1-2\Delta\). We study how the quantile level affects learnability.

Fix \(c\in[0,1)\), put \(\alpha=1-c\), and set \(D_c=\max_{i:\Delta_i>0}d_i(1-c\Delta_i)\). Balanced Historical Optimism (BHO) satisfies
\[
\begin{gathered}
R_T=O\Bigl(\sum_{i:\Delta_i>0}\frac{\log(eKT)}{\alpha^2\Delta_i}\\
+\frac{D_c\log^2(e\max\{1,\min\{D_c,T\}\})}{\alpha}\Bigr)
\end{gathered}
\]
without requiring optimal-arm reports (Theorems 1--2). At \(c=1/4\), the delay term removes OPSE's optimal-arm quantile at the cost of squared logarithms and larger constants. OPSE can be better when the optimum reports promptly (App. H). Theorem 5 gives the \(\alpha^{-2}\) cost in the logarithmic regime, and Theorem 6 gives linear minimax regret at \(c=1\).

BHO fills missing rewards with one, retains each arm's smallest upper index, and balances pulls among arms within a confidence width of the leader. Grouping choices by the leader's optimism lets the proof charge one shared interval across arms, even as arms leave and re-enter eligibility (Section 4, Figure 1).

On a multiscale family with growing \(K\), each learner incurs worst-case regret \(\Omega(G\log K)\), where \(G=\max_i\Delta_i h_i\). Theorem 7 separates this delay cost from the statistical term and horizon clipping. Pending-triggered balance (PTB) preserves BHO's weighted-cutoff bound for \(c<1\) and matches historical-greedy exploration under Theorem 4's deterministic bounded-delay conditions.

Calendar-rate blocks (RATE) pay a gap-weighted sum; NC-lean has both maximum and sum guarantees (Proposition 1, Theorem 3). Both have smaller-order regret than BHO and the three elimination learners on a reward-dependent family (Corollary 2). On a common-delay family, BHO has smaller-order regret than RATE, including after averaging over a uniform relabeling (Corollary 3). Section 6 examines the finite-horizon cost of balancing.

\section{Model and Algorithm}
\subsection{Actions, Reports, and Regret}
There are \(K\ge2\) arms and a known integer horizon \(T\ge1\). After the learner selects \(A_s\) using visible history and private randomness, a joint vector \(((Y_s(i),D_s(i)))_{i=1}^K\) is drawn independently of that history. Vectors are iid across rounds, with rewards in \([0,1]\) and delays in \(\mathbb N_0\cup\{\infty\}\). Within-round dependence across arms and between rewards and delays is unrestricted.

A finite-delay selected pull reports at the end of round \(s+D_s(A_s)\), after that round's decision. Reports reveal arm and reward, including zero and multiplicity, but omit pull time, identifier and delay. Each pull reports at most once; unselected outcomes remain hidden. The learner sees its actions, clock and arriving multisets.

Write \(\mu_i=\mathbb EY_s(i)\), \(\mu^*=\max_i\mu_i\), \(\Delta_i=\mu^*-\mu_i\), and \(\mathcal I=\{i:\Delta_i>0\}\). For pull counts \(N_i(T)\), actual pseudo-regret and expected regret are
\[
\Psi_T=\sum_{i\in\mathcal I}\Delta_iN_i(T), \qquad R_T=\mathbb E\Psi_T.
\]
Regret uses selected actions and latent means, including unreported outcomes. Bounds are clipped at \(T\); an empty \(\mathcal I\) gives zero regret, and infinite required cutoffs give only the trivial bound.

\subsection{Balanced Historical Optimism}
Time \(t\) counts completed rounds, including their reports. Fix \(\theta\in(0,1]\). BHO without a qualifier, or unit-width BHO, uses \(\theta=1\); BHO-half uses \(\theta=1/2\). Let \(m_i(t),n_i(t),G_i(t)\) be pull count, report count and reported reward sum. Put \(\delta=(T+1)^{-2}\), \(\ell=\log(4KT/\delta)\), and \(\beta_i(t)=\sqrt{\ell/(2m_i(t))}\), infinite at zero count. At positive counts, optimistic completion gives
\begin{equation}
\begin{gathered}
u_i^+(t)=\frac{G_i(t)+m_i(t)-n_i(t)}{m_i(t)},\\
\rho_i(t)=u_i^+(t)+\beta_i(t).
\end{gathered}
\tag{1}
\end{equation}
This estimator comes from Lancewicki et al.~(2021, Alg. 6). Each arm retains the smallest optimistic upper index seen so far, its historical index:
\begin{equation}
H_i(0)=1,\qquad H_i(t)=\min\{H_i(t-1),1,\rho_i(t)\}.
\tag{2}
\end{equation}
The update applies at positive count; otherwise the index stays one. Define
\begin{equation}
\begin{gathered}
M(t)=\max_jH_j(t),\\
\mathcal C_\theta(t)=\{i:H_i(t)+\theta\beta_i(t)\ge M(t)\}.
\end{gathered}
\tag{3}
\end{equation}
\begin{algorithm}[t]
\caption{Balanced Historical Optimism}
Initialize all pull/report counts and reported reward sums to zero, and every historical index to one. Repeat for \(t=0,\ldots,T-1\):
\begin{enumerate}

\def\labelenumi{\arabic{enumi}.}

\setlength{\itemsep}{8pt}\setlength{\parskip}{0pt}

\item

  Compute the widths from completed-round counts, using infinity at zero count, and form \(\mathcal C_\theta(t)\) by (3).

\item

  Choose the arm in \(\mathcal C_\theta(t)\) that lexicographically minimizes \((m_i,-H_i,i)\), pull it, and increment its pull count.

\item

  Process all reports arriving at this round's end, including reports of zero and reports from earlier actions. Each report adds one to its arm's report count and its reward to that arm's reward sum.

\item

  Recompute widths and update every positive-count historical index by (1)--(2); a zero-count index stays one.

\end{enumerate}

\end{algorithm}
The leader, an arm attaining \(M(t)\), is eligible; other arms can re-enter as the leading index falls. BHO uses \(O(K)\) scalars and \(O(K+r_t)\) work for \(r_t\) reports per round, storing no pull ages. Permanent elimination against an unresolved optimistic leader can discard the optimum: that leader's upper index supplies no lower confidence bound.

\section{Guarantees from Suboptimal-Arm Cutoffs}
For \(p\in[0,1]\), let
\begin{equation}
d_i(p)=\min\{h\in\mathbb N_0:\Pr(D_i\le h)\ge p\},
\tag{4}
\end{equation}
with infinity if no integer qualifies. The inclusive inequality matters at the critical endpoint. Put \(L=\log(eKT)\), \(S_T=L\sum_{i\in\mathcal I}\Delta_i^{-1}\), and \(\Phi_T(d)=d\log^2(e\min\{d,T\})\) for finite \(d>0\), with \(\Phi_T(0)=0\) and \(\Phi_T(\infty)=\infty\).

\textbf{Theorem 1 (BHO without optimal-arm feedback).} For \(c\in[0,1)\), set \(\alpha=1-c\) and \(D_c=\max_{i\in\mathcal I}d_i(1-c\Delta_i)\). Unit-width BHO satisfies
\begin{equation}
R_T\le\min\left\{T, \frac{18473S_T}{\alpha^2}+\frac{9264\Phi_T(D_c)}{\alpha}\right\}.
\tag{5}
\end{equation}
Write \(C_0=18500\) for a rounded common coefficient dominating both terms in (5). For a fixed \(\theta\in(0,1]\), let \(N_\theta=\lceil32/\theta\rceil\) and \(C_\theta=3(96N_\theta/\theta+16)\). Then
\begin{equation}
R_T^{(\theta)}\le\min\left\{T, \frac{18473S_T}{\alpha^2\theta^2} +\frac{C_\theta\Phi_T(D_c)}{\alpha}\right\}.
\tag{6}
\end{equation}
The same numerical bounds hold for \(\Psi_T\) with probability at least \(1-\delta\). Optimal-arm delays are unrestricted. Finite ordinary cutoffs at \(c=0\) require bounded suboptimal delays almost surely. A positive \(c\) allows permanent missingness. BHO does not receive \(c\), the gaps or the cutoffs, so its bound can be minimized over \(c\) (Apps. A, M).

\subsection{Weighted Completion and Gap Profiles}
An unreported reward contributes deficit \(1-Y\). Define
\begin{equation}
\begin{gathered}
b_i(h)=\mathbb E[(1-Y_i)\mathbf1\{D_i>h\}],\\
w_i(c)=\min\{h\in\mathbb N_0:b_i(h)\le c\Delta_i\},
\end{gathered}
\tag{7}
\end{equation}
and \(W_c=\max_{i\in\mathcal I}w_i(c)\), using infinity for a missing finite cutoff.

\textbf{Theorem 2 (weighted completion bias).} Theorem 1, including its high-probability and fixed-policy minimization statements, holds with \(D_c\) replaced by \(W_c\).

Since \(b_i(h)\le\Pr(D_i>h)\), \(W_c\le D_c\). Hidden binary successes contribute no deficit: if all suboptimal failures report by \(h\), then \(b_i(h)=0\), even if all successes and optimal-arm outcomes are hidden. If such an arm hides every success, its ordinary cutoff \(d_i(1-c\Delta_i)\) is infinite whenever \(c\Delta_i<\mu_i\), while \(w_i(c)\le h\).

Finite cutoffs are sufficient, not necessary. With completed limiting means \(\nu_i=\mu_i+\mathbb E[(1-Y_i)\mathbf1\{D_i=\infty\}]\), BHO is pointwise consistent if all maximizers are optimal (App. G); mixed optimal/suboptimal maximizers are not covered. App. G.5 compares completion with returned averaging.

\textbf{Corollary 1 (bounded gap spread removes one logarithm).} For nonempty \(\mathcal I\), put \(q_\Delta=\lceil\log_2(\Delta_{\max}/\Delta_{\min})\rceil\) and \(\Xi_T(d)=d\log(e\min\{d,T\})\), with the same zero/infinity conventions. For \(D=D_c\) or \(W_c\) and \(C_\Delta=6160+3088q_\Delta\), unit-width BHO satisfies
\begin{equation}
R_T\le\min\left\{T, \frac{18473S_T}{\alpha^2} +\frac{C_\Delta\Xi_T(D)}{\alpha}\right\}.
\tag{8}
\end{equation}
The high-probability form and positive-width constants are in App. A.8; Theorem 1 extends to arm-specific slack and to regret on a subset of suboptimal arms (App. B).

\subsection{Gap-Weighted Delays and Same-Run Guarantees}
Calendar-rate blocks (RATE, Algorithm 2 in App. Q) pull one arm per block. Each arm's block lengths are \(2,4,8,\ldots\). After a completed block of length \(2n\), RATE caps that arm's historical index at \(1-\zeta/n\) plus a confidence width, where \(\zeta\) sums the deficits \(1-y\) of all the arm's reports that arrive in the block's second half, including reports of earlier pulls. At each block boundary it selects the smallest label with the largest historical index.

\textbf{Proposition 1 (calendar-rate blocks).} Fix \(c\in[0,1)\) and finite ordinary or weighted cutoffs \(h_i\). Set \(\delta=(T+1)^{-2}\), \(\ell_{\rm rate}=\log(2KT(T+1)^3)\), and \(S_{\rm rate}=\ell_{\rm rate}\alpha^{-2}\sum_{i\in\mathcal I}\Delta_i^{-1}\). RATE satisfies
\begin{equation}
R_T^{\rm rate}\le\min\Bigl\{T,\,8\sum_{i\in\mathcal I}\Delta_ih_i+1024S_{\rm rate}+\delta T\Bigr\}.
\tag{9}
\end{equation}
No optimal-arm reports are required. For two arms, (9) is \(O(\Delta h+\log(eT)/(\alpha^2\Delta))\). It supplies a gap factor for the sum term, motivated by the conjecture after Theorem 6 of Lancewicki et al.~(2021); their lower bound uses \(d(1-2\Delta)\). At \(c=1/4\), \(\sum_i\Delta_ih_i\le\sum_ih_i\) improves the delay term of their Optimistic-UCB expression, though a sum can exceed a maximum by a factor of \(K-1\).

\begin{table*}[t]
\caption{\textbf{Guarantees and lower bounds, up to absolute constants.} Upper bounds are clipped at \(T\); RATE/NC-lean also include \(\delta T\). Let \(Z=\sum_{i\in\mathcal I}\Delta_i^{-1}\). Prior rows use a unique optimum. For the L21 rows, except for PSE's free \(q_i\), use \(c=1/4\), \(d_i^{\rm q}=d_i(1-\Delta_i/4)\), \(D=\max_{i\in\mathcal I}d_i^{\rm q}\), and \(D_*=d_*(1-\Delta_{\min}/4)\). BHO, RATE, NC-lean and PTB use their stated fixed \(c<1\). Rows compare guarantees, not actual regrets.}
\vskip -0.05in
\begin{center}
\begin{small}
\begin{tabular}{>{\raggedright\arraybackslash}p{103.9pt}>{\raggedright\arraybackslash}p{111.7pt}>{\raggedright\arraybackslash}p{134.6pt}>{\raggedright\arraybackslash}p{87.8pt}}
\toprule
\textbf{Result} & \textbf{Statistical term} & \textbf{Delay term} & \textbf{Optimal arm; assumptions/proof} \\
\midrule
OPSE (L21, Thm. 5) & \((\log T)Z\) & \((\log K)(D+D_*)\) & Finite \(D_*\); proved there \\
PSE (L21, Thm. 3) & \((\log T)\sum_{i\in\mathcal I}(q_i\Delta_i)^{-1}\) & \((\log T)(\log K)\max_{i\in\mathcal I}\Delta_id_i(q_i)\) & No optimal-arm cutoff; reward-independent delays \\
Optimistic-UCB (L21, Eq.~8) & \((\log T)Z\) & \(\sum_{i\in\mathcal I}d_i^{\rm q}\) & Unrestricted; stated without expanded proof \\
BHO (Theorems 1--2) & \(LZ/\alpha^2\) & \(\Phi_T(D_c)/\alpha\) or \(\Phi_T(W_c)/\alpha\) & Unrestricted; Apps. A--B, M \\
RATE (Proposition 1) & \(\ell_{\rm rate}Z/\alpha^2\) & \(\sum_i\Delta_ih_i\) & Unrestricted; App. Q \\
NC-lean (Theorem 3) & \(\ell_CZ/\alpha^2\) & Two full bounds in (10) & Unrestricted; same-run proof, App. R \\
PTB (Theorem 4) & \(LZ/\alpha^2\) & \(\Phi_T(W_c)/\alpha\) & Optimal-arm reporting unrestricted; App. V \\
Every policy (Theorem 5) & Inverse-square slack lower bound & Zero cutoffs & Constructed family; App. C \\
Every policy (Theorem 6) & Linear regret at \(c=1\) & Zero critical cutoffs & Uniform impossibility; App. D \\
Every policy (Theorem 7) & -- & \(\Omega(G\log K)\) & Growing-\(K\) family; \(G=\max_i\Delta_i h_i\); App.~U \\
\bottomrule
\end{tabular}
\end{small}
\end{center}
\vskip -0.1in

\end{table*}
NC-lean (Algorithm 3, App. R) starts with one all-arm cohort. Among cohorts whose best historical index is within a confidence width of the largest, it serves one of minimum exposure \(x\); members a width below the cohort's best split off, and the \(m\) others are sampled uniformly for a fixed \(2mx\)-round block, which scores second-half arrivals and doubles exposure. Its logarithm is \(\ell_C=\log(8K^2(T+1)^3J)\), \(J=1+\lfloor\log_2(T+1)\rfloor\).

For \(c\in[0,1)\), let \(h_i=w_i(c)<\infty\) and \(S_C=\ell_C\alpha^{-2}\sum_{i\in\mathcal I}\Delta_i^{-1}\). Let \(M_i\) be the cohort size in the last started block containing \(i\), including truncated blocks and members never drawn.

\begingroup\clubpenalty=10000
\textbf{Theorem 3 (same-run maximum and cohort-sum guarantees).} NC-lean satisfies
\begin{equation}
R_T\le\min\left\{\begin{array}{c}\displaystyle T,\\\displaystyle 128S_C+8\mathbb E\sum_{i\in\mathcal I}\frac{\Delta_i h_i}{M_i}+\delta T,\\\displaystyle 524288S_C+\frac{6420\Phi_T(W_c)}{\alpha}+\delta T\end{array}\right\}.
\tag{10}
\end{equation}\endgroup
Both bounds on predictable regret hold on one event for the same trajectory (App. R). Taking expectations adds a single \(\delta T\) term, with \(\delta=(T+1)^{-2}\). The learner receives neither \(c\) nor cutoffs. Since \(M_i\ge1\), the random sum implies a deterministic sum bound; expressing its group benefit through deterministic instance quantities remains open. App. R.1 explains why the two cohort charges coexist.

\begin{samepage}
\textbf{Corollary 2 (reward-dependent separation).} For integers \(n\ge64\), let arm 1 give reward zero immediately or reward one with infinite delay, each with probability \(1/2\). Let arm 2 have Bernoulli mean \(1/2-1/n\) and deterministic delay \(n^3\). At \(T=\lfloor n^4/16\rfloor\), native RATE and NC-lean have regret \(O(n^2+n\log n)\), whereas unit-width BHO and SE, PSE and OPSE have regret \(\Theta(n^3)\) (App. J.2).\par
\end{samepage}

SE and OPSE cannot eliminate arm 2 because arm 1 reports only zeros and its lower confidence bound cannot exceed arm 2's upper bound. Once both arms meet PSE's first report quota, PSE instead eliminates arm 1 with high probability. For BHO, pending arm-2 pulls keep its historical index high. On the concentration event, if arm 1 reaches \(b=O(n^2\log n)\) pulls, its eligibility threshold falls below that index. Subsequent arm-2 pulls sustain the backlog, so arm 1 cannot exceed \(b\) pulls. RATE and NC-lean pay for delay at the gap scale \(\Delta_2h_2=n^2\).

\begin{samepage}
\textbf{Corollary 3 (RATE loses on a common long delay).} Let \(K-1=D=n\ge2\), \(T=n^3\), with a unique last-labeled optimum and suboptimal arms of common fixed gap \(\Delta\) and deterministic delay \(D\). RATE incurs at least \(\Delta n(n+1)\), while BHO has \(O(n\log n)\) regret. The order separation persists after averaging over a uniform relabeling (App. J.1).\par
\end{samepage}

\subsection{Pending-Triggered Balancing}
PTB retains (1)--(3), sets \(P_i=m_i-n_i\), and selects a leader \(g\) maximizing \(H_i\), with ties resolved by \((m_i,-H_i,i)\). It chooses \(g\) if \(m_g=0\) or \(P_g/m_g\le\beta_g\); otherwise it uses Algorithm 1's count priority on the full \(\mathcal C_1\). It retests and updates all indices each round, using \(O(K)\) storage.

\begin{samepage}
\textbf{Theorem 4 (protection and benign efficiency).} Fix \(c\in[0,1)\) and set \(\alpha=1-c\). With finite \(W_c\), PTB satisfies
\[
R_T\le\min\{T,18473S_T/\alpha^2+9264\Phi_T(W_c)/\alpha\}.
\]
The same bound holds for \(\Psi_T\) with probability at least \(1-(T+1)^{-2}\).\par
\end{samepage}

For fixed finite \(K\), deterministic rewards with a unique optimum and fixed positive gaps, and an unknown fixed bounded delay support, the same learner satisfies, for \(i\in\mathcal I\), as \(T\to\infty\),
\[
\frac{\mathbb E N_i(T)}{\ell_T}\longrightarrow\frac1{2\Delta_i^2}, \qquad \ell_T=\log(4KT(T+1)^2).
\]
On the concentration event in App. V.2, every mature suboptimal choice uses count priority. For a fixed delay bound \(d_0\), the trigger is inactive throughout the run once \(\ell_T\ge2d_0\) (App. V.4). For deterministic immediate rewards, the coefficient is one quarter of unit-width BHO's (App. N).

\section{One Interval per Optimism Bin}
Fix finite admissible weighted cutoffs \(h_i\), \(D=\max_i h_i\), and \(\gamma_i=\alpha\Delta_i\). An event \(E\) of probability at least \(1-\delta\) controls latent reward and cutoff-weighted deficit prefixes in adaptive pull order. Every historical index is then at least its arm's mean, so \(M(t)\ge\mu^*\). Define
\begin{equation}
B_i(t,h_i)=m_i(t)-m_i(\max\{0,t-h_i\}).
\tag{11}
\end{equation}
Old deficits have expected mass at most \(c\Delta_i\); each recent pull contributes at most one. At every positive count, including idle or ineligible arms,
\begin{equation}
\rho_i(t)\le\mu^*-\gamma_i+\frac{B_i(t,h_i)}{m_i(t)}+3\beta_i(t).
\tag{12}
\end{equation}
The analysis can thus track hidden ages without revealing them to the learner.

Selections before \(4608\ell/\gamma_i^2\) pulls contribute the statistical term. Later, eligibility and (12) force \(B_i/m_i\ge23\gamma_i/24\). Since \(B_i\le h_i\), these mature selections require \(h_i\gamma_i\ge4416\ell\); immediate feedback therefore incurs only the statistical term. Group mature selections by \(M(t)-\mu^*\), splitting each dyadic scale \([a,2a)\) into 32 bins; Figure 1 illustrates one bin.

\begin{figure*}[t]
\centering
\includegraphics[width=\textwidth]{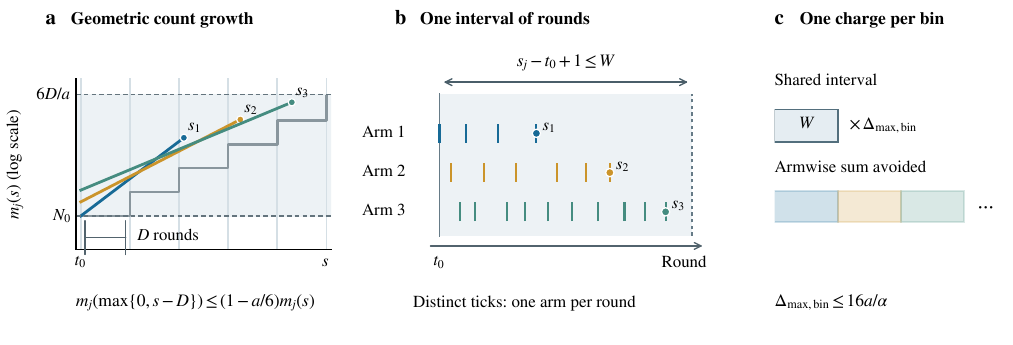}
\caption{\textbf{One shared interval per high-optimism bin (on \(E\); schematic, not to scale).} \textbf{a,} A charged arm's count starts at no less than \(N_0=81\ell/(8a^2)\), grows by a factor of at least \((1-a/6)^{-1}\) per \(D\) rounds, and stays at most \(6D/a\). \textbf{b,} Hence all charged decisions, one arm per round, fit in one interval of length \(W=D(1+6\log\min\{D,T\}/a)\). \textbf{c,} The bin costs at most \((16a/\alpha)W\le(96D/\alpha)\log(e\min\{D,T\})\): one interval times the largest charged gap, not a sum of armwise waits.}

\end{figure*}
\begin{samepage}
\textbf{Lemma 1 (one interval per bin).} On \(E\), with \(D\ge1,T\ge2\), fix a nonempty high-optimism bin at scale \(0<a\le1\) and its mature selections with \(\gamma_i\le16a\). Let \(t_0\) be their first time and \(s_j\) arm \(j\)'s last such selection. Every charged arm satisfies \(m_j(t_0)\ge81\ell/(8a^2)\) and \(B_j(s,h_j)\ge(a/6)m_j(s)\) throughout \(s\in[t_0,s_j]\), and
\begin{equation}
s_j-t_0+1\le D\Bigl(1+\frac{6\log\min\{D,T\}}{a}\Bigr).
\tag{13}
\end{equation}
\end{samepage}
First, eligibility and maturity at \(s_j\) imply \(H_j(s_j)\ge\mu^*+b-a/6\), where \([b,b+a/32)\) is the fixed bin. Historical monotonicity transports that lower bound to \(t_0\). If \(m_j(t_0)<N_0=81\ell/(8a^2)\), its width exceeds \(2a/9\), giving \(H_j(t_0)+\beta_j(t_0)>\mu^*+b+a/18>M(t_0)\). It would be eligible. The selected mature arm, however, has count at least \(18\ell/a^2>N_0\), contradicting minimum-count priority. Thus every future charged arm already has the stated common count floor at the first charge. Its width stays at most \(2a/9\), and the completion bound gives the recent-pull fraction in Lemma 1 at every intervening time. Eligibility need not persist between the two charged decisions.

Second, (12) gives \(m_j(\max\{0,s-D\})\le(1-a/6)m_j(s)\). Across each complete \(D\)-round step, the count therefore grows by at least \((1-a/6)^{-1}\). Yet one action per round gives \(B_j\le D\), so its endpoint count is at most \(\min\{T,6D/a\}\). Relative to the starting floor, the count ratio satisfies
\[
\begin{gathered}
R_H=\min\left\{\frac{8a^2T}{81\ell},\frac{48aD}{81\ell}\right\}
\le\min\{D,T\}.
\end{gathered}
\]
For a nonempty bin this ratio is at least one. The logarithmic count growth therefore permits at most \((6/a)\log\min\{D,T\}\) complete steps. Writing \(s_j-t_0=qD+r\), with \(0\le r<D\), bounds the inclusive number of rounds by \(D(q+1)\), giving (13) (Apps. A.3--A.4).

The charged decisions can be scattered across arms, but all occur between the same first charge and the last of the \(s_j\). One action per round makes this one interval a bound on their total number. Every charged gap is at most \(16a/\alpha\), so multiplying (13) by that gap costs at most \((96D/\alpha)\log(e\min\{D,T\})\) per bin. Apart from total regret at most \(9/\alpha\), the remaining mature charges have optimism below \(\gamma_i/8\). Group their effective gaps into bands \([\varepsilon,2\varepsilon)\). The same starting-count argument gives floor \(512\ell/(81\varepsilon^2)\), recent-pull fraction at least \(\varepsilon/8\), and at most \((16D/\alpha)\log(e\min\{D,T\})\) regret per band. Maturity bounds both the high-optimism and low-gap scale counts by \(3\log(e\min\{D,T\})\). The 32 high bins per scale and the low bands then contribute coefficient \(3(32\cdot96+16)=9264\). Early selections cost \((4608\ell/\alpha^2+1)\sum_i\Delta_i^{-1}\); using \(\ell\le4L\) and adding the small remaining and failure costs gives coefficient 18473 in (5). Bounded gap spread limits the scale sum instead and gives (8).

\section{Information Limits and Policy Costs}
\textbf{Theorem 5 (inverse-square statistical slack is necessary).} Fix \(c\in[0,1)\), \(\alpha=1-c\), \(0<\Delta\le1/8\), \(K\ge2\), and \(T\ge1\). There are \(K\) admissible instances with \(D_c=W_c=0\) and gaps in \([\Delta,2\Delta]\) such that every policy satisfies
\begin{equation}
\max_jR_T(P_j)\ge \frac{K-1}{256\alpha^2\Delta} \log\left(1+\frac{T\alpha^2\Delta^2}{K-1}\right).
\tag{14}
\end{equation}
Changing one arm's visible distribution by order \(\alpha\Delta\) while adjusting hidden rewards creates a testing cost on its pulls. The proof needs no asymptotic-efficiency assumption (App. C). In the logarithmic regime \(T\alpha^2\Delta^2\gg K-1\), (14) exhibits the inverse-square slack factor.

\begin{samepage}
\textbf{Theorem 6 (critical-cutoff impossibility).} (i) At \(c=1\), a fixed infinite-delay family with equal positive gaps and \(d_i(1-\Delta_i)=0\) has linear minimax regret. (ii) For every \(T\ge1\), two finite-support two-arm instances with gap \(1/8\), whose suboptimal arm reports at once with probability above \(7/8\) (zero critical cutoff), satisfy
\begin{equation}
\max\{R_T(P_T),R_T(Q_T)\}\ge T/32.
\tag{15}
\end{equation}
\end{samepage}
In (i), the instances differ only in which arm's never-reported mass \(\Delta\) has reward one, so all observable laws coincide (App. D.1). Part (ii) holds at every horizon, with finite-support laws that depend on \(T\). For \(K\ge2\), \(T\ge1\), integer \(d\ge0\), \(c\in[0,1]\), and \(0<\Delta\le1/8\), a family with \(D_c=d\) forces every policy's worst-case regret to be at least \(\Delta(1-1/K)\min\{T,d+1\}\) (Proposition D.1). A maximum-delay scale is necessary at fixed gaps.

\begin{samepage}
\textbf{Theorem 7 (multiscale delay information).} Fix \(c=1/4\). For integers \(m,H\ge1\), set \(K=2^m\) and \(T=HK\). There is a finite family of deterministic instances, each with one optimum and \(G=\max_i\Delta_i h_i=H\) under both ordinary and weighted cutoffs, such that every learner satisfies
\begin{equation}
\max_P R_T(P)\ge\frac{Hm}{4}=\frac{G\log K}{4\log2}.
\tag{16}
\end{equation}
\end{samepage}

At scale \(j\), \(2^{m-j}\) arms have gap \(2^{-j}\) and delay \(d_j=H2^j\). The optimum never reports. Assign types uniformly to labels and reveal each type for free at the end of round \(d_j\), before any of its reports can arrive. During rounds \(d_{j-1}+1\) to \(d_j\), with \(d_0=0\), half of the unrevealed labels are of type \(j\). Their assignment remains uniform given the visible history, so each unrevealed label has conditional expected gap at least \(2^{-j-1}\). Earlier types have larger gaps. For \(j\ge2\), this phase lasts \(H2^{j-1}\) rounds and costs at least \(H/4\) in expectation. The first phase costs at least \(H/2\). Averaging over the fixed assignments gives (16) for the worst assignment.

Taking \(H=K^2\) makes the statistical term \(o(G)\) while \(G\log K=o(T)\), ruling out a uniform \(O(S_T/\alpha^2+G)\) bound (App. U).

BHO itself pays inverse slack in delay: for \(\alpha\in(0,1]\), \(0<\Delta\le1/32\), integer \(d\ge2(b+1)\), \(b=\lfloor2\ell/(9\alpha^2\Delta^2)\rfloor+1\), \(\ell=\log(8T(T+1)^2)\), a two-arm instance at \(T=\lfloor d/(16\alpha\Delta)\rfloor\) has \(D_c=d\) and \(R_T^{\rm BHO}\ge d/(128\alpha)\) (Proposition E.1).

With deterministic immediate rewards, BHO's leading counts are \((1+\theta)^2\ell/(2\Delta_i^2)\); unit width pays four times historical greedy's coefficient (App. N). Smaller widths retain (6) with larger constants, while PTB retains protection and the low coefficient (Theorem 4). Historical greedy incurs larger-order regret than BHO on the common-delay family of Proposition I.1, which uses at least 65,536 arms and horizon \(T=K^3\).

\section{Experiments}
The experiments measure finite-horizon cost and sensitivity to reporting. PTB is outside their scope. Learners share potential rewards and delays within each seed, and seeds are the sampling units. Baselines include historical greedy (G), OUCB-current, OPSE, and BHO-half. SE and PSE are compared under their reward-independent assumptions and also stress-tested under dependent reporting (Lancewicki et al., 2021, Algs. 4--5 with Proc. 2). Primary intervals are nominal Bonferroni-\(t\) simultaneous intervals over seeds; a fixed-SE max-absolute-\(z\) bootstrap is the sensitivity check (Apps. L, O, P).

\begin{figure*}[t]
\centering
\includegraphics[width=\textwidth]{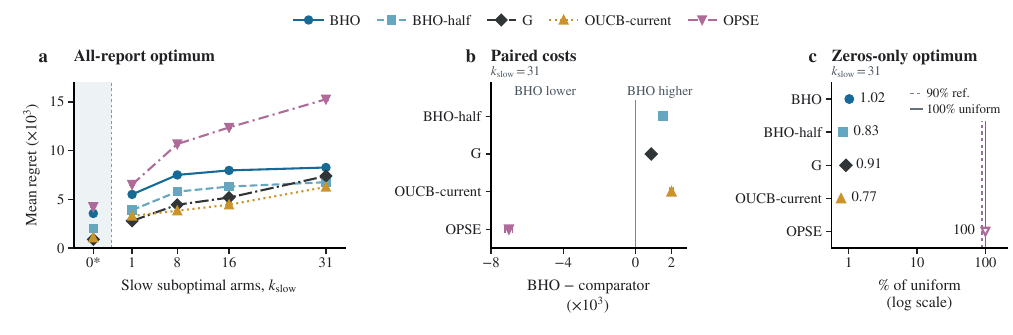}
\caption{\textbf{Reporting robustness coexists with exploration costs.} \(K=32\), \(T=2^{20}\), Bernoulli means 0.9 for the optimum and 0.1 otherwise, 32 paired seeds (App. P). \textbf{a,} All-report optimum; \(k_{\mathrm{slow}}\) suboptimal arms have delay 4096, and the shaded \(0^*\) strip is the zero-delay control. \textbf{b,} BHO minus each comparator at \(k_{\mathrm{slow}}=31\) with an all-report optimum. Positive values mean higher BHO regret. \textbf{c,} Each learner's regret at \(k_{\mathrm{slow}}=31\) with a zeros-only optimum, as a percentage of expected uniform regret (log axis; dashed 90\%, solid 100\%; open markers at least 90\%). OPSE attains uniform regret, while the four completion learners repeat their all-report trajectories, because completion by one restores each hidden success; BHO stays above G, OUCB-current and BHO-half. Intervals are marginal 95\% Student-\(t_{31}\) in a and c, and nominal Bonferroni-\(t_{31}\) over 20 distinct contrasts in b.}

\end{figure*}
\textbf{Synthetic designs.} The Bernoulli designs of Apps. L and O use means \(3/4\) for the optimal arm and \(1/2\) for the others, \(T=2^{18}\), and 32 paired seeds. App. L's fixed-delay design crosses \(K\in\{2,4,8,16,32,64\}\) with optimal-arm and suboptimal-arm delays of 0 or 2048 rounds. App. O's 59 reporting laws use \(K\in\{8,64\}\) and suboptimal delays up to 65,536. Suboptimal arms may also hide their failures, their successes, or outcomes independently of the reward, with total missing probability \(1/8\) or \(3/16\); the optimal arm reports every reward, never reports, or reports only its zeros. Twenty-five laws have reward-independent delays, the setting of SE and PSE, and two negative controls report only successes.

\textbf{Optimal-arm reports.} In the fixed-delay comparisons (App. L), delaying the optimal arm's reports by 2048 rounds raises OPSE's regret 1.28--4.06-fold in all 12 matched cells. BHO's regret decreases in nine cells and rises at most 4.22\% otherwise, G's decreases in ten and rises at most 1.50\% otherwise, and OUCB-current's rises by up to 50.40\%. In the eight laws of App. O whose optimal arm never reports, BHO beats G in none and OUCB-current in four, all at \(K=64\).

\textbf{Total regret.} In App. O, an optimal arm that never reports or reports only zeros has a negative lower bound in OPSE and cannot eliminate other arms; OPSE's regret equals uniform allocation's in all 32 such laws and in both negative controls. BHO's mean regret is below OPSE's in all 25 laws whose optimum reports every outcome and in 56 of all 59 (three ties); in 11 of these 56 it is at least 90\% of uniform regret. G beats BHO in 40 of the 59 laws, and BHO-half beats BHO in 57 (two ties). SE beats BHO in 10 of the 25 reward-independent laws, with BHO/SE ratios 1.19--2.86, including eight of the nine with suboptimal delay at least 8192. In the 24 zeros-only laws, which violate the assumptions of PSE and SE, each eliminates the optimal arm in all 768 environment--seed pairs, and uniform allocation also beats them (App. O). Returned-UCB (App. O.7) has lower regret than BHO in 19 laws, including all eight never-report laws, where its zero-count convention sends 96\%--99\% of pulls to the optimum, and higher in 40, including all 26 zeros-only and control laws, where it exceeds uniform regret; every interval excludes zero.

\textbf{Where BHO leads.} In App. L, BHO has the lowest mean regret in six of the 24 cells, exactly those with \(K\ge16\) and suboptimal delay 2048; there G's mean regret is 1.50--2.70 times BHO's and OUCB-current's 1.69--2.68 times. Paired intervals for BHO minus each comparator's delay excess (same-seed regret increase from suboptimal delay 0 to 2048) lie below zero in 32 of 36 contrasts with G, OUCB-current and OPSE (33 under the bootstrap). Against PSE and SE, BHO's intervals lie below zero in 20 and 14 of the 25 reward-independent laws and above zero in 4 and 10 (App. O). Ten deterministic common-delay constructions, smaller-\(K\) versions of the construction in Proposition I.1, are reported separately (App. L.2): with \(K\ge64\), G and OUCB-current incur 2.8--10.2 times BHO's regret, while with \(K\le32\) one of them matches or beats BHO in four of the six constructions.

{\brokenpenalty=10000 \textbf{Exploration cost.} Balanced exploration is costly at finite horizons. In Figure 2, BHO's regret is 0.44\%--1.02\% of uniform regret; raising the number of slow arms from 1 to 31 adds 2760 to it, versus 4603 for G, whose regret stays lower (Figure 2a), and all 20 primary Bonferroni-\(t\) intervals exclude zero (App. P). In App. L's fixed-delay design with every delay zero, BHO/G is 3.92--4.01, close to the factor four proved for deterministic rewards in Appendix N. At~\(K=64\) and suboptimal delay 65,536, raising the number of slow arms from 1 to 63 raises BHO's regret from 56.39\% to 99.55\% of uniform regret (43.16 points), versus increases of 48.28 points for G, 30.73 for OUCB-current and 16.50 for SE (App. O).\par}

\textbf{Cohort learners.} NC-wide shares NC-lean's rules but uses a wider logarithm and tolerance. At \(T\in\{2^{14},2^{16}\}\) with eight paired seeds per environment, both trail BHO in all 14 cells that satisfy our assumptions: NC-lean has 44\%--93\% lower mean regret than NC-wide but 19\%--135\% higher than BHO. RATE's is 1.79--5.58 times BHO's, and NC-wide exceeds 99.8\% of uniform regret in eight cells.

\textbf{Fixed classifiers.} Four classifiers on UCI optdigits (Alpaydin and Kaynak, 1998) define a 1,797-record empirical reward law. Reports arrive immediately (E0), are delayed by 128 rounds with probability \(1-\mu_i\) independently of correctness (E1), are delayed by 128 rounds exactly for errors (E2), or follow E2 except that the best classifier follows E2 with probability 0.05 on each pull and otherwise never reports (E3). In E0--E3, returned averages target true means as the returned counts grow. At \(T=2^{14}\), returned-UCB (App. T.3) incurs 6.0\%--15.4\% of uniform regret, against 39.1\%--40.7\% for BHO and 75.4\%--77.4\% for NC-lean. Across all eight channel--horizon cells (32 paired seeds), returned-UCB is lowest and NC-lean exceeds BHO.

\textbf{Selective confirmation.} In E4 and E4b, correct predictions of the best classifier, or of the two best, are confirmed with probability \(p\in\{0,0.05\}\) and otherwise never. For a model of accuracy \(\mu\) that reports every error and confirms each correct prediction with probability \(p\), the returned accuracy converges to \(p\mu/(1-\mu+p\mu)\) as its pull count diverges.

{\clubpenalty=10000 At \(p=0.05\), the best model has limiting returned accuracy 0.661, below even the two lowest-accuracy models. In E4b the second-best model has an infinite ordinary cutoff but a zero weighted cutoff for \(c<1\) because every error reports immediately. BHO's completed reward sum counts every pull except those with reported errors. Under the same draws and ties, hiding correct predictions therefore leaves BHO's trajectory unchanged (Apps. O.5 and T.5).\par}

BHO has 18.9\% of uniform regret at \(T=2^{16}\); returned-UCB has 30.8\%--32.0\% in E4 and 133.1\%--168.4\% in E4b, and is still lower in E4 for \(T\le2^{14}\) and in E4b at \(p=0.05\), \(T=2^{12}\) (App. T.5).

\section{Related Work}
Lancewicki et al.~(2021) formulate the closest protocol. Their PSE already gives a gap-weighted suboptimal maximum under reward-independent delays (Table 1). Wu and Wager (2022b) allow arm-specific infinite delays under independence but retain an optimal-arm quantile. Joulani et al.'s (2013) queue reduction uses independence to preserve iid observations. Outcome-dependent reporting also motivates proportional-hazards partial likelihood (Wu and Wager, 2022a).

Tang et al.~(2024) use a known reward--delay relation and the originating pull time to construct censored rewards. With payoff proportional to delay and a known cap, Schlisselberg et al.~(2025a) use the second-largest expected delay. Other delay summaries support best-of-both-worlds and linear-bandit bounds (Masoudian et al., 2022; 2024; Schlisselberg et al., 2025b; 2026). Van der Hoeven and Cesa-Bianchi (2022) pursue comparator-loss/delay dependence in a nonstochastic partially concealed setting.

Classical missing-data theory states when the reporting mechanism can be ignored (Rubin, 1976). For bounded outcomes, assigning the support endpoints to unobserved rewards gives selection bounds (Manski, 1989); completion by one uses their upper endpoint. Decision learning without value identification connects to partial monitoring (Bartók et al., 2014). Guinet et al.~(2022) study bandits under censorship. Mahrooghi et al.~(2025) use mediators and completeness for identification under nonignorable missingness. Here suboptimal-arm cutoffs suffice for decision learning. Delayed-conversion methods use delay laws, tails or interaction ages (Vernade et al., 2017; 2020; Manegueu et al., 2020; Gigli and Stella, 2024); conversion-only feedback lacks our explicit zeros.

\newpage
Arrival-window estimation appears with bridge periods (Pike-Burke et al., 2018), composite-feedback blocks (Wang et al., 2021) and multi-batch reductions (Yang et al., 2023). Batched policies are well studied (Perchet et al., 2016). Our block analysis adds reward-dependent truncation and ancestry control. PTB retains BHO's shared-interval accounting.

\section{Conclusion and Limitations}
On fixed instances with finite suboptimal weighted cutoffs for some \(c<1\), BHO and PTB have sublinear regret without requiring optimal-arm reports. Theorem 7 rules out a uniform \(O(S_T/\alpha^2+G)\) bound, and a general \(O(S_T/\alpha^2+G\log K)\) upper bound remains open. Under Theorem 4's deterministic bounded-delay conditions, PTB attains the historical-greedy exploration coefficient.

Open questions include general stochastic-reward efficiency for PTB, the necessity of inverse delay slack or squared cutoff logarithms for all learners, a \(K\)-independent single-logarithm bound for heterogeneous gaps, and reports crossing restarts. The explicit constants are large. In the admissible Bernoulli designs of Appendix K, cutoff--gap products lie below the proof's maturity threshold. Slow-arm scans do not establish maximum-versus-sum rates, and equal-gap designs do not test Corollary 1. NC-lean was designed on its evaluation cells, and App. S's good-event bounds omit the failure term. The fixed-classifier study measures costs under simulated reporting, without establishing deployment value.

\pdfbookmark[0]{Code and Data}{r060.nav.1}
\section*{Code and Data}
Code and saved experimental results are included in the accompanying supplementary archive. Its README documents the fixed reward populations, seed coupling, available action logs and reconstruction instructions. It also indexes the complete experimental tables and paired contrasts.

\pdfbookmark[0]{Impact Statement}{r060.nav.2}
\section*{Impact Statement}
Selective reporting can bias model comparisons and decisions based on them. Applications should assess reporting mechanisms and distribution changes before using these guarantees. The experiments use public digit-recognition data with simulated feedback and involve no newly collected participant data.

\clearpage
\pdfbookmark[0]{References}{r060.nav.3}
\section*{References}
\begingroup
\setlength{\parindent}{0pt}
\interlinepenalty=10000

\noindent\hangindent=10pt\hangafter=1 Alpaydin, E., and Kaynak, C. (1998). \emph{Optical Recognition of Handwritten Digits} {[}Dataset{]}. UCI Machine Learning Repository. \href{https://doi.org/10.24432/C50P49}{DOI:10.24432/C50P49}.

\noindent\hangindent=10pt\hangafter=1 Bartók, G., Foster, D. P., Pál, D., Rakhlin, A., and Szepesvári, C. (2014). Partial monitoring---classification, regret bounds, and algorithms. \emph{Mathematics of Operations Research}, \textbf{39}(4), 967--997. \href{https://doi.org/10.1287/moor.2014.0663}{DOI}.

\noindent\hangindent=10pt\hangafter=1 Chen, X., Lee, K. M., Villar, S. S., and Robertson, D. S. (2022). Some performance considerations when using multi-armed bandit algorithms in the presence of missing data. \emph{PLoS ONE}, \textbf{17}(9), e0274272. \href{https://doi.org/10.1371/journal.pone.0274272}{DOI: 10.1371/journal.pone.0274272}.

\noindent\hangindent=10pt\hangafter=1 Garivier, A., Ménard, P., and Stoltz, G. (2019). Explore first, exploit next: The true shape of regret in bandit problems. \emph{Mathematics of Operations Research}, \textbf{44}(2), 377--399. \href{https://doi.org/10.1287/moor.2017.0928}{DOI: 10.1287/moor.2017.0928}. \href{https://arxiv.org/abs/1602.07182v3}{arXiv:1602.07182v3}.

\noindent\hangindent=10pt\hangafter=1 Gigli, M., and Stella, F. (2024). Bootstrap your conversions: Thompson sampling for partially observable delayed rewards. \emph{Proceedings of the Fortieth Conference on Uncertainty in Artificial Intelligence}, PMLR \textbf{244}, 1438--1452. \href{https://proceedings.mlr.press/v244/gigli24a.html}{PMLR}.

\noindent\hangindent=10pt\hangafter=1 Guinet, G., Amin, S., and Jaillet, P. (2022). Effective dimension in bandit problems under censorship. \emph{Advances in Neural Information Processing Systems}, \textbf{35}, 5243--5255. \href{https://papers.nips.cc/paper/2022/hash/228ffa71ce31ebbdebc6cf413a39cdce-Abstract-Conference.html}{NeurIPS}.

\noindent\hangindent=10pt\hangafter=1 Joulani, P., György, A., and Szepesvári, C. (2013). Online learning under delayed feedback. \emph{Proceedings of the 30th International Conference on Machine Learning}, PMLR \textbf{28}(3), 1453--1461. \href{https://proceedings.mlr.press/v28/joulani13.html}{PMLR}.

\noindent\hangindent=10pt\hangafter=1 Lancewicki, T., Segal, S., Koren, T., and Mansour, Y. (2021). Stochastic multi-armed bandits with unrestricted delay distributions. \emph{Proceedings of the 38th International Conference on Machine Learning}, PMLR \textbf{139}, 5969--5978. \href{https://proceedings.mlr.press/v139/lancewicki21a.html}{PMLR}. \href{https://arxiv.org/abs/2106.02436v1}{arXiv:2106.02436v1}.

\noindent\hangindent=10pt\hangafter=1 Mahrooghi, I., Moradi, M., Akbari, S., and Kiyavash, N. (2025). Multi-armed bandits with missing outcomes. \emph{Proceedings of the Forty-first Conference on Uncertainty in Artificial Intelligence}, PMLR \textbf{286}, 2844--2875. \href{https://proceedings.mlr.press/v286/mahrooghi25a.html}{PMLR}.

\noindent\hangindent=10pt\hangafter=1 Manegueu, A. G., Vernade, C., Carpentier, A., and Valko, M. (2020). Stochastic bandits with arm-dependent delays. \emph{Proceedings of the 37th International Conference on Machine Learning}, PMLR \textbf{119}, 3348--3356. \href{https://proceedings.mlr.press/v119/gael20a.html}{PMLR}. \href{https://arxiv.org/abs/2006.10459v1}{arXiv:2006.10459v1}.

\noindent\hangindent=10pt\hangafter=1 Manski, C. F. (1989). Anatomy of the selection problem. \emph{The Journal of Human Resources}, \textbf{24}(3), 343--360. \href{https://doi.org/10.2307/145818}{DOI: 10.2307/145818}. \href{https://www.jstor.org/stable/145818}{JSTOR}.

\noindent\hangindent=10pt\hangafter=1 Masoudian, S., Zimmert, J., and Seldin, Y. (2022). A best-of-both-worlds algorithm for bandits with delayed feedback. \emph{Advances in Neural Information Processing Systems}, \textbf{35}, 11752--11762. \href{https://doi.org/10.52202/068431-0854}{DOI: 10.52202/068431-0854}.

\noindent\hangindent=10pt\hangafter=1 Masoudian, S., Zimmert, J., and Seldin, Y. (2024). A best-of-both-worlds algorithm for bandits with delayed feedback with robustness to excessive delays. \emph{Advances in Neural Information Processing Systems}, \textbf{37}, 141071--141102. \href{https://doi.org/10.52202/079017-4479}{DOI: 10.52202/079017-4479}. \href{https://arxiv.org/abs/2308.10675}{arXiv:2308.10675}.

\noindent\hangindent=10pt\hangafter=1 Perchet, V., Rigollet, P., Chassang, S., and Snowberg, E. (2016). Batched bandit problems. \emph{The Annals of Statistics}, \textbf{44}(2), 660--681. \href{https://doi.org/10.1214/15-AOS1381}{DOI: 10.1214/15-AOS1381}. \href{https://arxiv.org/abs/1505.00369v3}{arXiv:1505.00369v3}.

\noindent\hangindent=10pt\hangafter=1 Pike-Burke, C., Agrawal, S., Szepesvári, C., and Grünewälder, S. (2018). Bandits with delayed, aggregated anonymous feedback. \emph{Proceedings of the 35th International Conference on Machine Learning}, PMLR \textbf{80}, 4105--4113. \href{https://proceedings.mlr.press/v80/pike-burke18a.html}{PMLR}.

\noindent\hangindent=10pt\hangafter=1 Rubin, D. B. (1976). Inference and missing data. \emph{Biometrika}, \textbf{63}(3), 581--592. \href{https://doi.org/10.1093/biomet/63.3.581}{DOI: 10.1093/biomet/63.3.581}.

\noindent\hangindent=10pt\hangafter=1 Schlisselberg, O., Cohen, I., Lancewicki, T., and Mansour, Y. (2025a). Delay as payoff in MAB. \emph{Proceedings of the AAAI Conference on Artificial Intelligence}, \textbf{39}(19), 20310--20317. \href{https://doi.org/10.1609/aaai.v39i19.34237}{DOI: 10.1609/aaai.v39i19.34237}. \href{https://arxiv.org/abs/2408.15158v2}{arXiv:2408.15158v2}.

\noindent\hangindent=10pt\hangafter=1 Schlisselberg, O., Lancewicki, T., Auer, P., and Mansour, Y. (2025b). Improved best-of-both-worlds regret for bandits with delayed feedback. \emph{Advances in Neural Information Processing Systems}, \textbf{38}, 2035--2072. \href{https://proceedings.neurips.cc/paper_files/paper/2025/hash/02f0ac0a323dc17d964d4bbf8a62e01b-Abstract-Conference.html}{NeurIPS}.

\noindent\hangindent=10pt\hangafter=1 Schlisselberg, O., Zhang, M., and Mansour, Y. (2026). Near-optimal stochastic linear bandits with delay. Accepted at \emph{NeurIPS 2026} (poster). arXiv:2606.16656v1. \href{https://arxiv.org/abs/2606.16656v1}{arXiv:2606.16656v1}. \href{https://neurips.cc/virtual/2026/poster/149837}{NeurIPS}.

\noindent\hangindent=10pt\hangafter=1 Tang, Y., Wang, Y., and Zheng, Z. (2024). Stochastic multi-armed bandits with strongly reward-dependent delays. \emph{Proceedings of the 27th International Conference on Artificial Intelligence and Statistics}, PMLR \textbf{238}, 3043--3051. \href{https://proceedings.mlr.press/v238/tang24c.html}{PMLR}.

\noindent\hangindent=10pt\hangafter=1 van der Hoeven, D., and Cesa-Bianchi, N. (2022). Nonstochastic bandits and experts with arm-dependent delays. \emph{Proceedings of the 25th International Conference on Artificial Intelligence and Statistics}, PMLR \textbf{151}, 2022--2044. \href{https://proceedings.mlr.press/v151/van-der-hoeven22a.html}{PMLR}.

\noindent\hangindent=10pt\hangafter=1 Vernade, C., Cappé, O., and Perchet, V. (2017). Stochastic bandit models for delayed conversions. \emph{Proceedings of the 33rd Conference on Uncertainty in Artificial Intelligence}, 730--739; reissued as PMLR \textbf{R15} (2026). \href{https://proceedings.mlr.press/r15/vernade17a.html}{PMLR}.

\noindent\hangindent=10pt\hangafter=1 Vernade, C., Carpentier, A., Lattimore, T., Zappella, G., Ermis, B., and Brückner, M. (2020). Linear bandits with stochastic delayed feedback. \emph{Proceedings of the 37th International Conference on Machine Learning}, PMLR \textbf{119}, 9712--9721. \href{https://proceedings.mlr.press/v119/vernade20a.html}{PMLR}.

\noindent\hangindent=10pt\hangafter=1 Wang, S., Wang, H., and Huang, L. (2021). Adaptive algorithms for multi-armed bandit with composite and anonymous feedback. \emph{Proceedings of the AAAI Conference on Artificial Intelligence}, \textbf{35}(11), 10210--10217. \href{https://doi.org/10.1609/aaai.v35i11.17224}{DOI: 10.1609/aaai.v35i11.17224}. \href{https://arxiv.org/abs/2012.07048v2}{Technical version}.

\noindent\hangindent=10pt\hangafter=1 Wu, H., and Wager, S. (2022a). Partial likelihood Thompson sampling. \emph{Proceedings of the Thirty-Eighth Conference on Uncertainty in Artificial Intelligence}, PMLR \textbf{180}, 2138--2147. \href{https://proceedings.mlr.press/v180/wu22c.html}{PMLR}.

\noindent\hangindent=10pt\hangafter=1 Wu, H., and Wager, S. (2022b). Thompson sampling with unrestricted delays. \emph{Proceedings of the 23rd ACM Conference on Economics and Computation}, 937--955. \href{https://doi.org/10.1145/3490486.3538376}{DOI: 10.1145/3490486.3538376}. \href{https://arxiv.org/abs/2202.12431v2}{arXiv:2202.12431v2}.

\noindent\hangindent=10pt\hangafter=1 Yang, Y., Zhong, H., Wu, T., Liu, B., Wang, L., and Du, S. S. (2023). A reduction-based framework for sequential decision making with delayed feedback. \emph{Advances in Neural Information Processing Systems}, \textbf{36}, 46362--46389. \href{https://proceedings.neurips.cc/paper_files/paper/2023/hash/915125efea950af378435518b3542e6a-Abstract-Conference.html}{NeurIPS}.
\par\endgroup
\clearpage
\onecolumn
\appendix
\begin{multicols}{2}
\pdfbookmark[0]{Appendix A. Upper-Bound Proof}{r060.nav.4}
\section*{Appendix A. Upper-Bound Proof}
\textbf{Roadmap.} Appendices A--B and M prove BHO's guarantees and extensions; V treats PTB. Information lower bounds are in C--D and U. Policy costs and separations are in E, I--J and N; Q--R analyze RATE and NC-lean. F--G discuss decision learning, H compares delay assumptions and K relates the Bernoulli designs to the maturity threshold. Experiments are in L, O--P and S--T.

\subsection*{A.1 A fixed-cutoff concentration event}
First take arbitrary \(\delta\in(0,1)\), with \(\ell=\log(4KT/\delta)\), and fix \(c<1\). Suppose \(\mathcal I\ne\varnothing\) and all required cutoffs are finite. For Theorem 2 set \(h_i=w_i(c)\), \(r_i=b_i(h_i)\le c\Delta_i\), and \(D=\max_i h_i\). For Theorem 1 we may use its ordinary cutoffs and the same weighted argument, because \(b_i(h_i)\le\Pr(D_i>h_i)\le c\Delta_i\). Write \(Z=\sum_{i\in\mathcal I}\Delta_i^{-1}\) and \(\gamma_i=\alpha\Delta_i\).

Represent each arm by an infinite iid stack of its marginal reward-delay pairs, using the next pair when selected. This preserves visible-history laws: conditional on any legal past and selected arm, the fresh vector's relevant component has that marginal law independently of the past; unused components generate no reports. The union bound needs no independence between arm stacks.

Require, for all \(i\) and every integer \(1\le m\le T\),
\begin{equation}
\begin{gathered}
\left|\frac1m\sum_{k=1}^mY_{i,k}-\mu_i\right|\\
\le\sqrt{\frac\ell{2m}}.
\end{gathered}
\tag{A1}
\end{equation}
For each \(i\in\mathcal I\), also require
\begin{equation}
\begin{gathered}
\sum_{k=1}^m(1-Y_{i,k})\mathbf1\{D_{i,k}>h_i\}\\
\le mr_i+\sqrt{m\ell/2}.
\end{gathered}
\tag{A2}
\end{equation}
The (A2) summands are iid in \([0,1]\), though possibly dependent on the (A1) rewards. Hoeffding and a union bound give failure probability at most \(3KT e^{-\ell}=3\delta/4\le\delta\). Call the event \(E\). Its uniform prefix bounds apply at adaptive counts without additional conditional independence.

\subsection*{A.2 Historical indices and the completion bound}
Work on \(E\). At a positive count \(m=m_i(t)\), let \(\overline Y_i(t)\) denote the mean of all latent rewards pulled from arm \(i\). Completing missing rewards as one gives the pathwise identity
\begin{equation}
\begin{gathered}
m\bigl(u_i^+(t)-\overline Y_i(t)\bigr)\\
=\sum_{\text{pending pulls of }i}(1-Y).
\end{gathered}
\tag{A3}
\end{equation}
It is nonnegative. Equation (A1) implies \(\rho_i(t)\ge\mu_i\), so the minimum in (2) satisfies \(H_i(t)\ge\mu_i\), including at zero count. Thus \(M(t)\ge\mu^*\). Whenever an already-pulled arm is selected,
\begin{equation}
\rho_i(t)\ge H_i(t)\ge M(t)-\beta_i(t).
\tag{A4}
\end{equation}
A pull made at or before round \(t-h_i\) with delay at most \(h_i\) has returned by the end of round \(t\). Consequently every pending old pull has delay exceeding \(h_i\), while the number of newer pulls is \(B_i(t,h_i)\). For \(t<h_i\), all pulls are newer. In either case,
\begin{equation}
\begin{gathered}
\sum_{\rm pending}(1-Y)\\
\le\sum_{k=1}^{m}(1-Y_{i,k})\mathbf1\{D_{i,k}>h_i\}\\
+B_i(t,h_i).
\end{gathered}
\tag{A5}
\end{equation}
Double-counting some new pulls preserves the upper bound. Combining (A1)--(A3) and (A5), at every positive-count calendar time,
\begin{equation}
\begin{gathered}
\rho_i(t)\le\mu_i+r_i+\\
\frac{B_i(t,h_i)}{m_i(t)}+3\beta_i(t)\\
\le\mu^*-\gamma_i+\\
\frac{B_i(t,h_i)}{m_i(t)}+3\beta_i(t).
\end{gathered}
\tag{A6}
\end{equation}
For ordinary cutoffs, \(q_i=\Pr(D_i>h_i)\) satisfies \(r_i=b_i(h_i)\le q_i\le c\Delta_i\), giving (A6) with the same \(\gamma_i=\alpha\Delta_i\). This weighted event proves both theorems without concentration of \(n_i(t)\).

\subsection*{A.3 Cold selections and calendar growth}
Let \(A=96\), \(Q_i=A^2\ell/(2\gamma_i^2)\). A selection made with pre-action count below \(Q_i\) is cold; otherwise it is mature. Each selection increases its arm's count by one, so arm \(i\) has at most \(Q_i+1\) cold selections. Since \(\Delta_i\le1/\Delta_i\),
\begin{equation}
\begin{gathered}
\Psi_{\rm cold}\le\\
\left[\frac{A^2\ell}{2\alpha^2}+1\right]Z.
\end{gathered}
\tag{A7}
\end{equation}
A mature selected arm has \(\beta_i\le\gamma_i/96\). On \(E\), (A4)--(A6) imply
\begin{equation}
\begin{gathered}
\frac{B_i(t,h_i)}{m_i(t)}\ge\gamma_i-4\beta_i(t)\\
\ge\frac{23}{24}\gamma_i,\\
\qquad h_i\gamma_i\ge4416\ell.
\end{gathered}
\tag{A8}
\end{equation}
The last inequality uses \(B_i\le h_i\) and \(m_i\ge4608\ell/\gamma_i^2\). Let \(\gamma_{\max}=\max_{i\in\mathcal I}\gamma_i\). If \(D\gamma_{\max}<4416\ell\), all bad selections are cold on \(E\), so (A7) bounds their entire regret. This includes \(D=0\); equality is not excluded by (A8). When \(T=1\), all selections are cold as well. For the calendar proof assume \(T\ge2\) and finite \(D\ge1\).

We use a deterministic endpoint-count lemma. Suppose arm \(j\) has \(m_j(t_0)\ge N>0\), and
\begin{equation}
B_j(s,h_j)\ge(g/2)m_j(s),\qquad 0<g\le1,
\tag{A9}
\end{equation}
at every integer \(s\in[t_0,s_j]\), where \(h_j\le D\). The one-action-per-round convention gives \(B_j(s,h_j)\le h_j\le D\). In particular, the last charged pre-action count satisfies \(m_j(s_j)\le U:=\min\{T,2D/g\}\). Let \(R=U/N\) and \(\lambda=-\log(1-g/2)\). A nonempty charged group has \(R\ge1\). For \(s\ge t_0+D\), \(m_j(s-D)\le m_j(s-h_j)\le(1-g/2)m_j(s)\). Write \(s_j-t_0=qD+r\), with integers \(q\ge0\) and \(0\le r<D\). Iteration from \(t_0+r\) gives \(N(1-g/2)^{-q}\le U\), and hence
\begin{equation}
\begin{gathered}
s_j-t_0+1\\
\le D\left(1+\left\lfloor\frac{\log R}{\lambda}\right\rfloor\right)\\
\le D\left(1+\frac{2\log R}{g}\right).
\end{gathered}
\tag{A10}
\end{equation}
The first inequality uses \(r+1\le D\). For a hindsight collection sharing \(t_0,N,g\) and satisfying (A9) until each arm's last charged selection, (A10) bounds the common interval, including idle and uncharged times, without summing arm durations. Positive-count arms with \(h_j=0\) cannot satisfy (A9).

\subsection*{A.4 High-optimism bins}
Put \(v(t)=M(t)-\mu^*\). Since there is a bad arm, \(\mu^*>0\), so \(0\le v(t)<1\). Let \(P=\lceil\log_2T\rceil+2\). The bands \([a,2a)\), \(a=2^{-r}\), \(r=0,\ldots,P-1\), cover every relevant \(v\ge1/T\). Divide each into 32 bins \([b,b+a/32)\), where \(b=a+ka/32\), \(k=0,\ldots,31\).

Fix a bin, and charge its mature bad selections with \(\gamma_i\le16a\). For a nonempty charge let \(t_0\) be its first time, \(F\) its charged-arm set, and \(s_j\) the last charged time of \(j\in F\). Eligibility and maturity imply
\begin{equation}
\begin{gathered}
H_j(s_j)\ge M(s_j)-\beta_j(s_j)\\
\ge\mu^*+b-a/6.
\end{gathered}
\tag{A11}
\end{equation}
Historical monotonicity transports (A11) to every \(s\le s_j\). Set \(N_0=81\ell/(8a^2)\). If \(m_j(t_0)<N_0\), its width exceeds \(2a/9\), including at zero count. Therefore
\[
\begin{gathered}
H_j(t_0)+\beta_j(t_0)\\
>\mu^*+b+(2/9-1/6)a\\
=\mu^*+b+a/18>M(t_0),
\end{gathered}
\]
so it is eligible, using \(1/18>1/32\). The actual mature selection instead has
\[
\begin{gathered}
m_i(t_0)\ge\frac{A^2\ell}{2\gamma_i^2}\\
\ge18\ell/a^2>N_0.
\end{gathered}
\]
This contradicts minimum-count selection. Consequently every \(j\in F\) has \(m_j(t_0)\ge N_0\) and \(\beta_j(s)\le2a/9\) throughout \([t_0,s_j]\). From (A6), (A11), and \(\rho_j(s)\ge H_j(s)\),
\begin{equation}
\begin{gathered}
\frac{B_j(s,h_j)}{m_j(s)}\\
\ge b-a/6+\gamma_j-2a/3\\
\ge a/6+\gamma_j\ge a/6.
\end{gathered}
\tag{A12}
\end{equation}
We apply (A10) with \(N=81\ell/(8a^2)\) and \(g=a/3\). The endpoint ratio is
\[
R_H=\min\left\{\frac{8a^2T}{81\ell},\frac{48aD}{81\ell}\right\}.
\]
Set \(M_D=\min\{D,T\}\). Since \(\ell>\log8>81/64\), both initial count bounds used here and below exceed one, and \(48a/(81\ell)<1\). Thus every nonempty high group has \(1\le R_H\le M_D\). Ratios below one imply an empty group. Every charged gap is at most \(16a/\alpha\), so the bin regret is at most
\begin{equation}
\begin{gathered}
\frac{16aD}{\alpha}\left(1+\frac{6\log M_D}{a}\right)\\
\le\frac{96D}{\alpha}\log(eM_D).
\end{gathered}
\tag{A13}
\end{equation}
Here \(a\le1\); charged decisions may be nonconsecutive, and eligibility need not be nested.

\subsection*{A.5 Low-optimism bands}
Charge mature selections with \(v(t)<\gamma_i/8\) and \(\gamma_i\ge1/T\). Partition their effective gaps into the same \(P\) bands \([\varepsilon,2\varepsilon)\); include \(\gamma_i=1\) in the \(\varepsilon=1\) band. Define \(t_0,F,s_j\) as before. The first charged selection has \(\gamma_i<2\varepsilon\), so \(M(t_0)<\mu^*+\varepsilon/4\). Future eligibility and maturity give
\begin{equation}
\begin{gathered}
H_j(s_j)\ge\mu^*-2\varepsilon/A\\
=\mu^*-\varepsilon/48.
\end{gathered}
\tag{A14}
\end{equation}
This bound also holds at all earlier times. Let \(N_0=512\ell/(81\varepsilon^2)\). If \(m_j(t_0)<N_0\), then \(\beta_j(t_0)>9\varepsilon/32\), and
\[
\begin{gathered}
H_j(t_0)+\beta_j(t_0)\\
>\mu^*+(9/32-1/48)\varepsilon\\
=\mu^*+25\varepsilon/96\\
>\mu^*+\varepsilon/4>M(t_0).
\end{gathered}
\]
Arm \(j\) is eligible, but the selected mature arm has count at least \(A^2\ell/(8\varepsilon^2)=1152\ell/\varepsilon^2>N_0\), contradicting minimum count. All future charged arms therefore have \(\beta_j(s)\le9\varepsilon/32\) throughout their intervals. The upper bound (A6) and (A14) imply
\begin{equation}
\begin{gathered}
\frac{B_j(s,h_j)}{m_j(s)}\\
\ge\gamma_j-\varepsilon/48-27\varepsilon/32\\
\ge13\varepsilon/96\ge\varepsilon/8.
\end{gathered}
\tag{A15}
\end{equation}
Take \(N=512\ell/(81\varepsilon^2)\) and \(g=\varepsilon/4\) in (A10). Here
\[
R_L=\min\left\{\frac{81\varepsilon^2T}{512\ell}, \frac{648\varepsilon D}{512\ell}\right\}.
\]
Because \(\ell>81/64\), \(648\varepsilon/(512\ell)<1\), and \(N>1\). Hence a nonempty low group has \(1\le R_L\le M_D\), while \(R_L<1\) means it is empty. Charged gaps are less than \(2\varepsilon/\alpha\), so the band cost is at most
\begin{equation}
\begin{gathered}
\frac{2\varepsilon D}{\alpha}\\
\left(1+\frac{8\log M_D}{\varepsilon}\right)\\
\le\frac{16D}{\alpha}\log(eM_D).
\end{gathered}
\tag{A16}
\end{equation}
\subsection*{A.6 Finite scale counts, exhaustion, and expectation}
Every mature bad selection is charged above or is globally inexpensive. Those with \(\gamma_i<1/T\) cost at most \(1/\alpha\). Otherwise, for \(v\ge1/T\), its band \([a,2a)\) gives (A13) when \(\gamma_i\le16a\); if not, \(v<2a<\gamma_i/8\) gives (A16). For \(v<1/T\), selections with \(\gamma_i\le8/T\) cost at most \(8/\alpha\); the remainder satisfy the low charge. Weak inequalities cover boundaries; overlap only increases the bound.

Let \(M_D=\min\{D,T\}\), \(\gamma_{\min}=\min_{i\in\mathcal I}\gamma_i\), and retain \(P=\lceil\log_2T\rceil+2\). In the cold-only case \(D\gamma_{\max}<4416\ell\), set \(H=L_{\rm sc}=0\). Otherwise define
\begin{equation}
\begin{gathered}
\Gamma=\max\{\gamma_{\min},4416\ell/D\}\le\gamma_{\max},\\
\quad\\
H=\max\{0,\min\{P-1,\lfloor\log_2(16/\Gamma)\rfloor\}\},
\end{gathered}
\tag{A17}
\end{equation}
\begin{equation}
L_{\rm sc}=\max\{0,\min\{P,\lceil\log_2(2/\Gamma)\rceil\}\}.
\tag{A18}
\end{equation}
Mature selected arms have \(\gamma_i\ge\Gamma\). High charges require \(a\ge\Gamma/16\); \(a\le v<1\) excludes scale \(r=0\), leaving at most \(H\) scales with 32 bins each. Low charges require \(\varepsilon\le\gamma_i<2\varepsilon\), hence strictly \(\varepsilon>\Gamma/2\), leaving at most \(L_{\rm sc}\) scales. The floor in (A17) includes dyadic equality, the ceiling in (A18) excludes \(\varepsilon=\Gamma/2\), and \(\gamma_i=1\) belongs to the \(\varepsilon=1\) band. Thus (A7), (A13), and (A16) give on \(E\)
\begin{equation}
\begin{gathered}
\Psi_T\le[4608\ell/\alpha^2+1]Z+9/\alpha\\
+(3072H+16L_{\rm sc})D\log(eM_D)/\alpha.
\end{gathered}
\tag{A19}
\end{equation}
For a simpler cutoff-only expression, define
\[
\Phi_T(D)=\begin{cases}
0,&D=0,\\
D\log^2(e\min\{D,T\}),&1\le D<\infty,\\
\infty,&D=\infty.
\end{cases}
\]
If \(D\ge T\), both counts are at most \(P\le3\log(eT)=3\log(eM_D)\). For \(D<T\), (A8) makes each nonempty high scale satisfy \(a\ge276\ell/D\), and each nonempty low scale satisfy \(\varepsilon>2208\ell/D\). Allowing the unused \(r=0\) high scale gives counts at most \([1+\lfloor\log_2(D/(276\ell))\rfloor]_+\) and \([\lceil\log_2(D/(2208\ell))\rceil]_+\), respectively, where \([x]_+=\max\{x,0\}\). When a mature selection is possible these are each at most \(1+\log_2D\le3\log(eD)\); otherwise the counts are zero. This proves \(H,L_{\rm sc}\le3\log(eM_D)\) in both regimes. Hence
\begin{equation}
\begin{gathered}
\Psi_T\le\\
\left[\frac{96^2\ell}{2\alpha^2}+1\right]Z\\
+\frac9\alpha+\frac{9264\Phi_T(D)}\alpha,
\end{gathered}
\tag{A20}
\end{equation}
since \(3(32\cdot96+16)=9264\). As \(Z\ge1\), \(\ell>1\), and \(\alpha\le1\),
\begin{equation}
\Psi_T\le\min\left\{T, 4618\frac{\ell Z}{\alpha^2} +9264\frac{\Phi_T(D)}\alpha\right\}
\tag{A21}
\end{equation}
on \(E\): \(96^2/2=4608\) and \(Z+9/\alpha\le10\ell Z/\alpha^2\). In the cold-only case, (A7) yields the sharper bound with no calendar charge or tiny-class cost.

On \(E^c\), regret is at most \(T\), adding at most \(\delta T\) in expectation. At \(\delta=(T+1)^{-2}\), use \(\ell\le4L\), \(\delta T\le1\), and \(LZ/\alpha^2\ge1\). For \(T\ge2\), the logarithmic comparison follows from \(4KT(T+1)^2\le16KT^3\le(eKT)^4\). In particular, the refined count bound implies
\begin{equation}
\begin{gathered}
R_T\le\min\bigl\{T,\frac{18473LZ}{\alpha^2}\\
+\frac{(3072H+16L_{\rm sc})D\log(eM_D)}\alpha\bigr\},
\end{gathered}
\tag{A22}
\end{equation}
and (A21) gives (5) with \(18473=4\cdot4618+1\) and \(9264\); the coefficient \(18473\) also covers the high-probability statement at \(\delta=(T+1)^{-2}\). The scale counts \(H\) and \(L_{\rm sc}\) of (A17)--(A18) use the actual \(\ell\) of the run. In the cold-only case, \(18473LZ/\alpha^2\) alone is a valid expectation bound and high-probability threshold. For fixed finite \(D\), the delay term is at most \(9264D\log^2(eD)/\alpha\), which does not depend on \(T\); only the statistical term grows with \(\log T\).

\textbf{Lemma A.1 (selective balancing).} Fix the cutoffs and event \(E\) of A.1. Consider a learner with the completion, widths and historical updates (1)--(2). Suppose that, on \(E\), every choice belongs to \(\mathcal C_1(t)\), and every mature suboptimal choice minimizes the pull count over \(\mathcal C_1(t)\), with maturity defined in A.3. Then (A21) holds for this learner. At \(\delta=(T+1)^{-2}\), it has the expectation and high-probability bounds (5), with \(D=\max_i h_i\).

\textbf{Proof.} The prefix event, completion bound, cold count and calendar-growth lemma in A.1--A.3 apply to these updates. In A.4 and A.5, eligibility gives the historical-index lower bound at each arm's last charged selection. Count priority is used only at the first charged time \(t_0\), when the selected arm is mature and suboptimal, to establish the common starting-count floor. The remaining interval argument uses only (A6), historical monotonicity and nondecreasing counts. Thus (A13) and (A16) hold, and A.6's exhaustion, scale counts and expectation conversion give the stated bounds. The cold-only cases cover \(D=0\) and \(T=1\); empty suboptimal sets have zero regret.

\subsection*{A.7 The scalar oracle}
For any deterministic threshold \(F(c)\) above, including clipping at \(T\), the policy and \(\Psi_T\) stay fixed as \(c\) varies, with \(\Pr(\Psi_T>F(c))\le\delta\). Put \(f=\inf_cF(c)\), choose deterministic \(c_n\) approaching it, and let \(g_n=\min_{k\le n}F(c_k)\). Each \(g_n\) is a threshold at a deterministic member of that finite list, so \(\Pr(\Psi_T>g_n)\le\delta\). Since \(g_n\downarrow f\), continuity on the increasing events \(\{\Psi_T>g_n\}\) gives \(\Pr(\Psi_T>f)\le\delta\), without intersecting cutoff events. Expectation thresholds pass directly to their infimum. The argument includes refined counts, cold-only thresholds and deterministic qualifying cutoff vectors. Appendix M permits it at each fixed positive \(\theta\), but changing \(\theta\) changes the policy and regret variable, so no oracle over widths follows.

\subsection*{A.8 Gap-weighted summation}
The gap-spread index is
\[
q_\Delta=\left\lceil\log_2 \frac{\max_{i\in\mathcal I}\Delta_i}{\min_{i\in\mathcal I}\Delta_i} \right\rceil.
\]
Keep the fixed-cutoff event, cold threshold, high bins, low bands and exhaustion argument of Appendices A and M. Let \(\gamma_i=\alpha\Delta_i\), let \(\gamma_{\min},\gamma_{\max}\) be their extrema, and take finite \(D\ge1\); empty bad-arm sets give zero regret, \(D=0\) and \(T=1\) give the cold-only bound of A.3 and A.6, and infinite cutoffs give only the trivial bound \(T\). For fixed \(\theta\in(0,1]\), define
\begin{equation}
\begin{gathered}
f_\theta(a)=\max\bigl(\{\gamma_i:\gamma_i\le16a/\theta\}\cup\{0\}\bigr),\\
\qquad\\
\mathcal S_\theta=\sum_{j\ge0}\\
\frac{f_\theta(2^{-j})}{2^{-j}}.
\end{gathered}
\tag{A23}
\end{equation}
Terms vanish below \(\theta\gamma_{\min}/16\), so the sum is finite. Let \(n_\gamma\) count distinct effective gaps and \(m_\gamma\) occupied dyadic bands \([\varepsilon,2\varepsilon)\), assigning gap one to the \(\varepsilon=1\) band. Then
\begin{equation}
\begin{gathered}
\mathcal S_\theta\le\frac1\theta\\
\min\{32+16q_\Delta,32n_\gamma,64m_\gamma\},\\
\qquad m_\gamma\le1+q_\Delta.
\end{gathered}
\tag{A24}
\end{equation}
The corresponding profile coefficient and regret bound are
\begin{equation}
\begin{gathered}
\mathcal G_\theta=6N_\theta\mathcal S_\theta+16m_\gamma,\\
\qquad\\
R_T^{(\theta)}\le\min\left\{T, \frac{18473S_T}{\alpha^2\theta^2} +\frac{\mathcal G_\theta\Xi_T(D)}\alpha\right\}.
\end{gathered}
\tag{A25}
\end{equation}
The same numerical threshold holds with probability at least \(1-(T+1)^{-2}\). Its spread-only coefficient is
\begin{equation}
\begin{gathered}
\mathcal G_\theta\le\\
\left(\frac{192N_\theta}{\theta}+16\right)\\
+\left(\frac{96N_\theta}{\theta}+16\right)q_\Delta.
\end{gathered}
\tag{A26}
\end{equation}
To prove these statements, write \(M_D=\min\{D,T\}\). Before multiplying by a gap bound, the common interval in M.5 has length at most \(D(1+6\log M_D/a)\). Every high-bin charged arm has effective gap at most \(f_\theta(a)\). Since \(a\le1\), that bin costs at most
\begin{equation}
\begin{gathered}
\frac{D f_\theta(a)}\alpha\\
\left(1+\frac{6\log M_D}{a}\right)\\
\le\frac{6\Xi_T(D)}\alpha\frac{f_\theta(a)}a.
\end{gathered}
\tag{A27}
\end{equation}
Summing over \(N_\theta\) bins and all scales gives the first term of \(\mathcal G_\theta\). M.6 charges at most \(m_\gamma\) low bands, each costing \(16\Xi_T(D)/\alpha\). M.7's unchanged exhaustion leaves noncalendar cost at most \(4618\ell\sum_{i\in\mathcal I}\Delta_i^{-1}/(\alpha^2\theta^2)\); the conversion of A.6 at \(\delta=(T+1)^{-2}\) gives (A25).

For (A24), put \(u=\theta\gamma_{\max}/16\) and \(l=\theta\gamma_{\min}/16\). The positive dyadic sum over \(a\ge u\) satisfies \(\sum\gamma_{\max}/a<32/\theta\). There are at most \(q_\Delta\) dyadic scales in \([l,u)\), each contributing at most \(16/\theta\). Indeed, \(r\) such scales imply \(2^{r-1}<u/l\), including the empty interval when the gap ratio is one. The strict upper endpoint also handles dyadic ratios. Extreme occupied low-band indices differ by at most \(q_\Delta\): if their difference is \(r\ge1\), the extremal gap ratio is strictly greater than \(2^{r-1}\). This proves the first majorant and the band count.

For the distinct-value majorant, bound \(f_\theta(a)\) by the sum of each distinct gap \(g\) times \(\mathbf1\{a\ge\theta g/16\}\); each contributes less than \(32/\theta\). For an occupied band's minimum and maximum gaps \(u_b,v_b\), use instead \(v_b\mathbf1\{a\ge\theta u_b/16\}\). Its sum is less than \(32v_b/(\theta u_b)\le64/\theta\). These are finite sums of nonnegative terms. Equation (A26) follows from (A24), and \(N_1=32\) gives (8).

\pdfbookmark[0]{Appendix B. Subsets, Tolerance, and Heterogeneous Residuals}{r060.nav.5}
\section*{Appendix B. Subsets, Tolerance, and Heterogeneous Residuals}
Fix deterministic \(J\subseteq\mathcal I\). We retain (A1) on every arm for \(M\ge\mu^*\), but require cutoff concentration and charge selections only on \(J\). Both count-floor comparisons still start with a selection in \(J\); (A10) includes time consumed by outside arms. Thus (5) and its probability version hold for \(\Psi_J=\sum_{i\in J}\Delta_iN_i(T)\), restricting the sum and delay maximum to \(J\), with value zero for \(J=\varnothing\). For \(\eta\ge0\), choose \(J=J_\eta=\{i\in\mathcal I:\Delta_i>\eta\}\) and add \(\Psi_{\mathcal I\setminus J}\le\eta T\). The same-policy oracle permits an infimum over \((c,\eta)\) of these full-regret thresholds.

For (B1), take deterministic \(h_i<\infty\), \(q_i=\Pr(D_i>h_i)\), and \(\gamma_i=\Delta_i-q_i>0\) on nonempty \(J\). Put \(D_J=\max_{i\in J}h_i\). The weighted bound (A6), using \(r_i=b_i(h_i)\le q_i\) as in Appendix A.2, gives that inequality with this \(\gamma_i\). Let
\[
\begin{gathered}
V_J=\sum_{i\in J}\frac{\Delta_i}{\gamma_i^2},\qquad\\
\varrho_J=\max_{i\in J}\frac{\Delta_i}{\gamma_i}.
\end{gathered}
\]
Then \(0<\gamma_i\le\Delta_i\le1\) and \(V_J\ge\varrho_J\ge1\). Set \(A=96\) as in Appendix A. The cold cost is at most \((4608\ell+1)V_J\). All count-floor and density arguments depend on \(\gamma_i\) alone. The mature condition is \(h_i\gamma_i\ge4416\ell\) on the corresponding fixed-cutoff event. Thus (A17)--(A19) apply with gaps and extrema restricted to \(J\); in particular \(D_J\max_{i\in J}\gamma_i<4416\ell\) gives a cold-only charge on that event. This also proves the cutoff-only scale bound \(\Phi_T(D_J)=D_J\log^2(e\min\{D_J,T\})\) used below. High-bin gaps are at most \(16\varrho_Ja\), low-band gaps are less than \(2\varrho_J\varepsilon\), and the two small classes cost at most \(9\varrho_J\). Replacing \(1/\alpha\) in (A13)--(A20) by \(\varrho_J\), and using \(\varrho_J\le V_J\), yields with probability at least \(1-\delta\)
\[
\begin{gathered}
\Psi_J\le4618\ell V_J\\
+9264\varrho_J\Phi_T(D_J).
\end{gathered}
\]
At \(\delta=(T+1)^{-2}\), adding \(\delta T\le1\le LV_J\) gives
\[
\begin{gathered}
\mathbb E\Psi_J\le\min\{T,18473LV_J\\
+9264\varrho_J\Phi_T(D_J)\},
\end{gathered}
\]
Since the rounded unit-width coefficient \(C_0\) defined after (5) dominates both coefficients, this also gives
\begin{equation}
\begin{gathered}
\mathbb E\sum_{i\in J}\Delta_iN_i(T)\\
\le\min\{T,C_0[LV_J+\varrho_J\Phi_T(D_J)]\}.
\end{gathered}
\tag{B1}
\end{equation}
Taking \(\gamma_i=\alpha\Delta_i\) recovers the common-slack bound. When \(J\) varies, \(\Psi_{\mathcal I\setminus J}\) must be added first, as for \(J_\eta\) above.

\pdfbookmark[0]{Appendix C. Statistical Lower Bound}{r060.nav.6}
\section*{Appendix C. Statistical Lower Bound}
\subsection*{C.1 A family of joint laws}
Fix the parameters of Theorem 5. Write \(a=1/2\), \(p=c\Delta\), and \(\alpha=1-c\). We construct worlds \(P_0,P_2,\ldots,P_K\). Entries in the table are unconditional masses. The last column specifies both a missing mass and the reward carried by it.

\end{multicols}
{\small
\begin{longtable}{>{\raggedright\arraybackslash}p{73.10pt}>{\raggedright\arraybackslash}p{98.58pt}>{\raggedright\arraybackslash}p{98.58pt}>{\raggedright\arraybackslash}p{169.56pt}}

\toprule
\textbf{World and arm} & \textbf{Mass at \((Y,D)=(1,0)\)} & \textbf{Mass at \((Y,D)=(0,0)\)} & \textbf{Infinite-delay mass and its latent reward} \\
\midrule
\endfirsthead
\multicolumn{4}{@{}l}{\footnotesize\itshape Continued from previous page}\\
\toprule
\textbf{World and arm} & \textbf{Mass at \((Y,D)=(1,0)\)} & \textbf{Mass at \((Y,D)=(0,0)\)} & \textbf{Infinite-delay mass and its latent reward} \\
\midrule
\endhead
\midrule\multicolumn{4}{r@{}}{\footnotesize\itshape Continued on next page}\\
\endfoot
\bottomrule
\endlastfoot
\(P_0\), arm 1 & \(a\) & \(1-p-a\) & \(p\), reward 1 \\
\(P_0\), arm \(j\ge2\) & \(a-\alpha\Delta\) & \(1-p-a+\alpha\Delta\) & \(p\), reward 0 \\
\(P_i\), arm 1 & \(a\) & \(1-p-a\) & \(p\), reward 0 \\
\(P_i\), arm \(i\) & \(a+\alpha\Delta\) & \(1-p-a-\alpha\Delta\) & \(p\), reward 1 \\
\(P_i\), arm \(j\notin\{1,i\}\) & \(a-\alpha\Delta\) & \(1-p-a+\alpha\Delta\) & \(p\), reward 0 \\
\end{longtable}
}
\begin{multicols}{2}
The masses are legal: \(p\le\Delta\le1/8\), immediate-one masses lie in \([3/8,5/8]\), and immediate-zero masses are at least \(1/2-p-\alpha\Delta=1/2-\Delta\ge3/8\). Independent arm components and rounds give a permitted model.

In \(P_0\), the means are \(\mu_1=a+p\) and \(\mu_j=a-\alpha\Delta\), giving gap \(\Delta\) for every bad arm. In \(P_i\), arm \(i\) has mean \(a+\Delta\), arm 1 has mean \(a\), and the remaining arms have mean \(a-\alpha\Delta\). Its bad gaps are therefore \(\Delta\) and \((1+\alpha)\Delta\). Every arm has missing probability \(p\), which is at most \(c\) times its actual bad gap. All required ordinary cutoffs equal zero. For every bad arm its missing rewards are zero, so its weighted bias is also \(p\), and \(W_c=0\).

\subsection*{C.2 Visible information and the chain rule}
The visible outcome of a pull is an immediate one, an immediate zero, or no report. Changing the latent reward of a permanently missing pull does not change this outcome. Between \(P_0\) and \(P_i\), only the observable law of arm \(i\) changes. Let \(k\) be the relative entropy of its three-symbol law under \(P_0\) with respect to that under \(P_i\). The common missing mass contributes zero. From \(\log x\le x-1\),
\begin{equation}
\begin{aligned}
k&\le\sum_z\frac{(P(z)-Q(z))^2}{Q(z)}\\
&=4\alpha^2\Delta^2
\left[\frac1{a+\alpha\Delta}
+\frac1{1-p-a-\alpha\Delta}\right]
\le32\alpha^2\Delta^2.
\end{aligned}
\tag{C1}
\end{equation}
Also \(k>0\). For any same policy, its action kernel after a common visible history is identical across worlds. The conditional relative entropy contribution of the next feedback is \(k\) if arm \(i\) is selected and zero otherwise. The likelihood-ratio chain rule for the full visible transcript therefore gives
\begin{equation}
\operatorname{KL}(P_0^T,P_i^T)=k\,\mathbb E_0N_i(T).
\tag{C2}
\end{equation}
The transcript excludes unavailable latent missing rewards. Reports carry no originating timestamps here; adding them reveals no further rewards.

\subsection*{C.3 Testing and finite-time inversion}
Draw an independent uniform round \(U\in\{1,\ldots,T\}\), let \(n_i=\mathbb E_0N_i(T)\), and put \(u=\max_jR_T(P_j)/\Delta\). Then \(\sum_{i\ge2}n_i\le u\), and \(\mathbb E_i[T-N_i(T)]\le u\) because every nonoptimal gap in \(P_i\) is at least \(\Delta\). The event \(A_U\ne i\) has probability \(1-n_i/T\) in \(P_0\) and at most \(u/T\) in \(P_i\).

We use a standard finite-time binary-KL reduction (Garivier et al., 2019). Grouping likelihood terms by this event and its complement, the log-sum inequality bounds its binary relative entropy by (C2). For fixed \(x\), \(y\mapsto\operatorname{kl}(x,y)\) is decreasing on \((0,x]\), since its derivative is \((y-x)/(y(1-y))\le0\). For probabilities \(x,y\), binary entropy is at most \(\log2\), so
\begin{equation}
\operatorname{kl}(x,y)\ge x\log(1/y)-\log2.
\tag{C3}
\end{equation}
If \(0<u\le T/4\), then \(n_i\le u\) and
\[
\begin{gathered}
kn_i\ge\frac34\log(T/u)-\log2\\
\ge\frac14\log(T/u).
\end{gathered}
\]
Summing over alternatives gives \(ku\ge(K-1)\log(T/u)/4\). The case \(u=0\) is impossible by the same binary inequality and finite visible relative entropy. If \(u>T/4\), no testing step is needed.

Set \(x=Tk/(K-1)>0\), \(v=ku/(K-1)\). In the large-\(u\) case, \(v>x/4\ge\log(1+x)/8\). In the other case, \(4v\ge\log(x/v)\), hence \(ve^{4v}\ge x\). The elementary inequality
\[
\log(1+x)\le x/\sqrt{1+x},\qquad x\ge0,
\]
follows by differentiating their difference: its derivative is \([1+x/2-\sqrt{1+x}]/(1+x)^{3/2}\ge0\). If \(v<\log(1+x)/8\), this inequality would give \(ve^{4v}<x/8<x\), a contradiction. Thus in both cases
\begin{equation}
u\ge\frac{K-1}{8k}\log\left(1+\frac{Tk}{K-1}\right).
\tag{C4}
\end{equation}
The function \(k\mapsto\log(1+Tk/(K-1))/k\) decreases for \(k>0\), since \(\log(1+z)\ge z/(1+z)\). Substitute (C1) into (C4), and weaken \(\log(1+32z)\ge\log(1+z)\). Multiplication by \(\Delta\) proves (14).

The infinite-delay family is fixed across \(T\). At any fixed horizon, replacing infinity by \(T+1\) preserves visible transcripts and quantile membership, giving the same finite-support lower bound. This horizon-dependent replacement gives no logarithmic asymptotic lower bound on one fixed finite-support instance.

\pdfbookmark[0]{Appendix D. Critical and Shared-Delay Lower Bounds}{r060.nav.7}
\section*{Appendix D. Critical and Shared-Delay Lower Bounds}
\subsection*{D.1 Exact nonidentification at the standard critical quantile}
Fix \(K\ge2\) and \(0<\Delta\le1/8\). In world \(j\), every arm has immediate-one mass \(a=1/2\), immediate-zero mass \(1-a-\Delta\), and infinite-delay mass \(\Delta\). The missing reward equals one for arm \(j\) and zero for every other arm. Thus \(j\) uniquely maximizes the mean, every bad gap equals \(\Delta\), and every bad arm has \(F_i(0)=1-\Delta\), so \(d_i(1-\Delta)=0\) under (4).

All observable laws, for every policy and horizon, agree across worlds. Under their common action law, average optimal pulls over the \(K\) worlds equal \(T/K\). Averaging regrets gives
\begin{equation}
\max_jR_T(P_j)\ge(1-1/K)T\Delta.
\tag{D1}
\end{equation}
These fixed worlds use infinite delays; at fixed \(T\), replacement by \(T+1\) preserves (D1) through that horizon. Equality \(F_i(0)=1-\Delta\) fails the strictly-greater convention, treated next.

\subsection*{D.2 Strict cumulative-distribution margin and finite support}
We use the two-arm table in Appendix C with auxiliary slack \(\eta=1/(2\sqrt T)\), true gap \(\Delta=1/8\), missing mass \(p=(1-\eta)\Delta\), and missing delay replaced by \(T+1\). Both worlds have gap \(\Delta\), finite-support delays, and \(F_{\rm bad}(0)=1-\Delta+\eta\Delta>1-\Delta\). In the positive-slack notation this is \(c'=1-\eta\) and \(\alpha'=\eta\), with \(\alpha'\) depending on \(T\). By (C1)--(C2),
\begin{equation}
\begin{gathered}
\operatorname{KL}(P^T,Q^T)\\
\le32\eta^2\Delta^2T=1/8.
\end{gathered}
\tag{D2}
\end{equation}
Binary relative entropy has second derivative \(1/[x(1-x)]\ge4\) in its first argument, with value and first derivative zero at equality. Thus \(\operatorname{kl}(x,y)\ge2(x-y)^2\); log-sum gives \(|P(E)-Q(E)|\le\sqrt{\operatorname{KL}(P,Q)/2}\) for every event \(E\), also after adding an independent uniform round.

Let \(n=\mathbb E_PN_2(T)\), \(m=\mathbb E_QN_1(T)\), and take the event \(A_U=1\). Its probabilities are \(1-n/T\) and \(m/T\). Equation (D2) implies \(1-(n+m)/T\le1/4\), so \(n+m\ge3T/4\). Therefore
\[
\begin{gathered}
\max\{R_T(P),R_T(Q)\}\\
\ge\frac{\Delta(n+m)}2\ge3T/64\ge T/32.
\end{gathered}
\]
This proves (15) under either standard or strictly-greater critical quantiles.

\subsection*{D.3 One shared waiting interval}
\textbf{Proposition D.1 (shared blind prefix).} For \(K\ge2\), \(T\ge1\), integer \(d\ge0\), \(c\in[0,1]\), and \(0<\Delta\le1/8\), there are \(K\) instances with \(D_c=d\) such that every policy satisfies
\begin{equation}
\max_jR_T(P_j)\ge\Delta(1-1/K)\min\{T,d+1\}.
\tag{D3}
\end{equation}
The blind-prefix argument follows the idea of Lancewicki et al.~(2021, Thm. 4), whose delay distribution is shared by all arms; the construction below instead permits the optimal arm to never report.

For the shared-prefix construction, in world \(j\) set \(Y_j=\Delta,D_j=\infty\) deterministically. Every other arm has \(Y_i=0,D_i=d\). The best gap is \(\Delta\). Since \(1-c\Delta>0\) for the specified parameters, every bad quantile is exactly \(d\). Before and including the decision in round \(d+1\), no selected pull can have reported: the earliest possible report is at that round's end. Thus the first \(q=\min(T,d+1)\) actions have a common law across all worlds. Averaging their bad pulls gives \((1-1/K)q\), and subsequent regret is nonnegative. Multiplying by \(\Delta\) proves (D3). For fixed \(T\), replacing the optimal arm's infinite delay by \(T+1\) changes nothing. The separate statistical (Theorem 5) and shared-delay (Proposition D.1) constructions give no additive lower bound on one instance.

\pdfbookmark[0]{Appendix E. A Delay-Slack Lower Bound for the Exact BHO Rule}{r060.nav.8}
\section*{Appendix E. A Delay-Slack Lower Bound for the Exact BHO Rule}
\textbf{Proposition E.1 (BHO-specific delay limitation).} Fix \(\alpha\in(0,1]\), \(0<\Delta\le1/32\), and a positive integer \(d\). Set
\[
\begin{gathered}
T=\left\lfloor\frac{d}{16\alpha\Delta}\right\rfloor,\\
\quad \ell=\log(8T(T+1)^2),\\
\quad b=\left\lfloor\frac{2\ell}{9\alpha^2\Delta^2}\right\rfloor+1.
\end{gathered}
\]
If \(d\ge2(b+1)\), a legal two-arm instance with an immediate deterministic optimal arm and \(D_c=d\), \(c=1-\alpha\), satisfies
\begin{equation}
\begin{gathered}
\Pr(\Psi_T\ge d/(64\alpha))\ge1-(T+1)^{-2}/8,\\
\qquad R_T\ge d/(128\alpha)
\end{gathered}
\tag{E1}
\end{equation}
for the exact BHO rule.

Let \(a=1/2\), \(q=c\Delta\), \(c=1-\alpha\). Arm 1 has deterministic reward \(a\) and zero delay. Arm 2 has joint law

\begin{center}\small
\begin{tabular}{>{\raggedright\arraybackslash}p{61.78pt}>{\raggedright\arraybackslash}p{44.84pt}>{\raggedright\arraybackslash}p{92.25pt}}
\toprule
\textbf{Reward} & \textbf{Delay} & \textbf{Probability} \\
\midrule
1 & \(d\) & \(a-\Delta\) \\
0 & \(d\) & \(1-a+\alpha\Delta\) \\
0 & \(\infty\) & \(c\Delta\) \\
\bottomrule
\end{tabular}
\end{center}
These masses are nonnegative and sum to one. The means are \(a\) and \(a-\Delta\). The bad-arm cumulative distribution is zero before \(d\) and equals \(1-c\Delta\) at \(d\); hence \(D_c=d\), including when \(c=0\). For \(c'<c\) the cutoff is infinite, whereas for \(c\le c'<1\) it equals \(d\). The numerical oracle therefore cannot choose a larger positive slack to evade the example.

Let \(V_k\) indicate that the \(k\)-th selected arm-2 outcome is a finite-delay zero. Its mean is \(z=1-a+\alpha\Delta\). Hoeffding and a union bound give an event \(E\) with
\begin{equation}
\begin{gathered}
\Pr(E)\ge1-Te^{-\ell}=1-\delta/8,\\
\qquad\\
\sum_{k\le m}V_k\le zm+\sqrt{m\ell/2}\\
\quad(1\le m\le T).
\end{gathered}
\tag{E2}
\end{equation}
At completed time \(t\), set \(m=m_2(t)>0\), \(m_{\rm old}=m_2((t-d)_+)\). Exactly the finite-zero outcomes in the old prefix have returned as zero reports. Observed ones cancel from \(G_2+m_2-n_2\); unreported zeros remain completed as one. Therefore
\[
u_2^+(t)=1-\frac{\sum_{k\le m_{\rm old}}V_k}{m}.
\]
On \(E\), the confidence term absorbs the prefix fluctuation because \(m_{\rm old}\le m\). Consequently
\begin{equation}
\begin{gathered}
\rho_2(t)\ge1-zm_{\rm old}/m\\
=a-\alpha\Delta+z\frac{B_2(t,d)}m.
\end{gathered}
\tag{E3}
\end{equation}
Let \(h_0=a+3\alpha\Delta<1\). Suppose the best arm's count has never exceeded \(b\) through time \(t\). At any earlier time \(s<d\), no arm-2 reward has reported, so \(\rho_2(s)\ge1\ge h_0\) whenever its count is positive. For \(s\ge d\), the last \(d\) rounds contain at least \(d-m_1(s)\ge d-b\ge d/2\) arm-2 pulls. Also \(m_2(s)\le T\le d/(16\alpha\Delta)\) and \(z\ge1/2\). Equation (E3) then gives
\begin{equation}
\begin{gathered}
\rho_2(s)\ge a-\alpha\Delta+\\
\frac12\frac{d/2}{T}\ge a+3\alpha\Delta=h_0.
\end{gathered}
\tag{E4}
\end{equation}
Zero counts initialize \(H_2=1\). Every historical score is at least \(h_0\), hence \(H_2(s)\ge h_0\).

For the immediate deterministic arm 1, at positive count \(n\), \(H_1=\min\{1,a+\sqrt{\ell/(2n)}\}\). The definition \(b=\lfloor2\ell/(9\alpha^2\Delta^2)\rfloor+1\) implies \(\beta_1<(3/2)\alpha\Delta\) at count \(b\). Thus
\[
H_1+\beta_1\le a+2\beta_1<h_0\le M.
\]
Thus the optimum is ineligible for its \((b+1)\)-st pull. Induction from zero counts gives \(m_1(t)\le b\) for all \(t\le T\) on \(E\). This controls the full historical minimum; neither ties nor minimum-count selection can choose an ineligible arm.

Since \(\Delta\le1/32\), \(T\ge2d-1\). Together with \(d\ge2(b+1)\), this gives \(T\ge2b\), and \(\Psi_T=\Delta(T-m_1(T))\ge\Delta T/2\). The size condition implies \(d\ge4\); the floor defining \(T\) also gives \(T\ge d/(32\alpha\Delta)\). Hence \(\Psi_T\ge d/(64\alpha)\) on \(E\). Since \(\Pr(E)\ge1/2\), (E1) follows.

For fixed \(\alpha,\Delta\), \(\ell=O(\log d)\), so the size condition holds for all sufficiently large \(d\). To separate the delay cost from the statistical term, fix \(\Delta=1/32\), let \(d=\lceil\alpha^{-3}\rceil\), and send \(\alpha\) to zero. Then \(T=\Theta(\alpha^{-4})\), \(\ell=\Theta(\log(1/\alpha))\), and the size condition holds eventually. Moreover, \(S_T/\alpha^2=\Theta(\log(1/\alpha)/\alpha^2)=o(d/\alpha)\). For any fixed \(p\ge0\) and \(\beta<1\), the same family has \(d\log^p(eKT)/\alpha^\beta=o(d/\alpha)\). Thus no universal bound of that smaller inverse-slack delay power, plus the stated statistical term, can hold for this exact BHO rule.

\pdfbookmark[0]{Appendix F. Decision Learning and Value Nonidentification}{r060.nav.9}
\section*{Appendix F. Decision Learning and Value Nonidentification}
Let \(b_i(\infty)=\mathbb E[(1-Y_i)\mathbf1\{D_i=\infty\}]\). If the bad-arm set is nonempty and \(b_i(\infty)<\Delta_i\) for each bad arm, choose a fixed \(c\) strictly between \(\max_{i\in\mathcal I}b_i(\infty)/\Delta_i\) and \(1\). Bounded convergence gives \(b_i(h)\downarrow b_i(\infty)\), so every \(w_i(c)\) is finite. Theorem 2 gives \(O(\log(eT))\) regret on this fixed instance without a delay moment condition. If there are no bad arms, regret is zero.

Let arm 1 never report. Its reward is Bernoulli\((3/4)\) in one instance and deterministically one in another. Arm 2 reports immediate Bernoulli\((1/2)\) rewards in both. Common policy randomness and arm-2 rewards couple the entire observed transcript across the two instances. For any estimator based on that transcript and any \(0<\varepsilon<1/8\), the two intervals of radius \(\varepsilon\) around the optimal means are disjoint. The two estimation failure probabilities therefore sum to at least one at each horizon. In both instances arm 2 is the only bad arm and its delay is zero, so Theorem 1 gives logarithmic regret.

\pdfbookmark[0]{Appendix G. Completed-Order Reduction}{r060.nav.10}
\section*{Appendix G. Completed-Order Reduction}
\subsection*{G.1 Setup}
We use the model and horizon-aware BHO family of Section 2. Write
\[
\begin{gathered}
\mu^*=\max_i\mu_i,\quad \Delta_i=\mu^*-\mu_i,\qquad\\
\nu_i=\mathbb E[Y_i1_{\{D_i<\infty\}}]+\Pr(D_i=\infty).
\end{gathered}
\]
Let \(\nu^*=\max_i\nu_i\), \(V=\arg\max_i\nu_i\), \(J=\{i:\nu_i<\nu^*\}\), and \(\Gamma_i=\nu^*-\nu_i\) for \(i\in J\).

\subsection*{G.2 Exact virtual environment (Lemma G.1)}
For each latent outcome define
\[
\begin{gathered}
X_i=Y_i1_{\{D_i<\infty\}}+1_{\{D_i=\infty\}},\\
\bar D_i=\begin{cases}D_i,&D_i<\infty,\\0,&D_i=\infty.\end{cases}
\end{gathered}
\]
The transformed law is still bounded, stationary and iid in time, and \(\mathbb E X_i=\nu_i\). Every transformed delay is finite almost surely; its support and mean may remain unbounded.

We couple both processes with common potential outcomes, private randomness, tie rules, \(K,T\), and confidence widths. Finite-delay outcomes report identically; permanently missing outcomes produce only an extra virtual \((i,1)\) report at the end of the round that selected arm \(i\).

For any pulled arm the optimistic-completion numerator is
\[
G_i+m_i-n_i=m_i-\sum_{\text{reports from }i}(1-Y).
\]
Extra unit reports add zero deficit. Equal past actions therefore imply equal pull counts, deficits, raw scores, historical indices \(H_i\), eligibility and the next lexicographic choice. The report counts \(n_i\), reward sums \(G_i\), and pending counts can differ. Updating the historical indices of all pulled arms at each round end preserves the induction. Origins, infinity indicators and virtual reports remain unavailable to the learner.

The identity and induction use exact real arithmetic. For binary rewards the deficits are integers, which double precision represents exactly up to \(2^{53}\).

\subsection*{G.3 Completed-order consistency (Corollary G.2)}
Assume every completed-mean maximizer is a true optimum:
\begin{equation}
V\subseteq\arg\max_i\mu_i.
\tag{CO}
\end{equation}
If there is a true suboptimal arm, it lies in \(J\) and
\[
\kappa=\max_{i:\Delta_i>0}\frac{\Delta_i}{\Gamma_i}<\infty.
\]
If all arms are truly optimal, regret is zero and no ratio is needed.

For every action sequence,
\[
\begin{gathered}
\sum_{t=1}^T\Delta_{A_t}\le\\
\kappa\sum_{t=1}^T(\nu^*-\nu_{A_t}).
\end{gathered}
\]
For true-optimal arms the left summand is zero; for true-suboptimal arms this is the definition of \(\kappa\). We apply the existing Theorem 1 to the legal virtual environment and use Lemma G.1. With the rounded unit-width coefficient \(C=C_0\) defined after (5), \(L=\log(eKT)\), \(0\le c<1\), and
\[
\bar D_c=\max_{i\in J}\bar d_i(1-c\Gamma_i),
\]
the expected true regret satisfies
\[
\begin{gathered}
R_T^\mu\le\\
\min\!\left\{T,\; \kappa C\left[ \frac{L}{(1-c)^2}\sum_{i\in J}\frac1{\Gamma_i} +\frac{\Phi_T(\bar D_c)}{1-c}\right]\right\}.
\end{gathered}
\]
The same domination transports the precise probability bound with its existing event probability.

Alternatively, we apply the weighted theorem in the virtual problem. Its bias is exactly
\[
\begin{gathered}
\bar b_i(h)=\mathbb E[(1-X_i)1_{\{\bar D_i>h\}}]\\
=\mathbb E[(1-Y_i)1_{\{h<D_i<\infty\}}],
\end{gathered}
\]
so permanently missing deficits have already been included in \(\nu_i\) and only the finite-report tail remains. The original theorem's weighted cutoff is then evaluated at \(c\Gamma_i\).

For fixed finite-\(K\) instances satisfying (CO), fix \(c\) strictly between zero and one. All required virtual quantiles are finite since \(\Pr(\bar D_i<\infty)=1\) and the index set is finite. Thus \(R_T^\mu/T\to0\) for horizon-aware BHO: pointwise expected consistency.

If several latent-optimal arms have different \(\nu_i\) values, \(J\) may contain latent-optimal arms. The displayed corollary can then include their virtual cutoffs. It must not replace the original theorem's stronger bad-only parameter statement. For a unique true optimum satisfying (CO), \(J\) is exactly the true bad-arm set.

\subsection*{G.4 Strict enlargement of the sufficient class}
There are two arms. Arm 1 has Bernoulli(3/4) rewards and never reports. Arm 2 has atoms
\[
\begin{gathered}
(Y,D)=(0,\infty)\text{ with mass }2/5,\quad\\
(1,0)\text{ with mass }1/2,\quad\\
(0,0)\text{ with mass }1/10.
\end{gathered}
\]
Then \(\mu_1=3/4,\mu_2=1/2,\Delta_2=1/4\), but \(\nu_1=1,\nu_2=9/10,\Gamma_2=1/10\) and \(\kappa=5/2\). For every finite \(h\), \(b_2(h)=2/5>c\Delta_2\) for all \(c<1\), so every original weighted cutoff is infinite. Its finite-report mass \(3/5\) is also below \(1-c\Delta_2>3/4\), so the original ordinary cutoff is infinite.

Both virtual delays are zero: arm 1 always rewards \(1\), and arm 2 is immediate Bernoulli(9/10). At \(c=0\) the corollary gives logarithmic true regret, strictly enlarging BHO's sufficient class.

\subsection*{G.5 Wrong-order obstruction (Proposition G.3)}
Suppose every arm in \(V\) is truly suboptimal, and let
\[
\begin{gathered}
\eta=\min_{i\in V}\Delta_i>0,\qquad\\
\Gamma_{\min}=\min_{i\in J}\Gamma_i>0.
\end{gathered}
\]
\(J\) is nonempty, since a true-optimal arm exists outside \(V\). The virtual theorem for any fixed \(c\in(0,1)\) gives \(R_T^\nu=o(T)\), hence
\[
\begin{gathered}
\mathbb E\sum_{t=1}^T1_{\{A_t\notin V\}}\\
\le R_T^\nu/\Gamma_{\min}=o(T).
\end{gathered}
\]
Therefore \(R_T^\mu\ge\eta[T-o(T)]\). This is a failure of the BHO family.

For example, arm 1 is immediate Bernoulli(3/4). Arm 2 has mass \(1/2\) at \((1,0)\) and mass \(1/2\) at \((0,\infty)\). Its true mean is \(1/2\) but its completed mean is \(1\); arm 1 has completed mean \(3/4\). Virtual delays are zero, and the unique virtual winner is truly worse. Thus \(R_T^\mu=T/4-O(\log T)\), while trivially \(R_T^\mu\le T/4\), so \(R_T^\mu/T\to1/4\).

Completion is not uniformly safer than averaging returned rewards. If arm \(i\)'s never-reported outcomes are zeros of total probability \(h\) and the optimum reports every outcome, completed limits rank the two arms correctly iff \(h<\Delta_i\). With growing returned-sample counts, the returned averages converge to \(\mu_i/(1-h)\) and \(\mu^*\), so their limits rank the arms correctly iff \(h<\Delta_i/\mu^*\), a weaker requirement when \(\mu^*<1\). In the example above, \(h=1/2\) exceeds both \(\Delta_2=1/4\) and \(\Delta_2/\mu^*=1/3\). Completion's advantage lies with hidden successes: they never lower a completed index, whereas they bias returned averages downward, the optimum's included.

Mixed truly optimal/suboptimal \(V\) is covered by neither proposition nor (CO). General mixed ties remain open.

\pdfbookmark[0]{Appendix H. Comparison with Lancewicki et al.~(2021)}{r060.nav.11}
\section*{Appendix H. Comparison with Lancewicki et al.~(2021)}
L21 uses the same arm-labeled reports without originating rounds and allows infinite delays (Protocol 1). Its SE and PSE guarantees (Theorems 2--3) assume reward-independent delays. Its SE bound in Eq. (1) takes a gap-weighted maximum of sums of optimal and suboptimal cutoffs. PSE avoids the optimal cutoff through cumulative report counts and within-phase pausing. Its delay term is \((\log T)(\log K)\max_{i\in\mathcal I}\Delta_i d_i(q_i)\), with free quantile levels \(q_i\).

OPSE (L21, Theorem 5) allows reward dependence and pays \(4\log K(D+D_*)\) in the notation of Table 1. At the same quantile levels, BHO's bound uses the suboptimal maximum alone, with a squared logarithm in the delay term. If the optimal arm reports immediately, \(D_*=0\). For fixed gaps, \(K\) and finite suboptimal cutoffs, OPSE then has the smaller displayed upper bound for sufficiently large \(T\), because its statistical coefficient is smaller. When the matched optimal cutoff is infinite, OPSE's displayed bound clipped at \(T\) is \(T\).

The Optimistic-UCB expression in L21, Eq. (8), uses a sum of suboptimal cutoffs and is stated without an expanded proof. The paragraph does not specify clipping, radius or ties. Appendix O.1 defines the experimental OUCB-current baseline.

\pdfbookmark[0]{Appendix I. Serial Waiting for the Exact Greedy Rule}{r060.nav.12}
\section*{Appendix I. Serial Waiting for the Exact Greedy Rule}
\subsection*{I.1 Exact rule and statement}
Define the greedy historical comparator G in exact real arithmetic. At completed time \(t\), it selects the lexicographically smallest \((-H_i(t),m_i(t),i)\) over all arms, initializes \(H_i(0)=1\), and updates \(H_i(t)=\min\{H_i(t-1),1,\rho_i(t)\}\), where \(\rho_i=u_i^++\beta_i\), \(\beta_i=\sqrt{\ell/(2m_i)}\), and
\[
\ell=\log(4KT(T+1)^2).
\]
\textbf{Proposition I.1 (full statement).} For every integer \(K\ge65536\) there is a legal \(K\)-arm instance and horizon \(T=K^3\), with one optimal arm of deterministic reward \(3/4\), all other rewards deterministic and in \((49/100,1/2)\), and the same deterministic finite delay
\[
n=\lceil 2\ell\rceil,\qquad D=Kn
\]
on every arm, such that
\begin{equation}
\Psi_T=R_T\ge \frac{(K-1)D}{4}.
\tag{I1}
\end{equation}
In particular, no universal finite \(C\) makes exact G satisfy, on every legal instance and horizon,
\begin{equation}
\begin{gathered}
R_T\le C\left[L\sum_{i:\Delta_i>0}\Delta_i^{-1}+D_0L^2\right],\\
\qquad L=\log(eKT),\quad D_0=\max_{i:\Delta_i>0}d_i(1).
\end{gathered}
\tag{I2}
\end{equation}
On the same instances, Theorem 1 bounds BHO with \(c=0\), \(\alpha=1\), \(D_0=D\), separating worst-case orders without a universal instance ranking.

Deterministic laws satisfy independent fresh-vector sampling. Reports reveal arm and reward exactly \(D\) rounds after selection, after that round's action. G uses label ties without latent timing information.

\subsection*{I.2 A common first crossing cycle}
Take \(K\) large enough that \(\ell\ge10\); all sufficiently large \(K\) below meet this. Let \(n=\lceil2\ell\rceil\). Under a still-uninterrupted round-robin schedule, an arm of deterministic reward \(x\) that has just received its \(m\)-th pull also receives the report of pull \(m-n\) when \(m>n\), because \(D=Kn\). Its current score is
\begin{equation}
f_x(m)=x+(1-x)\frac nm+\sqrt{\frac\ell{2m}},\qquad m>n.
\tag{I3}
\end{equation}
For \(m\le n\) there are no reports and its score is \(1+\beta_m>1\). For \(m>n\), \(f_x(m)\) is strictly decreasing in \(m\). For every \(x\in[0.49,0.50]\), at integer \(m\le5\ell\) with \(m>n\),
\begin{equation}
f_x(m)\ge .49+.51\,(2/5)+1/\sqrt{10}>1.
\tag{I4}
\end{equation}
At \(m=\lceil6\ell\rceil\), using \(n\le2\ell+1\),
\begin{equation}
\begin{gathered}
f_x(m)\le .50+.50\,\frac{2\ell+1}{6\ell}+1/\sqrt{12}\\
\le .675+1/\sqrt{12}<1.
\end{gathered}
\tag{I5}
\end{equation}
For each of the finitely many integers \(n<m\le\lceil6\ell\rceil\), the equation \(f_x(m)=1\) excludes at most one \(x\), because the coefficient \(1-n/m\) of \(x\) is positive. Choose \(x_0\in(0.49,0.50)\) outside that finite set. Let \(N\) be its first integer crossing, so \(f_{x_0}(N)<1<f_{x_0}(N-1)\). By continuity there is a nonempty open interval \(J\subset(0.49,0.50)\) on which both strict inequalities hold. Choose \(K-1\) distinct rewards \(\mu_j\in J\). Strict decrease in \(m\) implies that all these arms first cross below 1 on their \(N\)-th pull. Moreover,
\begin{equation}
5\ell<N\le\lceil6\ell\rceil\le6\ell+1.
\tag{I6}
\end{equation}
Set the last remaining arm's reward to \(\mu_*=3/4\). Its gap from each bad arm is greater than \(1/4\) and less than \(0.26\).

\subsection*{I.3 Reaching the state from initialization}
While all \(H_i=1\), count/label ties force ascending-label round-robin. I.2 keeps bad scores at least one before cycle \(N\); through count \(N\), the optimum has
\begin{equation}
\rho_*\ge\mu_*+\beta_N>3/4+1/4=1,
\tag{I7}
\end{equation}
because \(N\le6\ell+1<8\ell\) for \(\ell\ge10\). Thus its \(H\) stays 1 too. The claimed fixed order is consequently justified inductively through the first \(N-1\) cycles.

In cycle \(N\), each bad arm's own pull and its old report give exactly (I3), so its \(H\) first becomes \(f_{\mu_j}(N)<1\) immediately after that round. The rest of cycle \(N\) still follows the same label order: every not-yet-acted arm has count \(N-1\) and \(H\)=1, every already-acted optimal arm has count \(N\), and every already-acted bad arm has \(H\)\textless1. Hence dropping a bad arm from the \(H\)=1 tie cannot make an already-acted arm precede an unacted one. At
\[
t_* = KN
\]
all bad arms have count \(N\) and historical index below one, while the optimal arm has count \(N\) and historical index one.

For the next \(D\) rounds G selects only the optimal arm. To verify this without circularity, suppose \(s\) of these new optimal pulls have been made, \(0\le s\le D\). None of these new reports has arrived by completed time \(t_*+s\). At most its \(N\) initial pulls can have reported. With \(m=N+s\),
\begin{equation}
\rho_*\ge1-\frac{N}{4m}+\beta_N\sqrt{\frac Nm}\ge1,
\tag{I8}
\end{equation}
since \(\beta_N>1/4\) and \(\sqrt{N/m}\ge N/m\). Thus \(H_*=1\) throughout the induction and every bad \(H\) remains below 1. At
\[
t_1=t_*+D
\]
all \(N\) initial reports from every bad arm have returned. Its count is still \(N\) and its report sum is \(N\mu_j\). Since its count is fixed while those reports arrive, its score decreases to its final value; the historical minimum is exactly
\begin{equation}
h_j=\mu_j+\beta_N.
\tag{I9}
\end{equation}
These \(h_j\) are distinct, and, using (I6) and \(\ell\ge10\),
\begin{equation}
\begin{gathered}
.77 < .49+\frac1{\sqrt{12.2}}\le h_j\\
<.50+\frac1{\sqrt{10}}<1.
\end{gathered}
\tag{I10}
\end{equation}
The optimal count is now \(N+D\). An as-yet-unrestarted bad arm receives no additional reports, so its count \(N\) and its historical index \(h_j\) remain unchanged until it is selected again.

\subsection*{I.4 A lemma on unchanged historical indices}
\textbf{Lemma I.2 (pending-pull score identity).} Suppose an arm has deterministic reward \(x\), \(N\) old pulls all reported, no other old pulls, and historical index \(h=x+\beta_N\). If it makes \(q\) new pulls whose reports are all still pending, then, with \(z=\sqrt{N/(N+q)}\),
\begin{equation}
\begin{gathered}
\rho(q)-h\\
=(1-z)\left[(1-x)(1+z)-\beta_N\right].
\end{gathered}
\tag{I11}
\end{equation}
Thus if \(\beta_N\le1-x\), the current score is at least \(h\) for every such \(q\) and the historical index stays exactly \(h\).

Proof: \(u^+=1-(1-x)N/(N+q)\), \(\beta=\beta_Nz\). We subtract \(x+\beta_N\) to obtain \((1-x)(1-z^2)+\beta_N(z-1)\), factoring as (I11) with the claimed signs.

In the construction \(\beta_N<1/\sqrt{10}<0.317<1-\mu_j\), so the lemma applies to every bad arm.

\subsection*{I.5 A separate delay block for each restart}
We order bad arms by the time of their first selection after \(t_1\). Suppose j is the next bad arm to be restarted, and its first new selected round is \(s\). Its historical index is \(h_j\) and it has count \(N\). Since G chooses it, all other current \(H\) are at most \(h_j\). Every other not-yet-restarted bad arm has a distinct frozen \(h\) and count \(N\), so its \(H\) is strictly below \(h_j\). Every already-restarted bad arm has at least \(N+D\) pulls by induction; the optimal arm has at least \(N+D\) pulls by Section I.3.

Until round \(s+D\) none of j's new reports can arrive. For the \(D\) decisions in rounds \(s,\ldots,s+D-1\), (I11) keeps its \(H\) exactly \(h_j\). Every other \(H\) is nonincreasing. At the decision with \(q\) preceding new pulls, \(0\le q\le D-1\), its count is \(N+q<N+D\), so it wins every \(H\) tie against the optimal arm and any previously restarted bad arm. It strictly outranks every other not-yet-restarted bad arm. Hence G selects j on all \(D\) of these rounds. This proves the induction.

\subsection*{I.6 Completing every block within the horizon}
It remains to show that G cannot remain forever with some bad arm never restarted. At any time t and positive count m, deterministic rewards and the common deterministic delay give
\begin{equation}
\begin{gathered}
u_i^+(t)=\mu_i+(1-\mu_i)\frac{\operatorname{pending}_i(t)}m\\
\le\mu_i+\frac Dm,
\end{gathered}
\tag{I12}
\end{equation}
since at most \(D\) selected pulls can be pending after a completed round. If \(m\ge100D\) and \(K\ge100\), then \(D=Kn\ge2K\ell\) implies
\begin{equation}
\begin{gathered}
\rho_i(t)\le .75+.01+\sqrt{\ell/(200D)}\\
\le .76+1/\sqrt{400K}\le .765<.77.
\end{gathered}
\tag{I13}
\end{equation}
While an unrestarted bad arm remains after \(t_1\), \(M\ge h_j>0.77\), so a selected arm's pre-action count is below \(100D\). No count can exceed \(100D\); summing counts bounds completed rounds by \(100KD\), including intervening optimal pulls. Hence every bad arm restarts by round \(100KD+1\).

The last first restart begins by round \(100KD+1\), so its \(D\)-action block ends by \(100KD+D\). For sufficiently large \(K\) this is at most \(T=K^3\). Explicitly,
\[
\begin{gathered}
\ell\le\log(16K^{10}),\qquad\\
n\le 7+20\log K,
\end{gathered}
\]
and any \(K\) satisfying \(K\ge102(7+20\log K)\) gives \(T\ge102KD>100KD+D\); this holds for every \(K\ge65536\): at that point \(\log K<11.2\), so \(102(7+20\log K)<23562<K\), and the difference has derivative \(1-2040/K>0\) thereafter. The time \(t_1\) is also below this horizon, since \(t_1=K(N+n)\), \(N\le6\ell+1\), and \(n\ge2\ell\) make it much smaller than \(100KD\). Thus Section I.5 contributes \(D\) bad decisions for each of the \(K-1\) arms within the prescribed horizon. Their gaps exceed \(1/4\), proving (I1).

\subsection*{I.7 Separation of regret orders}
All suboptimal ordinary cutoffs satisfy \(d_i(1)=D\), so \(D_0=D\); there is no infinite tail and no reward dependence of delay. The gap bounds imply
\[
L\sum_{i\ne *}\Delta_i^{-1}\le4(K-1)L.
\]
For \(T=K^3\), \(L=1+4\log K\), \(\ell=\Theta(\log K)\), \(D=\Theta(K\log K)\), and (I1) is \(\Omega(K^2\log K)\). The bracket in (I2) is \(O(K\log^3 K)\). Their ratio is \(\Omega(K/\log^2 K)\), which tends to infinity. Hence no fixed finite C can make (I2) true for exact G on all horizons and legal instances. The lower bound remains below \(T\) because \(K^2\log K=o(K^3)\).

The divergent ratio does not prove linear asymptotic regret on a fixed instance.

\pdfbookmark[0]{Appendix J. Regret Separations Between BHO and Block Learners}{r060.nav.13}
\section*{Appendix J. Regret Separations Between BHO and Block Learners}
\subsection*{J.1 A common long delay}
RATE incurs a separate initial wait for each suboptimal arm in this family. Corollary 1 controls BHO through one maximum cutoff.

\textbf{Proposition J.1 (RATE pays a sum of initial waits).} Fix integers \(K\ge2,D\ge1,T\ge1\) and reals \(0<\Delta\le\mu^*\le1\). Let arm \(K\) be the unique optimum with mean \(\mu^*\) and an arbitrary admissible reward--delay law. All other arms have mean \(\mu^*-\Delta\) and delay at least \(D\) almost surely.

\begin{enumerate}

\def\labelenumi{(\alph{enumi})}

\item

  On every path satisfying this delay condition, Algorithm 2 selects a suboptimal arm in at least \(\min\{T,(K-1)(D+1)\}\) rounds. Consequently,
\begin{equation}
\begin{gathered}
R_T^{\rm rate}\ge\\
\Delta\min\{T,(K-1)(D+1)\}.
\end{gathered}
\tag{J1}
\end{equation}
\item

  If every suboptimal delay equals \(D\) almost surely, unit-width BHO, with \(L=\log(eKT)\), satisfies
\begin{equation}
\begin{gathered}
R_T^{\rm BHO}\le\\
\frac{18473L(K-1)}{\Delta}\\
+6160D\log(e\min\{D,T\}).
\end{gathered}
\tag{J2}
\end{equation}
\item

  Under (b), if \(T\ge(K-1)(D+1)\), then
\begin{equation}
\begin{gathered}
\frac{R_T^{\rm BHO}}{R_T^{\rm rate}}\\
\le\frac{18473L}{\Delta^2(D+1)}\\
+\frac{6160\log(eD)}{\Delta(K-1)}.
\end{gathered}
\tag{J3}
\end{equation}
\end{enumerate}

\textbf{Proof.} For (a), fix a path on which every suboptimal delay is at least \(D\). Algorithm 2 initializes every historical index at one. A historical index changes only after a completed block of its own arm, never increases, and never exceeds one. An unvisited arm therefore retains historical index one.

Suppose arm \(j\le K-1\) is first selected in round \(t_j\), with smaller labels already below one and larger labels unvisited at one. This holds for \(j=1\) at initialization. Before the end of round \(t_j+D\), no report from arm \(j\) can have arrived. Every completed arm-\(j\) block ending earlier therefore has \(\zeta_j(W)=0\), so (Q2) leaves its historical index at one because \(\beta_{\rm rate}(n)>0\). At each such boundary, arm \(j\) remains the smallest label maximizing the historical index and is selected again.

Its historical index can first fall below one only after processing a completed block ending in round \(t_j+D\) or later. The decision in that round precedes its reports. Hence arm \(j\) is selected in every round from \(t_j\) through \(\min\{T,t_j+D\}\), giving \(D+1\) consecutive suboptimal actions unless the horizon ends first. If its historical index never falls, every remaining action selects that arm. Otherwise the next unvisited label is \(j+1\), proving the induction. Thus the last-labeled optimum cannot be selected before all \(K-1\) suboptimal arms have received their disjoint initial \(D+1\)-round intervals. Horizon truncation only ends this suboptimal prefix earlier. Multiplying its length by \(\Delta\) proves (J1), pathwise and in expectation.

For (b), deterministic delay \(D\) gives \(d_i(1)=D\) for every suboptimal arm. We apply Corollary 1 at \(c=0\), so \(\alpha=1\), \(q_\Delta=0\), \(S_T=L(K-1)/\Delta\), and \(\Xi_T(D)=D\log(e\min\{D,T\})\). This gives (J2) without an optimal-arm reporting condition. For (c), divide (J2) by the positive lower bound \(\Delta(K-1)(D+1)\), use \(D/(D+1)\le1\), and replace \(\min\{D,T\}\) by \(D\) inside the logarithm. \(\square\)

Proposition J.1 gives Corollary 3: at \(T\ge(K-1)(D+1)\), (J1) is the RATE lower bound, and (J2) with \(\log(e\min\{D,T\})\le\log(eD)\) is the BHO upper bound. With \(K-1=D=n\ge2\) and \(T=n^3\ge n(n+1)\), the right side of (J3) is \(O(\log n/n)\) at fixed \(\Delta\).

At \(c=0\), the delay term in RATE's upper bound (9) is \(8\Delta(K-1)D\). When the horizon covers all initial waits, (J1) matches this dependence up to a factor smaller than eight.

\textbf{Corollary J.2 (reverse order separation).} Fix \(0<\Delta\le\mu^*\le1\). Take \(K=n+1,D=n,T=n^3\), with integer \(n\ge2\), in Proposition J.1(b). Then
\begin{equation}
\begin{gathered}
\frac{R_{n^3}^{\rm BHO}}{R_{n^3}^{\rm rate}}\\
\le g_\Delta(n):=\\
\frac{18473L_n/\Delta+6160\log(en)}{\Delta(n+1)},\\
\qquad L_n=\log\!\bigl(e(n+1)n^3\bigr),
\end{gathered}
\tag{J4}
\end{equation}
which is \(O(\log n/n)\).

\textbf{Proof.} For \(n\ge2\), \(n^3\ge n(n+1)\). Thus (J1) is \(\Delta n(n+1)\), and division of (J2) by this value gives (J4). Its numerator is \(O(\log n)\) for fixed \(\Delta\). \(\square\)

The last-label condition is material: a never-reporting optimal arm retains historical index one (App. Q) and, once selected, prevents Algorithm 2 from visiting larger labels.

\textbf{Corollary J.3 (one fixed random relabeling).} Under the gap and delay assumptions of Proposition J.1(a), draw one uniform arm-label permutation independently before the run and keep it fixed. Write \(\bar R_T\) for regret averaged over this permutation and the original stochasticity. Then
\begin{equation}
\bar R_T^{\rm rate}\ge\frac\Delta2\min\{T,(K-1)(D+1)\}.
\tag{J5}
\end{equation}
Under deterministic suboptimal delay, on Corollary J.2's family the ratio of the label-averaged regrets, \(\bar R_T^{\rm BHO}/\bar R_T^{\rm rate}\), is at most \(2g_\Delta(n)=O(\log n/n)\).

\textbf{Proof.} Conditional on the complete permutation, let \(B\) be the number of suboptimal labels preceding the optimum. The same prefix argument gives at least \(\min\{T,B(D+1)\}\) suboptimal actions before the first optimal visit, or the horizon ends first. Here \(B\) is uniform on \(\{0,\ldots,K-1\}\). For every such integer \(b\),
\[
\begin{gathered}
\min\{T,b(D+1)\}\ge\\
\frac b{K-1}\min\{T,(K-1)(D+1)\}.
\end{gathered}
\]
Taking expectations and using \(\mathbb EB=(K-1)/2\) proves (J5). We apply Corollary 1 conditional on each full fixed permutation; its upper bound is unchanged by the labels. Averaging and dividing give \(2g_\Delta(n)\). \(\square\)

\subsection*{J.2 Reward-dependent separation from elimination learners}
\textbf{Corollary 2 (reward-dependent separation).} For each integer \(n\ge64\), let
\begin{equation}
\begin{gathered}
\Delta=1/n,\qquad d=n^3,\qquad\\
T=\lfloor n^4/16\rfloor,\qquad \delta=(T+1)^{-2}.
\end{gathered}
\tag{J6}
\end{equation}
Arm 1 gives \((Y,D)=(0,0)\) or \((1,\infty)\), each with probability \(1/2\). Arm 2 has reward \(\operatorname{Bernoulli}(1/2-1/n)\) and delay \(d\). Take independent arms and iid origin rounds. Run unit-width BHO, RATE and NC-lean with their native rules, and SE, PSE and OPSE as in Lancewicki et al.~(2021, Algs. 4--6, with Proc. 2 for SE/PSE). With
\[
\begin{split}
\ell_H&=\log(8T(T+1)^2),\\
\ell_R&=\log(4T(T+1)^3),\\
J&=1+\lfloor\log_2(T+1)\rfloor,\qquad
\ell_C=\log(32(T+1)^3J),
\end{split}
\]
the expected regrets satisfy
\begin{equation}
\begin{gathered}
\begin{split}\\
R_T^{\rm BHO}&\ge n^3/128,\\\\
R_T^{\rm SE},\ R_T^{\rm OPSE}\\
&\ge n^{-1}\lfloor T/2\rfloor,\\\\
R_T^{\rm PSE}\\
&\ge\frac{T-d-8a_T-4}{n}\\
\left(1-T^{-32}-58T^{-9/8}\right)\\
\ge n^3/128,\\
\end{split}
\end{gathered}
\tag{J7}
\end{equation}
where \(a_T=\lceil64\log T\rceil\), whereas
\begin{equation}
\begin{split}
R_T^{\rm NC}&\le U_C(n)
 =8n^2+128n\ell_C+\delta T,\\
R_T^{\rm RATE}&\le U_R(n)
 =8n^2+1024n\ell_R+\delta T.
\end{split}
\tag{J8}
\end{equation}
Consequently, BHO, SE, PSE and OPSE have regret \(\Theta(n^3)\), and RATE and NC-lean have regret \(O(n^2)\). More precisely,
\begin{equation}
\begin{gathered}
R_T^{\rm BHO}=(1/16+o(1))n^3,\qquad\\
\frac{R_T^{\rm RATE}}{R_T^{\rm BHO}},\\
\ \frac{R_T^{\rm NC}}{R_T^{\rm BHO}}\\
\le\frac{128+o(1)}n.
\end{gathered}
\tag{J9}
\end{equation}
The upper bounds in (J8) are strictly below \(n^3/128\) for \(n\ge2048\) (NC-lean) and \(n\ge8192\) (RATE).

We count \(T\) individual actions, stopping an unfinished final scan at the horizon and retaining each learner's updates after complete scans. The literal one-based stopping convention in the cited pseudocode can omit one action: then replace \(\lfloor T/2\rfloor\) by \(\lfloor(T-1)/2\rfloor\) and subtract one from the numerator in (J7). The common \(n^3/128\) lower bounds remain valid.

\textbf{Proof.} The means are \(1/2\) and \(1/2-\Delta\), so \(\Psi_T=\Delta N_2(T)\). Each learner operates separately on this common environment law.

\textbf{RATE and NC-lean.} At \(c=0\), the suboptimal arm has
\[
d_2(1)=w_2(0)=d,
\qquad
b_2(h)=
\begin{cases}
1/2+\Delta,&h<d,\\
0,&h\ge d.
\end{cases}
\]
Proposition 1 gives \(R_T^{\rm RATE}\le8\Delta d+1024\ell_R/\Delta+\delta T\). The first branch of Theorem 3 and \(M_2\ge1\) give
\[
\begin{gathered}
R_T^{\rm NC}\le128\ell_C/\Delta\\
+8\mathbb E[\Delta d/M_2]+\delta T.
\end{gathered}
\]
These are (J8); neither bound requires reports from arm 1.

\textbf{BHO.} Generate each arm's iid potential sequence in pull order. Let \(S_m\) count the successes among the first \(m\) pulls of arm 1 and \(V_m\) the zeros among the first \(m\) pulls of arm 2; set \(S_0=V_0=0\) and \(z=1/2+\Delta\). Hoeffding and a union bound over the two sequences and \(1\le m\le T\) give an event \(E\) of probability at least \(1-\delta/4\) on which
\begin{equation}
\begin{gathered}
S_m\le m/2+\sqrt{m\ell_H/2},\qquad\\
V_m\le zm+\sqrt{m\ell_H/2}.
\end{gathered}
\tag{J10}
\end{equation}
Indeed, its failure probability is at most \(2Te^{-\ell_H}=\delta/4\). The event controls every prefix before adaptive pull counts are substituted.

Write \(o_i(t)\) for the report count at completed time \(t\). All zeros of arm 1 return immediately and all its ones remain pending. Thus, at pull count \(m>0\),
\[
\begin{gathered}
G_1=0,\qquad o_1=m-S_m,\qquad\\
u_1^+=S_m/m.
\end{gathered}
\]
On \(E\), the current index and historical minimum satisfy
\[
\begin{gathered}
H_1+\beta_1\le\rho_1+\beta_1\\
\le1/2+3\beta_1.
\end{gathered}
\]
Put
\begin{equation}
\begin{gathered}
b=\left\lfloor\frac{\ell_H}{2\Delta^2}\right\rfloor+1,\\
\qquad h_0=1/2+3\Delta<1.
\end{gathered}
\tag{J11}
\end{equation}
At \(m_1=b\), one has \(\beta_1<\Delta\), so \(H_1+\beta_1<h_0\).

Suppose arm 1's count has not exceeded \(b\) through completed time \(t\). At any earlier completed time \(s\), let \(m=m_2(s)>0\), \(q=m_2((s-d)_+)\) and \(B=m-q\). Exactly the first \(q\) arm-2 outcomes have returned, so \(u_2^+(s)=1-V_q/m\). The fluctuation in (J10) is absorbed by \(\beta_2\), since \(\sqrt{q\ell_H/2}/m\le\sqrt{\ell_H/(2m)}\). Hence
\begin{equation}
\begin{gathered}
\rho_2(s)\ge1-zq/m\\
=1/2-\Delta+zB/m.
\end{gathered}
\tag{J12}
\end{equation}
For \(s<d\), there are no arm-2 reports and \(\rho_2(s)\ge1\). For \(s\ge d\), at least \(d-b\ge d/2\) (by the size bound below) of the last \(d\) actions select arm 2. Using \(m\le T\le d/(16\Delta)\) and \(z\ge1/2\), (J12) gives \(\rho_2(s)\ge h_0\). Zero counts have historical index one. Every historical update is therefore at least \(h_0\), and \(H_2(t)\ge h_0\).

If arm 1 were about to receive its \((b+1)\)-st pull for the first time, its count would be \(b\) and the preceding argument would yield
\[
H_1+\beta_1<h_0\le H_2.
\]
It would be ineligible. Thus \(N_1(T)\le b\) on \(E\), and
\begin{equation}
\Psi_T^{\rm BHO}\ge\Delta(T-b).
\tag{J13}
\end{equation}
The needed size condition holds for every \(n\ge64\):
\[
\begin{gathered}
\ell_H\le12\log n,\qquad\\
2(b+1)\le12n^2\log n+4<n^3=d.
\end{gathered}
\]
For the last inequality, \(x-12\log x\) increases on \(x\ge64\), and \(\log64=6\log2<9/2\) gives a margin greater than ten at 64. Also \(T\ge4d\), \(T\ge2b\) and \(T\ge n^4/32\). Consequently,
\[
\Pr\{\Psi_T^{\rm BHO}\ge n^3/64\}\ge1-\delta/4,
\]
which implies the first bound in (J7). Since \(b=O(n^2\log n)\), (J13), its event probability, and the universal upper bound \(\Psi_T\le\Delta T\) give the first relation in (J9).

\textbf{SE and OPSE.} For SE, Procedure 2 uses
\[
\begin{gathered}
\widehat\mu_i=G_i/(o_i\vee1),\qquad\\
r_i=\sqrt{2\log T/(o_i\vee1)},\qquad\\
L_i=\widehat\mu_i-r_i,\quad U_i=\widehat\mu_i+r_i.
\end{gathered}
\]
Every returned reward of arm 1 is zero, so \(L_1<0\), whereas \(U_2>0\). The elimination condition \(U_2<L_1\) never holds. For OPSE the lower and upper centers are \(G_i/m_i\) and \((G_i+m_i-o_i)/m_i\), with radius \(\sqrt{2\log T/m_i}\). Again \(L_1<0<U_2\) after the first scan. Both learners also have \(U_2>L_2\) because the confidence radius is positive. Arm 2 is never eliminated by either learner. While both arms remain, scans sample them equally; if arm 1 is eliminated, all later actions choose arm 2. Thus \(N_2(T)\ge\lfloor T/2\rfloor\) on every path. Clipping intervals to \([0,1]\) leaves the strict elimination condition impossible.

\textbf{PSE.} Its first phase has cumulative report threshold \(a_T=\lceil64\log T\rceil\): the phase counter is incremented from zero to one before sampling. Arms that reach the threshold pause until the phase ends; elimination follows only after all active arms reach it.

Let \(E_1\) be the event that the first \(4a_T\) potential outcomes of arm 1 contain at least \(a_T\) zeros. Let \(E_2\) be the event that every arm-2 reward prefix of length \(k\ge a_T\) has mean at least \(3/8\). Hoeffding gives
\[
\Pr(E_1^c)\le e^{-a_T/2}\le T^{-32}.
\]
Since \(\mu_2\ge15/32\), a geometric union bound gives
\begin{equation}
\begin{split}
\Pr(E_2^c)
&\le\sum_{k=a_T}^{\infty}e^{-9k/512}\\
&=\frac{e^{-9a_T/512}}{1-e^{-9/512}}
\le58T^{-9/8}.
\end{split}
\tag{J14}
\end{equation}
Here \((1-e^{-x})^{-1}\le1+x^{-1}\), and \(1+512/9<58\). Arm 2's constant delay preserves its pull order, so its returned rewards form a prefix controlled by \(E_2\), including at a random phase-end count.

On \(E_1\), arm 1 pauses within \(4a_T\) pulls. An unpaused arm is sampled at least once per two actions, with at most two more actions to complete a scan. The first \(a_T\) pulls of arm 2 similarly occur within \(2a_T+2\) actions and return at most \(d\) rounds later. Allowing scan-end checks, the first phase ends by
\[
\tau\le d+8a_T+4.
\]
On \(E_1\cap E_2\), both report counts are then at least \(a_T\). Both radii are at most \(r=1/\sqrt{32}\), and
\begin{equation}
U_1\le r<3/8-r\le L_2.
\tag{J15}
\end{equation}
Thus PSE strictly eliminates arm 1. It retains arm 2 because \(U_2\ge0>L_1\) and \(U_2>L_2\), and all subsequent phases contain only arm 2. This proves the first PSE bound in (J7).

For \(n\ge64\), \(a_T\le256\log n+1\), so \(8a_T+4\le2048\log n+12\le d\). Since \(T\ge4d\), the phase ends by \(T/2\). Moreover \(T\ge2^{20}\), so the failure bound \(T^{-32}+58T^{-9/8}\) is below \(1/2\). Hence
\[
R_T^{\rm PSE}\ge T/(4n)\ge n^3/128.
\]
With one action omitted, the conservative bound \((T-2)/(4n)\) still exceeds \(n^3/128\), using \(T\ge n^4/16-1\) and \(n^4\ge96\).

Finally, \(J\le n\), \(\ell_C\le13\log n\), \(\ell_R\le16\log n\) and \(\delta T\le1\) give
\begin{equation}
\begin{gathered}
U_C\le8n^2+1664n\log n+1,\qquad\\
U_R\le8n^2+16384n\log n+1.
\end{gathered}
\tag{J16}
\end{equation}
The envelopes in (J16), divided by \(n^3/128\), decrease for \(n\ge64\). At \(n=2048\) and \(8192\), respectively, they are bounded by
\begin{equation}
\begin{gathered}
\frac12+\frac{13(11\log2)}{256}\\
+\frac{128}{2048^3}<1,\qquad\\
\frac18+\frac{13\log2}{32}\\
+\frac{128}{8192^3}<1.
\end{gathered}
\tag{J17}
\end{equation}
These prove the stated sufficient thresholds. The logarithms are \(O(\log n)\), so (J8) has leading upper term \((8+o(1))n^2\). Combining this with the BHO asymptotic proves the ratios in (J9); \(\Psi_T\le\Delta T\) supplies the remaining upper orders. \(\square\)

The family varies with \(n\); it does not give fixed-instance linear regret for BHO. SE/PSE's reward-independent guarantees do not apply, and OPSE's optimal-arm high quantile is infinite.

\pdfbookmark[0]{Appendix K. Maturity Threshold in the Bernoulli Designs}{r060.nav.14}
\section*{Appendix K. Maturity Threshold in the Bernoulli Designs}
For unit-width BHO, (A8) requires \(h_i\alpha\Delta_i\ge4416\ell\) at a mature suboptimal selection on the fixed-cutoff event. In the admissible Bernoulli cells of Apps. L and O at \(T=2^{18}\), these products are at most \(65536/4\), below \(4416\ell\) at \(\delta=(T+1)^{-2}\). The analysis therefore charges these selections to the cold term on that event. Their finite-horizon allocation can still depend on delay.

\pdfbookmark[0]{Appendix L. Fixed-Delay Allocation Comparisons}{r060.nav.15}
\section*{Appendix L. Fixed-Delay Allocation Comparisons}
\subsection*{L.1 Fixed design, information and interval calibration}
The 24 shared-delay cells cross six arm counts, two optimal delays and two bad delays, each with 32 paired seeds and four policies. Their rewards are Bernoulli with mean \(3/4\) for the optimal arm and \(1/2\) for every suboptimal arm, so \(\Delta=1/4\); policies, information and reporting follow O.1. Ten deterministic construction cells use \(K=8,16,32,64,128\), each at cubic and fixed 1048576 horizons. Four \(K=8\) cells add an eligibility-multiplier diagnostic; the data archive retains the half-width endpoints at \(K=8\) under both optimum-reporting conditions, and Appendix M gives the guarantee. There are 3,240 completed policy runs and 869,419,008 actions.

A policy's delay excess is its regret at bad delay 2048 minus its same-seed regret at bad delay zero, with the optimal delay held fixed. The secondary family has 36 paired BHO-minus-comparator delay-excess contrasts: six arm counts, two optimal delays and three comparators. Calibration follows O.2 with \(p=36\): the primary Bonferroni-\(t_{31}\) critical value is 3.511670; fixed-SE bootstrap sensitivity uses 10,000 whole-seed-vector resamples and critical value 3.103631. Other intervals are descriptive pointwise Student-t; deterministic single runs have none. No asymptotic slope is fitted. Of the 36 primary intervals, 32 lie below zero (BHO favored), 1 lies above zero (comparator favored) and 3 include zero; the bootstrap sensitivity intervals give 33, 1 and 2, respectively. The difference comes from the BHO-minus-G contrast at \(K=4\), optimal delay zero, whose primary interval includes zero: mean −68.4375, paired-difference SE 20.8302, primary interval {[}−141.5864, 4.7114{]}, versus {[}−133.0869, −3.7881{]} under the fixed-SE bootstrap. In the matched \(K=8,D=2048\) all-report setting, mean delay excesses are 716.07 for BHO, 448.00 for PSE and 364.69 for SE (the last two from Z03 minus Z01 in Appendix O).

All 24 primary intervals at \(K\ge8\) lie below zero. The sole positive interval is against OUCB-current at \(K=4\) with immediate optimal feedback. Delay excess and total regret can rank methods differently. The data archive supplies the full 24-cell endpoint table and all 36 contrasts.

\subsection*{L.2 Deterministic probes}
The ten probes adapt the construction of I.2 to \(K=8,16,32,64,128\), each at \(T=K^3\) and \(T=2^{20}\), with a fixed label permutation per cell. At \(K\ge64\), G and OUCB-current incur 2.8--10.2 times BHO's regret. At \(K\le32\), one of the two matches or beats BHO in four of six probes. The arm counts are below the sufficient range of Proposition I.1. The data archive gives the construction, label permutations and full endpoints.

\pdfbookmark[0]{Appendix M. Positive Eligibility Widths Preserve the Shared-Delay Guarantee}{r060.nav.16}
\section*{Appendix M. Positive Eligibility Widths Preserve the Shared-Delay Guarantee}
\subsection*{M.1 Model and policy}
We use Section 2.2 with fixed deterministic \(\theta\in(0,1]\), arbitrary \(\delta\in(0,1)\), and \(\ell=\log(4KT/\delta)\). We adopt Appendix A.1's \(Z,c,\alpha,h_i,\gamma_i,D\) and finite ordinary or weighted cutoff conditions. Optimal-arm delays remain unrestricted. We replace unit-width eligibility by
\begin{equation}
\mathcal C_\theta(t)=\{i:H_i(t)+\theta\beta_i(t)\ge M(t)\}.
\tag{M1}
\end{equation}
Define \(N_\theta=\lceil32/\theta\rceil\) and
\begin{equation}
C_\theta=3(96N_\theta/\theta+16).
\tag{M2}
\end{equation}
\subsection*{M.2 Guarantee}
\textbf{Theorem M.1 (Positive eligibility widths).} Under the model, policy, and fixed parameter choices in Section M.1, with probability at least \(1-\delta\),
\begin{equation}
\Psi_T\le\min\left\{T, \frac{4618\ell Z}{\alpha^2\theta^2} +\frac{C_\theta\Phi_T(D)}{\alpha}\right\}.
\tag{M3}
\end{equation}
At \(\delta=(T+1)^{-2}\), the expected bound is (6), with the same weighted-cutoff extension as Theorem 2. In particular,
\begin{equation}
R_T\le\min\left\{T,\frac{18500}{\theta^2} \left[\frac{LZ}{\alpha^2}+\frac{\Phi_T(D)}{\alpha}\right]\right\}.
\tag{M4}
\end{equation}
The same policy has these bounds for every qualifying \(c\). At \(\theta=1/2\), \(N_\theta=64\), \(C_{1/2}=3(96\cdot64/(1/2)+16)=36912\), and the expected statistical coefficient is \(18473/(1/2)^2=73892\).

\subsection*{M.3 Prefix event and the upper bound (A6)}
A.1's event \(\mathcal E=E\) and its probability are width-independent. A.2 gives \(H_i\ge\mu_i\), \(M\ge\mu^*\) and (A6), with \(B_i\) from (11). Selection under (M1) gives \(H_i\ge M-\theta\beta_i\), requiring neither nested eligibility nor an eligible optimum.

The count-only lemma (A9)--(A10) applies without equating unit-width and general-width actions.

\subsection*{M.4 Cold selections}
Put \(A=96/\theta\) and \(Q_i=A^2\ell/(2\gamma_i^2)\). Each suboptimal arm has at most \(Q_i+1\) selections with pre-action count below \(Q_i\), so
\begin{equation}
\Psi_{cold}\le[4608\ell/(\alpha^2\theta^2)+1]Z.
\tag{M5}
\end{equation}
At a mature selection, \(\beta_i\le\theta\gamma_i/96\). On \(\mathcal E\), selected eligibility and (A6) imply
\begin{equation}
\begin{gathered}
\frac{B_i}{m_i}\ge\gamma_i-(3+\theta)\beta_i\\
\ge\left[1-\frac{\theta(3+\theta)}{96}\right]\gamma_i,\\
\qquad h_i\gamma_i\ge C_m(\theta)\ell,
\end{gathered}
\tag{M6}
\end{equation}
where \(C_m(\theta)=48(96-3\theta-\theta^2)/\theta^2\), equal to 4416 at \(\theta=1\) and 18096 at \(\theta=1/2\). Thus \(D\gamma_{\max}<C_m(\theta)\ell\) forces cold bad selections on \(\mathcal E\), including \(D=0\); equality and \(\mathcal E^c\) are not covered. All selections are cold for \(T=1\). Otherwise assume \(T\ge2\), finite \(D\ge1\) for calendar counting.

\subsection*{M.5 High-optimism bins}
Let \(v(t)=M(t)-\mu^*\); if a suboptimal arm exists, \(0\le v<1\). Set \(P=\lceil\log_2T\rceil+2\). The dyadic bands \([a,2a)\), with \(a=2^{-r}\) for \(r=0,\ldots,P-1\), cover the relevant values \(v\ge1/T\). Divide each band into \(N_\theta\) equal bins of width \(a/N_\theta\le\theta a/32\).

Within one bin \([b,b+a/N_\theta)\), charge mature suboptimal selections satisfying \(\gamma_i\le16a/\theta\). Let \(t_0\) be the first charged time, \(F\) the set of charged arms, and \(s_j\) the last charged time of arm \(j\). At \(s_j\), maturity gives \(\beta_j\le\gamma_j/A\le a/6\), so
\begin{equation}
H_j(s_j)\ge\mu^*+b-\theta a/6.
\tag{M7}
\end{equation}
Historical monotonicity transports this bound to every earlier time. If \(m_j(t_0)<81\ell/(8a^2)\), including zero count, then \(\beta_j(t_0)>2a/9\) and
\[
\begin{gathered}
H_j(t_0)+\theta\beta_j(t_0)\\
>\mu^*+b+\theta a/18\\
>\mu^*+b+a/N_\theta>M(t_0).
\end{gathered}
\]
Thus \(j\) is eligible. The actual mature selected arm satisfies
\[
m_i(t_0)\ge A^2\ell/(2\gamma_i^2)\ge18\ell/a^2>81\ell/(8a^2),
\]
contradicting count priority. Hence every future charged arm satisfies \(\beta_j(s)\le2a/9\) throughout \([t_0,s_j]\). Equations (A6) and (M7) give
\begin{equation}
\begin{gathered}
B_j(s,h_j)/m_j(s)\\
\ge b-\theta a/6+\gamma_j-2a/3\\
\ge a/6+\gamma_j\ge a/6,
\end{gathered}
\tag{M8}
\end{equation}
using \(\theta\le1\). We apply (A10) with \(N=81\ell/(8a^2)\) and \(g=a/3\). Its ratio is \(R_H=\min\{8a^2T/(81\ell),48aD/(81\ell)\}\). As in Appendix A, \(\ell>\log8>81/64\) gives \(1\le R_H\le M_D:=\min\{D,T\}\) for a nonempty group; \(R_H<1\) means that group is empty. Since every charged selection has \(\Delta_i\le16a/(\alpha\theta)\), the bin cost is at most
\begin{equation}
\begin{gathered}
\frac{16aD}{\alpha\theta}\left(1+\frac{6\log M_D}{a}\right)\\
\le\frac{96D}{\alpha\theta}\log(eM_D).
\end{gathered}
\tag{M9}
\end{equation}
There are \(N_\theta\) bins in each band.

\subsection*{M.6 Low-optimism bands}
Charge mature selections satisfying \(v(t)<\theta\gamma_i/8\) and \(\gamma_i\ge1/T\). Partition the effective gaps into bands \([\epsilon,2\epsilon)\) at the same \(P\) dyadic scales, including \(\gamma_i=1\) in the \(\epsilon=1\) band. At the first charged time \(t_0\), \(M(t_0)<\mu^*+\theta\epsilon/4\). Every future charged arm \(j\) satisfies
\begin{equation}
\begin{gathered}
H_j(s_j)\ge\mu^*-\theta\beta_j(s_j)\\
\ge\mu^*-\theta\epsilon/48,
\end{gathered}
\tag{M10}
\end{equation}
where \(\beta_j\le\gamma_j/A<2\epsilon/A\le\epsilon/48\). This deliberately loose bound is valid for \(\theta\le1\). If \(m_j(t_0)<512\ell/(81\epsilon^2)\), then \(\beta_j(t_0)>9\epsilon/32\), and its eligibility score satisfies
\[
\begin{gathered}
H_j(t_0)+\theta\beta_j(t_0)\\
>\mu^*+\theta(9/32-1/48)\epsilon\\
>\mu^*+\theta\epsilon/4>M(t_0).
\end{gathered}
\]
The actual mature selected count is at least \(A^2\ell/(8\epsilon^2)\ge1152\ell/\epsilon^2\), contradicting count priority. Therefore \(\beta_j(s)\le9\epsilon/32\) throughout the interval, and
\begin{equation}
\begin{gathered}
B_j(s,h_j)/m_j(s)\\
\ge\gamma_j-\theta\epsilon/48-27\epsilon/32\\
\ge13\epsilon/96\ge\epsilon/8.
\end{gathered}
\tag{M11}
\end{equation}
We apply (A10) with \(N=512\ell/(81\epsilon^2)\) and \(g=\epsilon/4\). The ratio is \(R_L=\min\{81\epsilon^2T/(512\ell),648\epsilon D/(512\ell)\}\). The same lower bound on \(\ell\) gives \(1\le R_L\le M_D\) whenever this group is nonempty; ratios below one imply an empty group. Since \(\Delta_i<2\epsilon/\alpha\), the band cost is at most
\begin{equation}
\begin{gathered}
\frac{2\epsilon D}{\alpha}\left(1+\frac{8\log M_D}{\epsilon}\right)\\
\le16D\log(eM_D)/\alpha.
\end{gathered}
\tag{M12}
\end{equation}
\subsection*{M.7 Exhaustion and constants}
Mature selections with \(\gamma_i<1/T\) cost at most \(1/\alpha\) over the entire horizon. For a selection with \(\gamma_i\ge1/T\) and \(v\ge1/T\), consider its \(v\) band. If \(\gamma_i\le16a/\theta\), charge it to (M9); otherwise, \(v<2a<\theta\gamma_i/8\), so charge it to (M12). When \(v<1/T\) and \(\gamma_i\le8/(\theta T)\), the total cost of that class is at most \(8/(\alpha\theta)\); otherwise, the low-optimism condition holds. Overlap only increases the bound. Thus all mature selections are accounted for, with small-class cost at most \(9/(\alpha\theta)\).

Let \(\gamma_{\min}\) and \(\gamma_{\max}\) range over the nonempty bad-arm set. In the cold-only case set \(H_\theta=L_\theta=0\). Otherwise put \(\Gamma_\theta=\max\{\gamma_{\min},C_m(\theta)\ell/D\}\le\gamma_{\max}\). The high condition requires \(a\ge\theta\Gamma_\theta/16\); the low condition requires \(\varepsilon>\Gamma_\theta/2\), strictly. Therefore the respective numbers of potentially nonempty scales are at most
\begin{equation}
\begin{gathered}
H_\theta=\max\{0,\min\{P-1,\lfloor\log_2(16/(\theta\Gamma_\theta))\rfloor\}\},\\
\quad\\
L_\theta=\max\{0,\min\{P,\lceil\log_2(2/\Gamma_\theta)\rceil\}\}.
\end{gathered}
\tag{M13}
\end{equation}
Combining the unchanged per-bin and per-band costs yields on \(\mathcal E\)
\begin{equation}
\begin{gathered}
\Psi_T\le[4608\ell/(\alpha^2\theta^2)+1]Z+9/(\alpha\theta)\\
+\left[\frac{96N_\theta}{\theta}H_\theta+16L_\theta\right]\\
\frac{D\log(eM_D)}{\alpha}.
\end{gathered}
\tag{M14}
\end{equation}
For \(0<\theta\le1\), \(\theta C_m(\theta)\ge4416\) and \(C_m(\theta)\ge4416\): both follow from \(96-3\theta-\theta^2\ge92\) and \(\theta\le1\). If \(D<T\), every nonempty high scale thus has \(a\ge276\ell/D\), and every nonempty low scale has \(\varepsilon>2208\ell/D\). The A.6 argument bounds both counts by \(3\log(eD)\). If \(D\ge T\), use \(P\le3\log(eT)\); for \(T\ge3\), this follows from \(\lceil\log_2T\rceil+2\le\log(T)/\log2+3\le3\log(eT)\), and \(T=2\) is direct. Hence both counts are at most \(3\log(eM_D)\) in every case. Combining (M5), (M9), and (M12) gives
\begin{equation}
\begin{gathered}
\Psi_T\le[4608\ell/(\alpha^2\theta^2)+1]Z\\
+9/(\alpha\theta)\\
+C_\theta\Phi_T(D)/\alpha.
\end{gathered}
\tag{M15}
\end{equation}
Since \(Z\ge1\), \(\ell>1\), \(\alpha\le1\), and \(\theta\le1\), the first two terms are bounded by \(4618\ell Z/(\alpha^2\theta^2)\). This proves (M3). Retaining (M14) instead gives the sharper expectation bound at \(\delta=(T+1)^{-2}\) with statistical term \(18473LZ/(\alpha^2\theta^2)\) and delay term \([96N_\theta H_\theta/\theta+16L_\theta]D\log(eM_D)/\alpha\).

For \(\delta=(T+1)^{-2}\), we have \(\ell\le4L\), failure contributes at most \(\delta T\le1\), and \(LZ/(\alpha^2\theta^2)\ge1\). This proves the expectation bound (6). Using \(N_\theta\le32/\theta+1\),
\[
C_\theta\theta^2\le9216+288\theta+48\theta^2\le9552<18500,
\]
and \(18473<18500\), which proves the unified form. At \(\theta=1\), \(N_\theta=32\) and \(C_\theta=9264\), recovering the coefficient in Appendix A; at \(\theta=1/2\), \(N_\theta=64\) and \(C_\theta=36912\).

\subsection*{M.8 Scope}
Theorem M.1 covers the half-width rule in Appendix O. At fixed finite cutoffs its delay term is \(D\log^2(eD)\). Every fixed positive width preserves pointwise sublinearity on the positive-slack fixed-instance class. The width may depend on the known horizon but must remain fixed during each run; for vanishing \(\theta_T\), \(\theta_T^{-2}\log(eT)=o(T)\) suffices.

At zero width, defining eligibility as \(\{i:H_i=M\}\) gives exactly G, with the separation in Proposition I.1. The best positive width, dominance over unit width, necessity of the cutoff logarithms and optimality of the \(\theta^{-2}\) dependence remain open. Unit width is the theoretical reference; the half-width experiments do not select a practical default.

\pdfbookmark[0]{Appendix N. Immediate Feedback and Exploration Cost}{r060.nav.17}
\section*{Appendix N. Immediate Feedback and Exploration Cost}
\subsection*{N.1 Setting}
We work in the iid-round model of Section 2.1 with the policy of Appendix M, for fixed \(\ell>0\) and \(\theta\in(0,1]\); the case \(\theta=0\) is the greedy rule G of M.8, with eligible set \(\{i:H_i=M\}\) and key \((m_i,-H_i,i)\). Counts are pre-action; zero-delay reports arrive at round end.

\subsection*{N.2 Exact exploration coefficients for deterministic immediate rewards}
Now all rewards \(\mu_i\) are deterministic, all delays zero and the optimum unique, allowing \(\mu^*=1\). We retain exact arithmetic and the same policy/ties. For each bad arm set
\begin{equation}
\begin{gathered}
a_i=\frac{(1+\theta)^2\ell}{2\Delta_i^2},\qquad\\
B_i=\lfloor a_i\rfloor+1,\qquad B=\sum_{i\in\mathcal I}B_i.
\end{gathered}
\tag{N1}
\end{equation}
\textbf{Proposition N.1 (Finite deterministic squeeze).} For every \(T\ge1\) and \(\ell>0\), \(N_i(T)\le B_i\). If
\begin{equation}
\begin{gathered}
s=T-B-1>\max_{i\in\mathcal I}B_i,\\
\qquad \varepsilon=\sqrt{\ell/(2s)},
\end{gathered}
\tag{N2}
\end{equation}
then every bad arm also satisfies
\begin{equation}
N_i(T)>\frac{(1+\theta)^2\ell}{2(\Delta_i+\varepsilon)^2}.
\tag{N3}
\end{equation}
\emph{Proof.} Deterministic immediate reports give the exact identity
\begin{equation}
H_i(t)=\min\{1,\mu_i+\beta_i(t)\}
\tag{N4}
\end{equation}
at every positive count. Indeed, the current score decreases with count and is unchanged between selections. Thus its historical minimum adds no further reduction. At zero count the historical index is one. In particular, \(M\ge\mu^*\). A bad arm selected at positive count obeys
\[
\begin{gathered}
\mu^*\le M\le H_i+\theta\beta_i\\
\le\mu_i+(1+\theta)\beta_i.
\end{gathered}
\]
At \(\theta=0\) the middle relation means \(H_i=M\). Its pre-action count is at most \(a_i\), proving \(N_i(T)\le B_i\).

Under (N2), there is an optimal action because there are at most \(B<T\) bad actions. At the last optimal action, let \(m_*\) be the optimum's pre-action count. Uniqueness gives
\begin{equation}
m_*=N_*(T)-1\ge T-B-1=s>\max_i B_i.
\tag{N5}
\end{equation}
Every bad arm then has strictly smaller count than the optimum. Since count is the first selection key, no bad arm can be eligible at this decision. An arm attaining \(M\) is always eligible, so necessarily \(M=H_*\). Every bad arm has positive count: a zero-count arm would be eligible through its infinite width at \(\theta>0\), or through its historical index one attaining \(M\) at \(\theta=0\). Its historical index cannot equal one either, because this would also imply eligibility. Formula (N4) is therefore unclipped for every bad arm. Ineligibility yields
\begin{equation}
\begin{gathered}
\mu_i+(1+\theta)\beta_i=H_i+\theta\beta_i\\
<M=H_*\le\mu^*+\sqrt{\ell/(2m_*)}\\
\le\mu^*+\varepsilon.
\end{gathered}
\tag{N6}
\end{equation}
At \(\theta=0\), the strict inequality is exactly nonmembership in \(\{i:H_i=M\}\). Rearranging (N6) gives the strict count lower bound in (N3) at this decision. Counts are nondecreasing, so the same bound holds at the horizon. \(\square\)

\textbf{Corollary N.2 (Deterministic leading coefficient).} Fix a deterministic immediate-feedback instance satisfying Proposition N.1, with \(K\ge2\) and all gaps fixed as \(T\) grows. If \(\ell_T\to\infty\) and \(\ell_T=o(T)\), then for each fixed \(\theta\in[0,1]\),
\begin{equation}
\begin{gathered}
\frac{N_i(T)}{\ell_T}\longrightarrow\\
\frac{(1+\theta)^2}{2\Delta_i^2},\qquad\\
\frac{R_T^{(\theta)}}{\ell_T}\longrightarrow\\
\frac{(1+\theta)^2}{2}\sum_{i\in\mathcal I}\frac1{\Delta_i}.
\end{gathered}
\tag{N7}
\end{equation}
For rules using the same confidence logarithm,
\begin{equation}
\begin{gathered}
\frac{R_T^{(1)}}{R_T^{(0)}}\to4,\qquad\\
\frac{R_T^{(1/2)}}{R_T^{(0)}}\to\frac94,\qquad\\
\frac{R_T^{(1)}}{R_T^{(1/2)}}\to\frac{16}{9}.
\end{gathered}
\tag{N8}
\end{equation}
\emph{Proof.} With fixed gaps, \(B=O(\ell_T)\), so (N2) holds eventually and \(\varepsilon\to0\). Divide (N1) and (N3) by \(\ell_T\) and apply the squeeze theorem to obtain the count limit. Sum against the fixed gaps to obtain (N7); the deterministic regrets equal their expectations. The limiting coefficients are positive, so their ratios give (N8). The paper's \(\ell_T=\log(4KT(T+1)^2)\) meets the two conditions. \(\square\)

Added eligibility width changes approximate exclusion from \(\mu_i+\beta_i<\mu^*\) to \(\mu_i+(1+\theta)\beta_i<\mu^*\), giving the squared leading factor. Stochastic historical minima need not satisfy (N4).

\pdfbookmark[0]{Appendix O. Synthetic Reporting-Mechanism Experiments}{r060.nav.18}
\section*{Appendix O. Synthetic Reporting-Mechanism Experiments}
\subsection*{O.1 Protocol, methods, and fixed population}
The 59 environments have 32 paired seeds and seven methods (13,216 endpoints): 1,152 matching saved fixed-delay controls and 12,064 new runs. Matching environment/seed/method/endpoint rows count once, preserving 32 seed units. Every run has \(T=262144\), best mean \(3/4\), bad means \(1/2\), and a separate seed-specific random relabeling of logical arms.

The seven methods are floating-point BHO, reconstructed historical-greedy G, specified OUCB-current, OPSE, BHO-half (eligibility \(\theta=1/2\)), PSE and SE. G implements the historical greedy rule in Appendix I. BHO-half has the separate Appendix M guarantee. Section O.7 adds returned-UCB as an eighth method in separate runs.

PSE and SE follow Lancewicki et al.~(2021), Algorithms 5 and 4 with Procedure 2. Their confidence means and radii use cumulative received counts: \(\widehat\mu_i=G_i/(n_i\vee1)\) and \(r_i=\sqrt{2\log(T)/(n_i\vee1)}\). At completed ascending-label sweeps, an active arm is eliminated simultaneously when its UCB is strictly below the largest lower confidence bound (LCB) over \textbf{all \(K\) arms}, including previously eliminated arms. For PSE, phase \(\ell=1\) begins with the permanent active set; an arm pauses after a full sweep once cumulative received count reaches \(16\log(T)4^\ell\). Counts and sums never reset. Paused-arm reports continue to arrive. Elimination occurs only after every arm in the current phase sampling set has paused. OUCB-current maximizes \(u_i^++\sqrt{2\log T/m_i}\) without clipping or historical minimization; a zero-count arm has infinite score, and ties favor smaller pull count then smaller observed label. OPSE uses Algorithm 6's active-only bound/elimination-reference convention: \(L_i=G_i/m_i-\sqrt{2\log T/m_i}\), \(U_i=u_i^++\sqrt{2\log T/m_i}\), with simultaneous elimination at completed ascending-label sweeps when \(U_i<\max_{j\in S}L_j\); counts never reset.

Each round selects one arm and increments its pull count, generates its latent reward/delay pair, then delivers all due reports at round end, sorted by observed label and reward. Delay-zero reports first affect the next action. Sampling sets stay fixed within sweeps; a final incomplete sweep is truncated without completion. Policies receive no originating clock, latent delay, missingness probability, true gap, logical best identity or unreported reward.

Bad-arm missing mass is \(p\in\{0,1/8,3/16\}\). Given rewards \((0,1)\), F/I/S have missing probabilities \((2p,0)\), \((p,p)\) and \((0,2p)\), hiding failures/independent outcomes/successes. Nonmissing outcomes return after \(d\). The common marginal is \(F(d)=1-p\), but completion deficits are \(p,p/2,0\); ordinary slack is \(\alpha=1-4p\). At \(p=0\), the identical couplings occur once. Only the first \(q\) logical bad arms are delayed in the slow-count scan; others report immediately. Optimal-arm laws are fixed-delay all-report (A), never-report (N), immediate zeros only (Z), or immediate successes only (R). R occurs solely in the all-success-reporting negative control, whose bad-arm missing mass and deficit are \(1/2\), outside positive slack.

PSE/SE's reward-independent assumptions fail for Z/R optimum laws and positive-missingness F/S bad-arm laws; these runs test their behavior outside the reward-independent class. A never-reporting optimum is compatible with reward independence: PSE can pause reporting bad arms and stall its phase while continuing to pull that optimum, and SE's optimal-arm quantile bound can be vacuous.

\subsection*{O.2 Contrasts and uncertainty}
Simultaneous intervals in Apps. L, O and P use paired seed differences as Monte Carlo units. For family dimension \(p\), nominal 95\% Bonferroni-\(t_{31}\) intervals are \(\bar d\pm t_{31,1-0.05/(2p)}\operatorname{SE}(d)\). Bootstrap sensitivity uses fixed-original-SE centered max-absolute-\(z\) calibration with whole-seed-vector resampling, preserving cross-coordinate covariance and without re-studentizing replicates. Both calibrations are approximate; neither has distribution-free finite-sample familywise coverage. Flagged zero sample variance remains descriptive.

The primary family has 118 paired contrasts of mean pseudo-regret, BHO minus PSE and BHO minus SE in all 59 environments; positive values favor the comparator. The secondary family has 140 signed delay-excess contrasts (positive-delay regret minus the same seed's zero-delay regret) over two reporting laws (all-report optima with no missingness and zeros-only optima with bad-arm failure hiding at \(p=1/8\)), two values of \(K\), five delays and seven methods. Bonferroni critical values are 3.946546 and 4.008127; bootstrap sensitivity uses 20,000 replicates and critical values 3.277389/3.432755. The plan for the 12,064 new runs was fixed before they were run but after earlier results were known; the 1,152 earlier endpoints are reused.

\subsection*{O.3 Environment key}
\end{multicols}
{\small
\begin{longtable}{>{\raggedright\arraybackslash}p{46.02pt}>{\raggedright\arraybackslash}p{12.86pt}>{\raggedright\arraybackslash}p{51.44pt}>{\raggedright\arraybackslash}p{98.14pt}>{\raggedright\arraybackslash}p{31.49pt}>{\raggedright\arraybackslash}p{62.62pt}>{\raggedright\arraybackslash}p{58.11pt}>{\raggedright\arraybackslash}p{31.14pt}}
\multicolumn{8}{@{}p{\linewidth}@{}}{\textbf{Table O.1. Environment key, grouped by identical coordinates.} A0/A2048: all optimal rewards after 0/2048. N: none. Z/R: immediate zeros/successes only. Bad laws F/I/S hide failures/independent outcomes/successes. Ranges denote consecutive cells; varying entries follow ascending cell order. Each Z01--Z59 appears once: 25 RI (reward independent), 32 stress (reward dependent), 2 NC (outside positive slack).}\\[3pt]
\toprule
\textbf{Cell} & \textbf{K} & \textbf{Optimum} & \textbf{Bad d} & \textbf{Bad p} & \textbf{Missing law} & \textbf{Slow bad q} & \textbf{Scope} \\
\midrule
\endfirsthead
\multicolumn{8}{@{}l}{\footnotesize\itshape Continued from previous page}\\
\toprule
\textbf{Cell} & \textbf{K} & \textbf{Optimum} & \textbf{Bad d} & \textbf{Bad p} & \textbf{Missing law} & \textbf{Slow bad q} & \textbf{Scope} \\
\midrule
\endhead
\midrule\multicolumn{8}{r@{}}{\footnotesize\itshape Continued on next page}\\
\endfoot
\bottomrule
\endlastfoot
Z01--Z03 & 8 & A0 & 0, 512, 2048 & 0 & -- & 7 & RI \\
Z04 & 8 & A0 & 2048 & 1/8 & F & 7 & stress \\
Z05 & 8 & A0 & 2048 & 1/8 & I & 7 & RI \\
Z06 & 8 & A0 & 2048 & 1/8 & S & 7 & stress \\
Z07--Z09 & 8 & A0 & 8192, 32768, 65536 & 0 & -- & 7 & RI \\
Z10--Z11 & 8 & A2048 & 0, 2048 & 0 & -- & 7 & RI \\
Z12 & 8 & N & 2048 & 0 & -- & 7 & RI \\
Z13 & 8 & N & 2048 & 1/8 & F & 7 & stress \\
Z14 & 8 & N & 2048 & 1/8 & I & 7 & RI \\
Z15 & 8 & N & 2048 & 1/8 & S & 7 & stress \\
Z16--Z17 & 8 & Z & 0, 512 & 1/8 & F & 7 & stress \\
Z18 & 8 & Z & 2048 & 0 & -- & 7 & stress \\
Z19--Z21 & 8 & Z & 2048 & 1/8 & F, I, S & 7 & stress \\
Z22--Z24 & 8 & Z & 2048 & 3/16 & F, I, S & 7 & stress \\
Z25--Z27 & 8 & Z & 8192, 32768, 65536 & 1/8 & F & 7 & stress \\
Z28 & 8 & R & 0 & 1/2 & F & 7 & NC \\
Z29--Z31 & 64 & A0 & 0, 512, 2048 & 0 & -- & 63 & RI \\
Z32 & 64 & A0 & 2048 & 1/8 & F & 63 & stress \\
Z33 & 64 & A0 & 2048 & 1/8 & I & 63 & RI \\
Z34 & 64 & A0 & 2048 & 1/8 & S & 63 & stress \\
Z35--Z36 & 64 & A0 & 8192, 32768 & 0 & -- & 63 & RI \\
Z37 & 64 & A0 & 65536 & 0 & -- & 1 & RI \\
Z38 & 64 & A0 & 65536 & 0 & -- & 8 & RI \\
Z39 & 64 & A0 & 65536 & 0 & -- & 32 & RI \\
Z40 & 64 & A0 & 65536 & 0 & -- & 63 & RI \\
Z41--Z42 & 64 & A2048 & 0, 2048 & 0 & -- & 63 & RI \\
Z43 & 64 & N & 2048 & 0 & -- & 63 & RI \\
Z44 & 64 & N & 2048 & 1/8 & F & 63 & stress \\
Z45 & 64 & N & 2048 & 1/8 & I & 63 & RI \\
Z46 & 64 & N & 2048 & 1/8 & S & 63 & stress \\
Z47--Z48 & 64 & Z & 0, 512 & 1/8 & F & 63 & stress \\
Z49 & 64 & Z & 2048 & 0 & -- & 63 & stress \\
Z50--Z52 & 64 & Z & 2048 & 1/8 & F, I, S & 63 & stress \\
Z53--Z55 & 64 & Z & 2048 & 3/16 & F, I, S & 63 & stress \\
Z56--Z58 & 64 & Z & 8192, 32768, 65536 & 1/8 & F & 63 & stress \\
Z59 & 64 & R & 0 & 1/2 & F & 63 & NC \\
\end{longtable}
}
\begin{multicols}{2}
\subsection*{O.4 Endpoint summaries}
Table O.2 compares F, I and S at the same marginal missing probability (Z04--Z06), and shows the two ends of the slow-arm-count scan (Z37 and Z40). The data archive gives all 413 environment--method summaries, of which 147 have mean regret at or above 90\% of uniform regret. These display rows were selected after the runs to illustrate the reporting mechanisms and the scan. All contrasts use the original families in O.2.

\end{multicols}
{\small
\begin{longtable}{>{\raggedright\arraybackslash}p{34.57pt}>{\raggedright\arraybackslash}p{49.27pt}>{\raggedright\arraybackslash}p{49.27pt}>{\raggedright\arraybackslash}p{52.57pt}>{\raggedright\arraybackslash}p{58.32pt}>{\raggedright\arraybackslash}p{49.27pt}>{\raggedright\arraybackslash}p{49.27pt}>{\raggedright\arraybackslash}p{49.27pt}}
\multicolumn{8}{@{}p{\linewidth}@{}}{\textbf{Table O.2. Selected endpoint means (seed SE).} The uniform reference \(T(K-1)/(4K)\) is 57344 for \(K=8\) and 64512 for \(K=64\). † marks a mean at least 90\% of the reference. In Z06=Z03 only BHO, BHO-half, G and OUCB-current share the Z03 trajectory (O.5).}\\[3pt]
\toprule
\textbf{Cell} & \textbf{BHO} & \textbf{BHO-half} & \textbf{G} & \textbf{OUCB-current} & \textbf{OPSE} & \textbf{PSE} & \textbf{SE} \\
\midrule
\endfirsthead
\multicolumn{8}{@{}l}{\footnotesize\itshape Continued from previous page}\\
\toprule
\textbf{Cell} & \textbf{BHO} & \textbf{BHO-half} & \textbf{G} & \textbf{OUCB-current} & \textbf{OPSE} & \textbf{PSE} & \textbf{SE} \\
\midrule
\endhead
\midrule\multicolumn{8}{r@{}}{\footnotesize\itshape Continued on next page}\\
\endfoot
\bottomrule
\endlastfoot
Z04 & 8114.33 (58.33) & 5062.64 (47.92) & 10411.75 (123.73) & 7451.98 (47.65) & 15748.90 (150.81) & 15264.75 (994.52) & 6169.62 (104.84) \\
Z05 & 4215.47 (40.17) & 2823.82 (28.02) & 4111.38 (255.94) & 7603.00 (116.72) & 8411.53 (88.93) & 6956.94 (4.51) & 3372.07 (44.85) \\
Z06=Z03 & 2824.43 (25.95) & 2101.40 (24.39) & 1442.77 (35.31) & 4602.12 (207.44) & 5571.27 (60.03) & 6662.96 (84.33) & 2157.53 (24.64) \\
Z37 & 36378.52 (57.30) & 29778.04 (37.93) & 27083.14 (59.79) & 26962.15 (48.17) & 43873.83 (405.66) & 64759.04 (0.03)† & 26530.48 (337.80) \\
Z40 & 64220.10 (16.18)† & 63864.50 (27.04)† & 58227.19 (103.82)† & 46789.69 (173.70) & 64511.99 (0.01)† & 64768.00 (0.00)† & 37173.82 (355.16) \\
\end{longtable}
}
\begin{multicols}{2}
\subsection*{O.5 Results and limits}
For BHO, BHO-half, G and OUCB-current, \(G_i+m_i-n_i=m_i-\text{reported failures}_i\). Under common potential rewards and ties, returning or hiding successes preserves this numerator; induction gives identical pull counts, completions, indices, eligibility and actions whenever the failure-arrival processes agree. Report counts and sums can differ. OPSE, PSE and SE are excluded. The 59 environments have 45 such mechanism classes, with 14 repeats marked in the full endpoint tables in the data archive. All 1792 representative/duplicate method-seed action/outcome records agree after removing the missingness flag. BHO's lower-mean counts versus G and OUCB-current are 11/45 and 17/45 classes, respectively, alongside the 17/59 and 29/59 cell counts.

All 16 BHO-minus-PSE/SE primary intervals are below zero in the eight original reward-independent baseline-completion cells. Across all 25 reward-independent environments, intervals favor BHO over PSE in 20 cells and PSE in four, with one including zero; against SE they favor BHO in 14 and SE in ten, with one including zero. Other methods can be better; BHO-half's descriptive mean is lower in 57 cells and equal in the two negative controls.

Under Z, BHO beats PSE and SE in all 24 environments (all 48 primary intervals below zero), but these environments violate reward independence and each comparator eliminates the optimum in all 768 environment--seed pairs. In Z19 (\(K=8\), failure-hiding, \(p=1/8\), delay 2048) the means are 8114.33 versus 64740.35/65349.12. BHO itself can still be poor: in Z53 (\(K=64\), failure-hiding, \(p=3/16\), delay 2048) it incurs 64503.19 with only 1.576\% optimal pulls, and in Z58 (\(K=64\), failure-hiding, \(p=1/8\), delay 65536) it reaches uniform regret 64512, with optimal fraction 1/64.

SE has lower mean total regret than BHO in 10 of the 25 reward-independent environments, with BHO/SE ratios 1.19--2.86: Z05, Z07--Z09, Z33 and Z36--Z40 (environment key in Table O.1). These include eight of the nine reward-independent environments with bad-arm delay at least 8192. The PSE/SE adverse delay-scan cells include Z07--Z09 (\(K=8\), immediate all-report optimum, \(p=0\), delays 8192, 32768 and 65536) and Z36 (\(K=64\), immediate all-report optimum, \(p=0\), delay 32768). Signed-delay estimates include 20 negative and 10 zero values; eight secondary intervals are wholly negative and 36 include zero.

\end{multicols}
{\small
\begin{longtable}{>{\raggedright\arraybackslash}p{30.95pt}>{\raggedright\arraybackslash}p{121.87pt}>{\raggedright\arraybackslash}p{121.99pt}>{\raggedright\arraybackslash}p{165.01pt}}
\multicolumn{4}{@{}p{\linewidth}@{}}{\textbf{Table O.3. Selected primary intervals with favorable and unfavorable results.} All intervals use the same nominal Bonferroni-t calibration with p=118; the complete family is supplied in the data archive.}\\[3pt]
\toprule
\textbf{Cell} & \textbf{Contrast} & \textbf{Mean difference} & \textbf{Primary 95\% interval} \\
\midrule
\endfirsthead
\multicolumn{4}{@{}l}{\footnotesize\itshape Continued from previous page}\\
\toprule
\textbf{Cell} & \textbf{Contrast} & \textbf{Mean difference} & \textbf{Primary 95\% interval} \\
\midrule
\endhead
\midrule\multicolumn{4}{r@{}}{\footnotesize\itshape Continued on next page}\\
\endfoot
\bottomrule
\endlastfoot
Z19 & BHO-minus-PSE & -56626.02 & {[}-56856.66, -56395.39{]} \\
Z19 & BHO-minus-SE & -57234.80 & {[}-57464.69, -57004.90{]} \\
Z53 & BHO-minus-PSE & -224.42 & {[}-243.69, -205.16{]} \\
Z53 & BHO-minus-SE & -895.49 & {[}-903.80, -887.19{]} \\
Z58 & BHO-minus-PSE & -225.05 & {[}-243.53, -206.56{]} \\
Z58 & BHO-minus-SE & -717.27 & {[}-721.31, -713.22{]} \\
Z09 & BHO-minus-PSE & 23169.09 & {[}22059.09, 24279.09{]} \\
Z09 & BHO-minus-SE & 28993.02 & {[}27923.79, 30062.24{]} \\
Z36 & BHO-minus-PSE & 2072.78 & {[}1462.63, 2682.93{]} \\
Z36 & BHO-minus-SE & 29246.70 & {[}27660.77, 30832.64{]} \\
\end{longtable}
}
{\small
\begin{longtable}{>{\raggedright\arraybackslash}p{156.46pt}>{\raggedright\arraybackslash}p{32.85pt}>{\raggedright\arraybackslash}p{59.56pt}>{\raggedright\arraybackslash}p{91.57pt}>{\raggedright\arraybackslash}p{87.37pt}}
\multicolumn{5}{@{}p{\linewidth}@{}}{\textbf{Table O.4. Descriptive total-regret comparisons with G, OUCB and OPSE.} Counts of cells with lower/equal unrounded BHO mean over 32 seeds; the remainder have higher BHO means. Optimum-reporting groups cover all 59 environments. The fixed-optimum group differs from the 25-cell reward-independent subset because some bad-arm laws depend on reward.}\\[3pt]
\toprule
\textbf{Optimum reporting law} & \textbf{Cells} & \textbf{Versus G} & \textbf{Versus OUCB} & \textbf{Versus OPSE} \\
\midrule
\endfirsthead
\multicolumn{5}{@{}l}{\footnotesize\itshape Continued from previous page}\\
\toprule
\textbf{Optimum reporting law} & \textbf{Cells} & \textbf{Versus G} & \textbf{Versus OUCB} & \textbf{Versus OPSE} \\
\midrule
\endhead
\midrule\multicolumn{5}{r@{}}{\footnotesize\itshape Continued on next page}\\
\endfoot
\bottomrule
\endlastfoot
All outcomes, fixed delay & 25 & 8 / 0 & 11 / 0 & 25 / 0 \\
Never reports & 8 & 0 / 0 & 4 / 0 & 8 / 0 \\
Only zeros report & 24 & 9 / 0 & 14 / 0 & 23 / 1 \\
Only successes report & 2 & 0 / 2 & 0 / 2 & 0 / 2 \\
\end{longtable}
}
\begin{multicols}{2}
\subsection*{O.6 Bonferroni-t versus bootstrap signs}
The calibration dimensions in O.2 remain \(p=118\) for total regret and \(p=140\) for delay excess, although there are respectively 115 and 130 distinct observed contrast vectors. The primary and sensitivity signs are:

\end{multicols}
{\small
\begin{longtable}{>{\raggedright\arraybackslash}p{182.05pt}>{\raggedright\arraybackslash}p{41.80pt}>{\raggedright\arraybackslash}p{37.26pt}>{\raggedright\arraybackslash}p{63.78pt}>{\raggedright\arraybackslash}p{102.92pt}}

\toprule
\textbf{Family and calibration} & \textbf{Negative} & \textbf{Positive} & \textbf{Includes zero} & \textbf{Zero sample variance} \\
\midrule
\endfirsthead
\multicolumn{5}{@{}l}{\footnotesize\itshape Continued from previous page}\\
\toprule
\textbf{Family and calibration} & \textbf{Negative} & \textbf{Positive} & \textbf{Includes zero} & \textbf{Zero sample variance} \\
\midrule
\endhead
\midrule\multicolumn{5}{r@{}}{\footnotesize\itshape Continued on next page}\\
\endfoot
\bottomrule
\endlastfoot
Total regret, Bonferroni-t primary & 96 & 18 & 4 & 2 \\
Total regret, fixed-SE max-z sensitivity & 96 & 19 & 3 & 2 \\
Delay excess, Bonferroni-t primary & 8 & 96 & 36 & 20 \\
Delay excess, fixed-SE max-z sensitivity & 9 & 97 & 34 & 20 \\
\end{longtable}
}
\begin{multicols}{2}
The counts include flagged zero-variance rows, which remain descriptive and do not establish equivalence. For delay excess, ten such rows have positive degenerate endpoints and ten are all zero; among the 120 positive-SE rows, primary signs are 8 negative, 86 positive and 26 zero-containing. For total regret, negative means lower BHO regret. The Z07 BHO-minus-PSE primary interval is {[}−52.88, 753.57{]}, versus {[}15.49, 685.20{]} in sensitivity. Two delay-excess primary intervals contain zero although their sensitivity intervals do not: BHO in Z19 and SE in Z50.

\subsection*{O.7 Returned-UCB}
Returned-UCB follows App. T.3 (returned mean plus \(\sqrt{2\log(\max\{2,t\})/n}\), infinite index at returned count \(n=0\), smaller-label ties). The 59 environments, seeds 0--31 and \(T=262144\) yield 1,888 endpoints. The 59 paired BHO-minus-returned-UCB contrasts use nominal Bonferroni-\(t_{31}\) intervals at \(p=59\) (critical value 3.694359).

All 59 intervals exclude zero: BHO is lower in 40 environments and returned-UCB in 19. Of the 25 environments whose optima report every outcome, returned-UCB is lower in 11 and BHO in 14. Returned-UCB reaches or exceeds uniform regret in Z08, Z09, Z35, Z36 and Z38--Z40, all with bad-arm delay at least 8192. It is lower in all eight never-report environments: no optimal-arm sample arrives, so its index stays infinite and the arm receives 96.14\%--99.29\% of pulls on average. The zero-count rule produces this allocation without a mean estimate. BHO is lower in all 24 zeros-only environments and both negative controls, where returned-UCB has 100.52\%--114.28\% of uniform regret. In the 25 reward-independent environments, BHO is lower in 12 and returned-UCB in 13.

\end{multicols}
{\small
\begin{longtable}{>{\raggedright\arraybackslash}p{21.12pt}>{\raggedright\arraybackslash}p{82.84pt}>{\raggedright\arraybackslash}p{96.27pt}>{\raggedright\arraybackslash}p{116.04pt}>{\raggedright\arraybackslash}p{111.55pt}}
\multicolumn{5}{@{}p{\linewidth}@{}}{\textbf{Table O.5. Selected returned-UCB endpoints and paired contrasts.} Mean pseudo-regret (seed SE) and percentage of uniform regret. The final columns give BHO minus returned-UCB, mean (paired seed SE), and the nominal 95\% interval in the full 59-contrast family. Positive differences favor returned-UCB. The rows cover three all-report cells, never-report cells at both arm counts, and zeros-only cells including Z53 and Z58. The data archive provides all 59 rows.}\\[3pt]
\toprule
\textbf{Cell} & \textbf{Returned-UCB} & \textbf{Percent of uniform} & \textbf{BHO − returned-UCB} & \textbf{Primary interval} \\
\midrule
\endfirsthead
\multicolumn{5}{@{}l}{\footnotesize\itshape Continued from previous page}\\
\toprule
\textbf{Cell} & \textbf{Returned-UCB} & \textbf{Percent of uniform} & \textbf{BHO − returned-UCB} & \textbf{Primary interval} \\
\midrule
\endhead
\midrule\multicolumn{5}{r@{}}{\footnotesize\itshape Continued on next page}\\
\endfoot
\bottomrule
\endlastfoot
Z01 & 646.47 (9.60) & 1.13 & 1461.89 (16.68) & {[}1400.26, 1523.53{]} \\
Z09 & 65535.20 (0.03) & 114.28 & -20914.85 (281.26) & {[}-21953.93, -19875.77{]} \\
Z12 & 464.91 (26.76) & 0.81 & 940.62 (27.12) & {[}840.42, 1040.82{]} \\
Z19 & 65467.12 (1.21) & 114.17 & -57352.79 (57.96) & {[}-57566.92, -57138.66{]} \\
Z31 & 32260.34 (17.65) & 50.01 & -12588.10 (61.71) & {[}-12816.09, -12360.11{]} \\
Z43 & 504.98 (15.80) & 0.78 & 6525.47 (21.89) & {[}6444.58, 6606.36{]} \\
Z53 & 65487.67 (0.94) & 101.51 & -984.48 (2.66) & {[}-994.32, -974.65{]} \\
Z58 & 65535.53 (0.09) & 101.59 & -1023.53 (0.09) & {[}-1023.86, -1023.20{]} \\
\end{longtable}
}
\begin{multicols}{2}
\pdfbookmark[0]{Appendix P. Fixed Synthetic Cohorts and Sensitivity Checks}{r060.nav.19}
\section*{Appendix P. Fixed Synthetic Cohorts and Sensitivity Checks}
\subsection*{P.1 Cohort identities and learner information}
Population B (Figure 2) uses seeds 4096--4127; population C reuses B's five all-report environments and seed vectors. Learners, information and reporting are as in O.1; BHO and BHO-half use \(\theta=1\) and \(\theta=1/2\) with \(\ell=\log(4KT(T+1)^2)\).

Population A reruns BHO on Z31 with further seeds. With the generator and tie rule used for the original seeds, its 256 seeds 32--287 give mean regret 19822.25 (SE 18.46), versus 19672.24 (SE 56.95) for the original seeds 0--31. The difference, 150.01, is 0.76\% of the original mean and 2.506 estimated independent joint SEs (two-sided normal p=0.0122; \(t_{31}\) p=0.0177). Replaying the archived implementation on the 32 original seeds reproduces every endpoint and all 8,388,608 normalized action/reward entries. We take the 256-seed mean as the more precise estimate of BHO's regret on Z31 and retain the original batch, which is paired with the other methods in App. O, for comparison.

\subsection*{P.2 Longer-horizon population and paired comparisons}
Population B fixes \(K=32,T=1048576,\mu_*=0.9,\mu_i=0.1\) and 32 paired seeds 4096--4127. The best arm either reports every outcome immediately or reports zeros immediately and permanently hides successes. For each law there is one shared \(D=0\) reference, and four \(D=4096\) cells with \(k_{\rm slow}=1,8,16,31\) bad logical arms delayed; remaining bad arms report immediately. No bad-arm outcome is permanently missing. The two zero-delay references are each evaluated once per method/seed.

The endpoint \(R_T/[T(31/32)(0.8)]\) normalizes by uniform expected regret 812646.4. The common 90\% threshold is a finite-horizon diagnostic. BHO spans 0.43797\%--1.01815\%; every completion-based cell stays below 90\%. OPSE's five zeros-only stress cells equal 100\% at every seed (zero sample variance); its largest all-report mean is 1.87992\%.

The family has 35 coordinates: BHO minus G, OUCB-current and BHO-half in ten cells, and BHO minus OPSE in the five all-report cells. By the coupling in P.3 only 20 are distinct, and all 35 paired standard errors are positive. Under O.2's protocol, \(p=20\) gives Bonferroni-\(t_{31}\) critical value 3.290276 for 32 paired seeds; bootstrap sensitivity uses 20,000 whole-seed-vector resamples and critical value 2.876627. The family was specified after the results of Appendix O were known but before the population-B runs. The analysis uses every coordinate in this family.

From \(k_{\rm slow}=1\) to 31, BHO's mean rises 2760.275 versus G's 4602.95, yet BHO stays above G at every tested count, including its higher zero-delay cost.

All 15 distinct BHO-minus-G/OUCB-current/BHO-half intervals lie above zero and all five all-report BHO-minus-OPSE intervals lie below zero. Figure 2b displays the largest slow-arm-count case. The data archive supplies all 50 endpoint summaries and the full paired contrast family.

In population C, OUCB-common and OPSE-common replace the radius \(r_i^2=2\log(T)/m_i\) by \(r_i^2=\ell/(2m_i)\), with all other rules fixed; neither variant inherits a theorem from its original coefficient. OUCB-common's mean regret stays below BHO's in all five all-report cells, and OPSE-common improves in all five, matching BHO at zero delay but remaining higher at positive delays.

For each of the 20 distinct contrasts, the bootstrap sensitivity interval has the same sign as the primary interval.

\subsection*{P.3 Coupling and independent sampling units}
By O.5, the two laws have identical failure-arrival processes and hence identical completion-policy actions. OPSE's lower bound uses \(G_i/m_i\), so its trajectory can differ.

For each of the 640 run pairs (four methods × five environments × 32 seeds), all 1048576 action/reward entries coincide after removing the permanent-missingness flag. The bootstrap resamples whole seeds, keeping the covariance across all 35 coordinates; the two laws share the same 32 seeds, which are the independent units.

\pdfbookmark[0]{Appendix Q. Calendar-Rate Blocks}{r060.nav.20}
\section*{Appendix Q. Calendar-Rate Blocks}
\textbf{Algorithm 2 (calendar-rate blocks).} Set
\begin{equation}
\begin{gathered}
\ell_{\rm rate}=\log\bigl(2KT(T+1)^3\bigr),\qquad\\
\beta_{\rm rate}(n)=\sqrt{2\ell_{\rm rate}/n}\\
+\frac{2\ell_{\rm rate}}{3n}.
\end{gathered}
\tag{Q1}
\end{equation}
Initialize \(H_i^{\rm rate}=1\) and block counters \(k_i=0\). At a block boundary choose the smallest label maximizing \(H_i^{\rm rate}\). Set \(n=2^{k_i}\), increment \(k_i\), and pull that arm exclusively for \(2n\) rounds, stopping at the horizon. If the block completes, let \(W\) be its second half and let \(\zeta_i(W)\) sum \(1-y\) over all arm-\(i\) reports arriving in \(W\), retaining multiplicity and including reports from earlier blocks. After processing its final round's reports, update
\begin{equation}
H_i^{\rm rate}\leftarrow\min\left\{ H_i^{\rm rate},1,1-\frac{\zeta_i(W)}n+\beta_{\rm rate}(n)\right\}.
\tag{Q2}
\end{equation}
Other historical indices stay fixed. Each arm's block lengths double independently across its visits. A final incomplete block needs no update. The learner uses \(K,T\), its clock, actions and reports; gaps and cutoffs are analysis parameters.

We prove Proposition 1 in a form permitting heterogeneous residuals. Fix one deterministic finite cutoff \(h_i\in\mathbb N_0\) and allowance \(q_i^{\rm rate}\) for each bad arm, with
\[
\begin{gathered}
0\le b_i(h_i)\le q_i^{\rm rate}<\Delta_i,\\
\qquad \gamma_i^{\rm rate}=\Delta_i-q_i^{\rm rate}>0.
\end{gathered}
\]
Write \(\delta_T=\delta=(T+1)^{-2}\), so \(\ell_{\rm rate}=\log(2KT(T+1)/\delta_T)\). We show, with probability at least \(1-\delta_T\),
\begin{equation}
\begin{gathered}
\Psi_T^{\rm rate}\le\\
\sum_{i\in\mathcal I}\left( 8\Delta_i h_i+ \frac{1024\ell_{\rm rate}\Delta_i}{(\gamma_i^{\rm rate})^2} \right).
\end{gathered}
\tag{Q3}
\end{equation}
We use the original calendar-round potential array from Section 2.1. For a deterministic integer window \(W=[a,b]\subseteq[1,T]\) of length \(n\), define
\begin{equation}
\begin{gathered}
X_i(W)=\sum_{s=1}^b(1-Y_s(i))\\
\mathbf1\{s+D_s(i)\in W\}.
\end{gathered}
\tag{Q4}
\end{equation}
For fixed \(i,W\), the summands are independent across origin rounds and lie in \([0,1]\), despite within-round reward--delay or cross-arm dependence. Inserting \(\mathbf1\{A_s=i\}\) gives the observed sum, so \(\zeta_i(W)\le X_i(W)\) pathwise for every adaptive action sequence, including reports from earlier blocks. The sums can exceed \(n\); concentration uses their mean and variance. Unselected potential variables supply no reports and serve only this analysis, which does not assert conditional independence of selected outcomes given observed history.

Put \(p_i(d)=\mathbb E[(1-Y_i)\mathbf1\{D_i=d\}]\) for finite \(d\). Each delay has at most \(n\) positive origins arriving in \(W\), whence
\begin{equation}
\begin{gathered}
\mathbb E X_i(W)\\
=\sum_{d\ge0}p_i(d)\#\{s\ge1:s+d\in W\}\\
\le n(1-\mu_i).
\end{gathered}
\tag{Q5}
\end{equation}
Infinite delays contribute nothing. Missing origins before round one only reduce the mean, so (Q5) also covers startup. Since each \([0,1]\) summand has variance at most its mean, total variance is at most \(n\). Bernstein's inequality gives, with failure probability \(e^{-\ell_{\rm rate}}\),
\begin{equation}
\begin{gathered}
X_i(W)\le n(1-\mu_i)\\
+\sqrt{2n\ell_{\rm rate}}+\frac23\ell_{\rm rate}.
\end{gathered}
\tag{Q6}
\end{equation}
The threshold \(t=\sqrt{2n\ell_{\rm rate}}+2\ell_{\rm rate}/3\) satisfies \(t^2\ge2\ell_{\rm rate}(n+t/3)\), giving Bernstein exponent at least \(\ell_{\rm rate}\). By domination, (Q2) is at least \(\mu_i\); initialization and historical minima preserve this for every arm. This also covers an optimum whose delay is always infinite: its observed deficit sum is zero and its historical index remains one.

For a bad arm also define \(X_i^{\le h_i}(W)\) by inserting \(\mathbf1\{D_s(i)\le h_i\}\) in (Q4). If \(a\ge h_i+1\), each \(d\le h_i\) has exactly the \(n\) positive origins \(a-d,\ldots,b-d\). Therefore
\begin{equation}
\begin{gathered}
\mathbb E X_i^{\le h_i}(W)\\
=n\bigl(1-\mu_i-b_i(h_i)\bigr),\\
\qquad\\
X_i^{\le h_i}(W)\ge\\
\mathbb E X_i^{\le h_i}(W)-\sqrt{2n\ell_{\rm rate}}
\end{gathered}
\tag{Q7}
\end{equation}
except with probability \(e^{-\ell_{\rm rate}}\). To obtain the lower tail without a linear remainder, let \(S\) be this sum and \(m=\mathbb ES\le n\). Convexity on \([0,1]\), independence and \(\log(1+x)\le x\) give
\[
\begin{gathered}
\log\mathbb E e^{-\lambda(S-m)}\\
\le m(e^{-\lambda}-1+\lambda)\\
\le n\lambda^2/2\qquad(\lambda\ge0).
\end{gathered}
\]
Markov's inequality at \(\lambda=\sqrt{2\ell_{\rm rate}/n}\) gives (Q7), including when its lower threshold is negative.

We apply a union bound to (Q6) over all arms and \(T(T+1)/2\) deterministic windows, and to (Q7) over bad arms and windows satisfying its boundary condition, for the fixed cutoff vector. At most \(KT(T+1)\) events fail, with total probability at most \(\delta_T/2\le\delta_T\). The simultaneous event applies to adaptive second-half windows without conditioning on their random starts. Cutoffs are fixed before the union; events are not intersected over cutoff vectors. The learner is independent of this choice, so the expectation bound holds for each vector.

Consider a completed arm-\(i\) block starting at round \(u\), with second half \([u+n,u+2n-1]\) and \(n\ge h_i\), so the lower-mean boundary condition holds. Every delay-at-most-\(h_i\) origin in this half lies between \(u+n-h_i\ge u\) and \(u+2n-1\), when arm \(i\) is selected. Hence \(\zeta_i(W)\ge X_i^{\le h_i}(W)\). Other reports add nonnegative deficits, and the last round's zero-delay reports are processed before updating. Equation (Q7) gives
\begin{equation}
\begin{gathered}
1-\zeta_i(W)/n+\beta_{\rm rate}(n)\\
\le\mu_i+b_i(h_i)\\
+2\sqrt{2\ell_{\rm rate}/n}\\
+\frac{2\ell_{\rm rate}}{3n}.
\end{gathered}
\tag{Q8}
\end{equation}
If \(n\ge128\ell_{\rm rate}/(\gamma_i^{\rm rate})^2\), the last two terms are at most
\begin{equation}
\begin{gathered}
\gamma_i^{\rm rate}/4+(\gamma_i^{\rm rate})^2/192\\
\le49\gamma_i^{\rm rate}/192<\gamma_i^{\rm rate}.
\end{gathered}
\tag{Q9}
\end{equation}
The updated historical index is strictly below \(\mu^*\), whereas every optimal arm's historical index stays at least \(\mu^*\). Historical minima and the maximizing choice prevent any later block of this bad arm.

Let \(n_i^*\) be the first power of two at least \(a_i^{\rm rate}:=\max\{h_i,128\ell_{\rm rate}/(\gamma_i^{\rm rate})^2\}\). Since \(a_i^{\rm rate}>1\), we have \(n_i^*<2a_i^{\rm rate}\). The total number of pulls of this arm through its threshold block is
\begin{equation}
2(1+2+\cdots+n_i^*)=4n_i^*-2<8a_i^{\rm rate}.
\tag{Q10}
\end{equation}
Horizon truncation only reduces this per-arm count; intervening blocks of other arms add no pulls to it. Multiply by \(\Delta_i\), use \(\max\{x,y\}\le x+y\), and sum to obtain (Q3). Failure costs at most \(T\). Taking \(q_i^{\rm rate}=c\Delta_i\) gives \(\gamma_i^{\rm rate}=\alpha\Delta_i\) and (9) for either cutoff choice. Cutoffs \(h_i=0\) obey the same geometry; absent an admissible finite cutoff, only the trivial \(T\) bound remains.

\pdfbookmark[0]{Appendix R. One Cohort Learner with Both Cutoff Guarantees}{r060.nav.21}
\section*{Appendix R. One Cohort Learner with Both Cutoff Guarantees}
\subsection*{R.1 Why the two cohort charges coexist}
The proof bounds the same learner's predictable regret in two ways on one event. A per-member block cap gives the cohort-weighted sum bound (R.4); ancestry controls old arrivals (R.3), and minimum-exposure selection gives a shared-interval bound (R.5--R.8). Taking expectations preserves both branches and incurs one failure term.

\subsection*{R.2 Policy and analysis quantities}
We use Section 2.1's joint-origin model. Private action coins are fresh and independent of the potential vectors. Set \(\delta=(T+1)^{-2}\), \(J=1+\lfloor\log_2(T+1)\rfloor\), \(\ell_C=\log(8K^2(T+1)^3J)>1\), and \(\varepsilon(x)=\min\{1,\sqrt{\ell_C/(2x)}\}\). These are the exact NC-lean parameters.

\textbf{Algorithm 3 (NC-lean).} Initially one cohort contains every arm, with exposure one, counts \(N_i=0\), cumulative reported deficits \(A_i=0\), and scores \(H_i=1\). At each boundary select the eligible cohort minimizing \((x_C,-M_C,\min C)\), where \(M_C=\max_{i\in C}H_i\), \(M=\max_iH_i\), and eligibility means \(M_C+\varepsilon(x_C)\ge M\). Retain \(S=\{i\in C:H_i\ge M_C-\varepsilon(x_C)\}\). If this is a proper split, both children inherit \(x_C\); select \(S\). With \(m=|S|\), freeze independent uniform draws from \(S\) for \(2mx\) rounds. After every action, process all arrivals, including delay zero, and update every positive-count arm by
\begin{equation}
H_i\leftarrow\min\{H_i,1,1-A_i/N_i+\sqrt{\ell_C/(2N_i)}\}.
\tag{R1}
\end{equation}
At full completion, for the actual second-half arrival window \(W\), also update each member by
\begin{equation}
H_i\leftarrow\min\{H_i,1,1-\zeta_i(W)/x+\sqrt{2\ell_C/x}+2\ell_C/(3x)\},
\tag{R2}
\end{equation}
then double its cohort exposure for every member, including members with zero realized draws. Here \(\zeta_i(W)\) sums deficits of \textbf{all} \(i\)-labeled reports arriving in \(W\), including old reports. A horizon-truncated block receives no calendar update or doubling. Cohorts never merge; arms are never deleted.

Fix finite integer cutoffs \(h_i\ge0\) and allowances \(0\le q_i<\Delta_i\), with \(b_i(h_i)\le q_i\), for bad arms only. Define
\[
\begin{gathered}
Q_i=\mathbb E[(1-Y_i)\mathbf1\{D_i<\infty\}],\\
\nu_i=1-Q_i,\\
\nu^*=\max_i\nu_i,\\
\Gamma_i=\nu^*-\mu_i-q_i,\\
\alpha=\min_{i\in\mathcal I}(\Delta_i-q_i)/\Delta_i,\\
Z_{\rm eff}=\sum_{i\in\mathcal I}\Delta_i/\Gamma_i^2,\\
D=\max_{i\in\mathcal I}h_i,\\
f_i=1-\mu_i-b_i(h_i).
\end{gathered}
\]
Then \(\nu^*\ge\mu^*\), \(0<\Gamma_i\le f_i\le1\), and \(\Gamma_i\ge\alpha\Delta_i\). No cutoff is imposed on an optimum. Let \(\mathcal F_t\) be the analysis filtration generated by all complete origin vectors and private action coins through round \(t\). For action \(A_s\), write \(p_i(s)=\Pr(A_s=i\mid\mathcal F_{s-1})\). This probability is \(1/|S|\) on the selected block group \(S\) and zero off it. Set \(B_i(t)=\sum_{s=1}^t p_i(s)\) and \(P_T=\sum_{i\in\mathcal I}\Delta_iB_i(T)\). Let \(M_i,x_i\) be the size and exposure of the last selected group containing \(i\) whose block began, including truncation, and set \(C_{\rm cap}=\sum_{i\in\mathcal I}\Delta_i\min\{4x_i,8h_i/M_i\}\). The first selected group contains all arms, so this metadata exists even for members never drawn. We prove, on one event of failure at most \(\delta/2\),
\begin{equation}
\begin{gathered}
P_T\le128\ell_CZ_{\rm eff}+C_{\rm cap},\qquad\\
P_T\le524288\ell_CZ_{\rm eff}+6420\Phi_T(D)/\alpha.
\end{gathered}
\tag{R3}
\end{equation}
Times below count completed rounds, and a block starting after \(t\) rounds uses origins \(t+1,\ldots,t+2mx\).

\subsection*{R.3 One confidence event for the actual rule}
The policy uses only visible history. Its current block group is determined by the past, and the fresh action coin and origin vector are independent of \(\mathcal F_{s-1}\) and of each other. For a fixed arm, let \(S_i(t)\) sum a quantity \(U\in[0,1]\) of mean \(u\) over its pulls through round \(t\), with each \(U\) computed from the corresponding fresh origin. Conditional Hoeffding gives the supermartingale
\[
\exp\{\lambda(S_i-uN_i)-\lambda^2N_i/8\}.
\]
At each fixed own count \(n\), take \(\lambda=\sqrt{8\ell_C/n}\) and apply the maximal inequality. Union over \(i,n\le T\), once for finite deficits and once with negative sign for cutoff-truncated deficits, costs \(2KT e^{-\ell_C}\). Simultaneously,
\begin{equation}
\begin{gathered}
S_i^{\rm fin}(t)\le Q_iN_i(t)+\sqrt{N_i(t)\ell_C/2},\qquad\\
S_i^h(t)\ge f_iN_i(t)-\sqrt{N_i(t)\ell_C/2}.
\end{gathered}
\tag{R4}
\end{equation}
Zero-count statements are deterministic. Actual deficits satisfy \(A_i(t)\le S_i^{\rm fin}(t)\) and \(A_i(t)\ge S_i^h((t-h_i)_+)\). Thus (R1), including historical minima, gives at every positive-count time
\begin{equation}
\begin{gathered}
H_i\ge\nu_i,\qquad\\
H_i(t)\le\mu_i+b_i(h_i)+f_iR_i(t)/N_i(t)\\
+\sqrt{2\ell_C/N_i(t)},\quad\\
R_i(t)=N_i(t)-N_i((t-h_i)_+).
\end{gathered}
\tag{R5}
\end{equation}
The upper bound sums the two prefix errors and uses \(N_i(t-h_i)\le N_i(t)\). It holds during dormancy and inside frozen blocks.

For action counts, \(\exp\{-\lambda N_i+(1-e^{-\lambda})B_i\}\) is a supermartingale. For deterministic \(v>0\), the event \(B_i=v\), \(N_i<v-\sqrt{2v\ell_C}\) implies a crossing of \(e^{\ell_C}\), using \(\lambda=\sqrt{2\ell_C/v}\) and \(1-e^{-\lambda}\ge\lambda-\lambda^2/2\). At boundaries, let \(\xi_i\) denote the current cohort's next exposure. Geometric completed memberships give \(B_i=2\xi_i-2\), even after splits and dormancy. There are at most \(J\) possible dyadic levels. A \(KJ e^{-\ell_C}\) union yields
\begin{equation}
N_i\ge2\xi_i-2-2\sqrt{\xi_i\ell_C},\qquad N_i\ge\xi_i\quad(\xi_i>8\ell_C).
\tag{R6}
\end{equation}
For the latter, \(2\sqrt{\xi_i\ell_C}<3\xi_i/4\) and \(\xi_i>8\). The \(\xi_i=1\) level has \(N_i=B_i=0\).

The calendar candidates are valid by the arrival compensator argument corresponding to Appendix Q's all-potential window bound, with a checked historical probability cap. Fix deterministic \(i,m,W\), where \(W=[u+mx,u+2mx-1]\), \(x\) is dyadic, and the full block fits in \(1,\ldots,T\). Its origin contributions \(X_s\in[0,1]\) have conditional means \(p_i(s)w_s\), where \(w_s=\mathbb E[(1-Y_i)\mathbf1\{s+D_i\in W\}]\). With \(V=\sum_s p_i(s)w_s\), conditional iteration gives
\[
\begin{gathered}
\mathbb E e^{\lambda(\zeta_i(W)-V)-\phi(\lambda)V}\le1,\\
\quad \phi(\lambda)\le\frac{\lambda^2}{2(1-\lambda/3)}.
\end{gathered}
\]
For \(t=\sqrt{2x\ell_C}+2\ell_C/3\), the choice \(\lambda=t/(x+t/3)<3\) gives exponent at least \(t^2/[2(x+t/3)]\ge \ell_C\). Hence \(\Pr\{\zeta_i(W)-V>t,V\le x\}\le e^{-\ell_C}\). Never-merge ancestry guarantees \(p_i(s)\le1/m\) at every earlier/current origin of an actual size-\(m\) block: earlier positive probabilities came from ancestor cohorts of size at least \(m\), and other probabilities were zero. On this path, \(V\le xQ_i\le x\); a violation of \(\zeta_i(W)\le xQ_i+t\) is contained in that joint event. Union over the deterministic metadata costs at most
\[
\begin{gathered}
K\sum_{m=1}^K\sum_{k:2m2^k\le T}(T-2m2^k+1)\\
\le K^2TJ
\end{gathered}
\]
events. Thus actual (R2) candidates are at least \(\nu_i\), without conditioning on realized windows or discarding old arrivals.

For contraction, condition at a full block's measurable start on its chosen set and exposure. If \(mx\ge h_i\), every fixed delay \(d\le h_i\) has exactly \(mx\) shifted origins hitting the second half, all inside the frozen block. The joint reward-delay law is used in their mean; no independent-delay premise is needed. Independent future origin contributions therefore have truncated mean \(xf_i\) and give \(\zeta_i(W)\ge xf_i-\sqrt{2x\ell_C}\) with conditional failure at most \(e^{-\ell_C}\): the centered lower exponential has log moment at most \(x\lambda^2/2\), and \(\lambda=\sqrt{2\ell_C/x}\) gives the claim. Other reports only add nonnegative deficits. The total membership count over full blocks is at most \(\lfloor T/2\rfloor\), since each occupies \(2mx\ge2m\) disjoint rounds. Conditional union and expectation consequently cost at most \(\lfloor T/2\rfloor e^{-\ell_C}\), giving
\begin{equation}
\begin{gathered}
H_i({\rm end})\le\mu_i+b_i(h_i)+e(x),\qquad\\
e(x)=2\sqrt{2\ell_C/x}+2\ell_C/(3x).
\end{gathered}
\tag{R7}
\end{equation}
The intersection \(\mathcal G\) has failure at most
\begin{equation}
\begin{gathered}
[K^2TJ+2KT+KJ+\lfloor T/2\rfloor]e^{-\ell_C}\\
\le4K^2(T+1)Je^{-\ell_C}=\delta/2.
\end{gathered}
\tag{R8}
\end{equation}
This event is for the fixed learner, law and cutoff vector. Delay endpoints \(d=0,h_i=mx\), startup, completion exactly at \(T\), and a terminal partial block follow the stated inclusive windows.

\subsection*{R.4 The cohort cap}
On \(\mathcal G\), selected members at their start satisfy
\begin{equation}
\begin{gathered}
H_i\ge M_C-\varepsilon(x)\ge M-2\varepsilon(x)\\
\ge\nu^*-2\varepsilon(x).
\end{gathered}
\tag{R9}
\end{equation}
Suppose a completed block has \(mx\ge h_i\) and \(x\ge16\ell_C/\Gamma_i^2\). By (R7), \(H_i({\rm end})\le\nu^*-\Gamma_i+e(x)\). Every later own exposure is at least \(2x\), but
\begin{equation}
\begin{gathered}
e(x)+2\varepsilon(2x)\\
\le(1/\sqrt2+1/24+1/4)\Gamma_i<\Gamma_i.
\end{gathered}
\tag{R10}
\end{equation}
Strictness follows from \(1/\sqrt2<17/24\). Historical monotonicity and (R9) therefore forbid every later membership.

Own exposures double, so \(B_i(T)<4x_i\). If \(x_i>1\), the preceding block completed at \(x_i/2\) with size at least \(M_i\). It cannot meet both thresholds in (R10). Thus \(B_i(T)<8\max\{h_i/M_i,16\ell_C/\Gamma_i^2\}\). The same bound covers \(x_i=1\) because \(\ell_C>1\). Combining both inequalities gives
\begin{equation}
B_i(T)\le128\ell_C/\Gamma_i^2+\min\{4x_i,8h_i/M_i\}.
\tag{R11}
\end{equation}
This proves the first branch.

\subsection*{R.5 Maturity and the frozen endpoint}
Set \(A=2^{17}\). Charge an individual bad member's block contribution as mature when \(x\ge A\ell_C/\Gamma_i^2\). A mixed cohort can contain both types. The geometric sum of all immature own exposures is less than \(4A\ell_C/\Gamma_i^2\) per arm, charging \(524288\ell_CZ_{\rm eff}\) once over the run.

At a mature start, (R6) gives \(N_i\ge x\), and both \(2\varepsilon(x)\) and \(\sqrt{2\ell_C/N_i}\) are at most \(\Gamma_i/256\). Equations (R5),(R9) imply
\begin{equation}
f_iR_i/N_i\ge\Gamma_i/2,\qquad D\Gamma_i\ge A\ell_C/2.
\tag{R12}
\end{equation}
The preceding own block at \(x/2\) exceeds the statistical threshold 16. By (R10) its half-duration must be smaller than \(h_i\), so the current prescribed duration satisfies
\begin{equation}
mx<2D,\qquad 2mx<4D.
\tag{R13}
\end{equation}
This bounds the cost of a final partial block without assuming its interior actions remain eligible. If \(D=0\), no contribution is mature. If \(D\ge T\), the trivial \(P_T\le\Delta_{\max}T\le D\) already proves the maximum branch. Hence assume \(1\le D<T\).

\subsection*{R.6 A shared calendar interval}
The count-growth argument of Appendix A applies verbatim: with \(N_j,R_j\) in place of \(m_j,B_j\), (A9)--(A10) give the bound used below.

\subsection*{R.7 High and low charges}
Write \(v=M-\nu^*\in[0,1)\); a nonempty bad set implies \(\mu^*>0\). A mature bad member is high when \(v\ge\Gamma_i/16\). Partition each dyadic optimism band \([a,2a)\) into 20 half-open bins \([b,b+a/20)\). In one nonempty bin, let \(t_0\) be its first charged start and \(s_j\) each charged arm's last start. Charged gaps satisfy \(\Gamma_j<32a\), so (R9) gives \(H_j(s_j)\ge\nu^*+b-a/8\), a bound transported to every earlier time by historical minima.

Let \(X_0\) be the largest dyadic at most \(16\ell_C/a^2\); then \(X_0>8\ell_C/a^2\). If \(j\)'s current cohort at \(t_0\) had \(\xi_j<X_0\), its tolerance would be at least \(a/(4\sqrt2)\). Clipping at one preserves this lower bound because \(a\le1\). Since
\[
1/(4\sqrt2)-1/8>1/20,
\]
that cohort would be eligible. The selected mature cohort has exposure at least \(A\ell_C/(32a)^2=128\ell_C/a^2>X_0\), contradicting minimum-exposure priority. Therefore (R6) gives \(N_j(t_0)>8\ell_C/a^2\). Throughout \([t_0,s_j]\), the cumulative error is below \(a/2\); (R5) yields \(f_jR_j/N_j\ge a/4\). We apply (A10) with \(N=8\ell_C/a^2\) and \(g=a/2\). Its ratio \(U/N\le aD/(2\ell_C)\le D\); adding the final block's \(<4D\) duration gives interval length at most \(D[5+4\log D/a]\). At each origin, only this bin's members are charged; their probabilities sum to at most one and their gaps are below \(32a/\alpha\). Hence its charge is at most
\begin{equation}
\begin{gathered}
(D/\alpha)[160a+128\log D]\\
\le160(D/\alpha)\log(eD).
\end{gathered}
\tag{R14}
\end{equation}
A remaining mature contribution is low, with \(v<\Gamma_i/16\). Partition effective gaps into dyadic \([\epsilon,2\epsilon)\), putting \(\Gamma_i=1\) in the \(\epsilon=1\) band. Its first charged start satisfies \(M(t_0)<\nu^*+\epsilon/8\); future last charged scores exceed \(\nu^*-\epsilon/128\). The same largest-dyadic threshold \(X_0\le16\ell_C/\epsilon^2\) and \(1/(4\sqrt2)-1/128>1/8\) force \(N_j(t_0)>8\ell_C/\epsilon^2\). The selected exposure is at least \(A\ell_C/(4\epsilon^2)>X_0\), so the priority contradiction is valid. At every intervening time, (R5) gives \(f_jR_j/N_j\ge\epsilon/4\). Equation (A10), with \(N=8\ell_C/\epsilon^2\) and \(g=\epsilon/2\), and one final overshoot give \(D[5+4\log D/\epsilon]\). Charging only these members, with gaps below \(2\epsilon/\alpha\), costs at most
\begin{equation}
\begin{gathered}
(D/\alpha)[10\epsilon+8\log D]\\
\le10(D/\alpha)\log(eD).
\end{gathered}
\tag{R15}
\end{equation}
Mixed memberships cause no duplication: each individual bad member is assigned its own start class and its own gap; unrelated members are excluded from that class's rate.

\subsection*{R.8 Finite scales and expectation}
The cases exhaust mature contributions. By (R12), a nonempty high scale has \(a>\Gamma_i/32\ge A\ell_C/(64D)>1/D\); a nonempty low scale has \(\epsilon>\Gamma_i/2\ge A\ell_C/(4D)>1/D\). Each family therefore has at most \(1+\lfloor\log_2D\rfloor\le2\log(eD)\) scales. Summing 20 copies of (R14) and one of (R15) per scale gives
\[
\begin{gathered}
P_T^{\rm mature}\le(20\cdot160+10)\,2D\log^2(eD)/\alpha\\
=6420\Phi_T(D)/\alpha.
\end{gathered}
\]
Adding the immature term, which is paid once, proves the second branch. The earlier \(D=0,D\ge T\) cases give the stated truncated-log convention.

Both branches hold on the same \(\mathcal G\). Fresh coins give \(\mathbb EN_i(T)=\mathbb EB_i(T)\) unconditionally. Consequently
\begin{equation}
\begin{gathered}
R_T\le\\
\mathbb E\left[\mathbf1_{\mathcal G}\min\left\{\begin{array}{c}\Delta_{\max}T,\\128\ell_CZ_{\rm eff}+C_{\rm cap},\\524288\ell_CZ_{\rm eff}+6420\Phi_T(D)/\alpha\end{array}\right\}\right]\\
+\delta\Delta_{\max}T.
\end{gathered}
\tag{R16}
\end{equation}
Random \(M_i,x_i\) belong to this learner, and their cap remains inside its event-restricted expectation. For Theorem 3 take \(q_i=c\Delta_i\) and \(h_i=w_i(c)\), so \(Z_{\rm eff}\le\alpha^{-2}\sum_{i\in\mathcal I}1/\Delta_i\). Since \(C_{\rm cap}\le8\sum_{i\in\mathcal I}\Delta_i h_i/M_i\), dropping the event indicator from this nonnegative quantity and using \(\Delta_{\max}\le1\) gives (10). The stronger exact expectation (R16) remains available for heterogeneous allowances.

\pdfbookmark[0]{Appendix S. Exploration Cost of the Cohort Learners}{r060.nav.22}
\section*{Appendix S. Exploration Cost of the Cohort Learners}
\textbf{BRB.} BRB initializes separate arms at \(x_i=1\), uses \(\ell_B=\log(8K^2(T+1)^6)\) and \(e_B(x)=\min\{1,2\sqrt{2\ell_B/x}+2\ell_B/(3x)\}\), and selects minimum \((x_i,-H_i,i)\) among \(H_i+e_B(x_i)\ge M\). It plays that arm for \(2x_i\) frozen rounds, with global cumulative minima (R1) using \(\ell_B\), and its own second-half update (R2) using \(\ell_B\). Full completion doubles \(x_i\); truncation does neither. BRB is the cohort architecture with singleton initialization and this wider tolerance.

This comparison reuses the environments used to design NC-lean: 128 NC-lean endpoints and 512 endpoints of NC-wide, BRB, BHO and RATE (Algorithm 2), that is, eight environments × two horizons × eight seeds per method. Within each environment, horizon and seed, all methods share the reward--delay inputs and label permutation; action randomness is method specific. NC-wide follows Algorithm 3 (initialization, minimum-exposure priority, parent-relative splits, frozen uniform blocks, all-arm cumulative updates) but uses \(\ell_{\rm wide}=\log(8K^2(T+1)^6)\) in both index updates and the tolerance \(\min\{1,2\sqrt{2\ell_{\rm wide}/x}+2\ell_{\rm wide}/(3x)\}\) for eligibility and splitting; NC-lean uses those of R.2.

All rewards are Bernoulli. E0 has four arms with means (0.9, 0.7, 0.4, 0.1) and immediate reports. E1 has eight arms with means (0.85, 0.65, 0.5, 0.35, 0.25, 0.2, 0.15, 0.1) and common delay 512. E2--E4 have one reward-one arm that never reports and bad arms of mean 1/4 that report only zero rewards: E2 uses K = 4 and bad-arm delay 2048, E3 uses K = 16 and delay 2048, and E4 uses K = 16 and immediate bad-arm reports. E5 has means (0.85, 0.65, 0.45, 0.25); the optimal arm never reports, and bad arms report zero rewards after delay 256. E6 has eight means decreasing from 0.85 to 0.15 in steps of 0.1 and respective delays (0, 0, 16, 64, 256, 512, 1024, 2048). E7 uses E0's means but reports only reward-one outcomes after delay 64.

The data archive gives actual pseudo-regret means (gap/count dot product) and sample standard errors over eight seeds. Figure S.1 summarizes the comparisons with NC-wide and BHO.

\subsection*{S.1 Good-event bounds}
For each endpoint we evaluate the two good-event bounds (R3) on predictable regret with \(q=0\), \(\alpha=1\) and the endpoint's last-begun membership size and exposure, without clipping and without the failure term. The sum bound lies below \(\Delta_{\max}T\) at 24 of the 112 covered endpoints and the maximum bound at none; the 16 E7 endpoints are not covered.

\end{multicols}
\par\medskip\noindent\begin{minipage}{\textwidth}
\centering
\includegraphics[width=\textwidth]{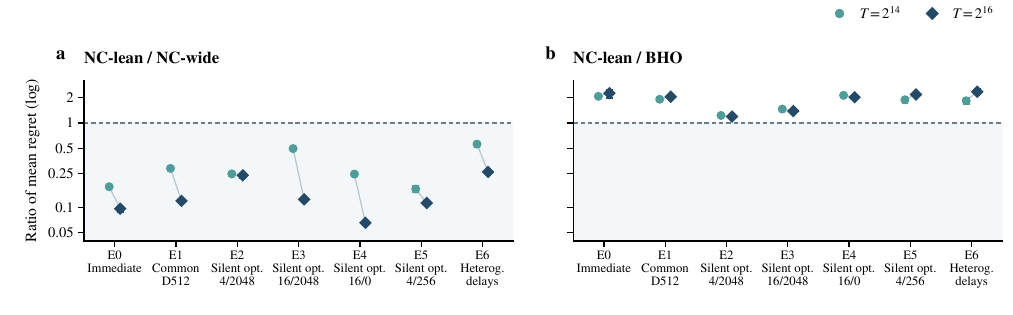}
\par\vspace{4pt}{\small \textbf{Figure S.1. Lower block exploration costs still exceed BHO's.} \textbf{a,} NC-lean versus NC-wide. \textbf{b,} NC-lean versus BHO. Points are ratios of arithmetic mean pseudo-regrets over eight paired seeds; bars are descriptive pointwise 95\% percentile intervals from 50,000 paired-seed bootstrap resamples. Both panels share a logarithmic ratio axis; shading marks ratios \(<1\), dashes equality, and segments join horizons. Panel \textbf{a} changes the logarithm and the tolerance together. Silent opt. means the optimum never reports; its labels give arm count/bad-arm delay. The data archive also reports E7, a success-only-reporting control outside the positive weighted-cutoff assumption.\par}
\end{minipage}\par\medskip
\begin{multicols}{2}
For E2, NC-lean, BRB and RATE each have the same cumulative regret at the two horizons because every seed has identical bad-arm pull counts at both endpoints; only the optimal-arm count increases, by 49,152. E3's NC-lean values also coincide at the two horizons.

Across E0--E6 and both horizons, the realized last-begun membership sizes \(M_i\) of bad arms in the NC-lean trajectories range from 1 to 16.

\pdfbookmark[0]{Appendix T. Fixed-Model Selection on Handwritten Digits}{r060.nav.23}
\section*{Appendix T. Fixed-Model Selection on Handwritten Digits}
\subsection*{T.1 Data and fixed classifiers}
We use UCI Optical Recognition of Handwritten Digits (Alpaydin and Kaynak, 1998): official training/test files of 3,823/1,797 records, each with 64 integer features and a digit label. A deterministic class-stratified split gives 3,054 fitting/769 calibration records; features are divided by 16. Table T.1's four classifiers are fitted once and retained independently of policy performance.

Before fitting, we required at least 5 calibration errors per model, an accuracy range of at least 0.03, and at least 2 distinct prediction and reward columns; all models qualify (observed range 146/769). Neighbor-distance ties follow original training-row order, and class ties choose the smallest label.

\end{multicols}
{\small
\begin{longtable}{>{\raggedright\arraybackslash}p{18.49pt}>{\raggedright\arraybackslash}p{221.77pt}>{\raggedright\arraybackslash}p{87.08pt}>{\raggedright\arraybackslash}p{66.24pt}>{\raggedright\arraybackslash}p{34.25pt}}
\multicolumn{5}{@{}p{\linewidth}@{}}{\textbf{Table T.1.} Fixed models. The last column is exact empirical accuracy on the 1,797-record test table, displayed to six decimals. Full constructor parameters and preprocessing are supplied in the data archive.}\\[3pt]
\toprule
\textbf{Arm} & \textbf{Classifier and main fixed parameters} & \textbf{Calibration errors/769} & \textbf{Test correct/1797} & \textbf{\(\mu_i\)} \\
\midrule
\endfirsthead
\multicolumn{5}{@{}l}{\footnotesize\itshape Continued from previous page}\\
\toprule
\textbf{Arm} & \textbf{Classifier and main fixed parameters} & \textbf{Calibration errors/769} & \textbf{Test correct/1797} & \textbf{\(\mu_i\)} \\
\midrule
\endhead
\midrule\multicolumn{5}{r@{}}{\footnotesize\itshape Continued on next page}\\
\endfoot
\bottomrule
\endlastfoot
M0 & Multinomial logistic regression: L2, C=1, lbfgs, tol=1e-4, \texttt{max\_iter=1000}, intercept & 19 & 1694 & 0.942682 \\
M1 & Five Euclidean neighbors, uniform weights; deterministic distance/vote ties & 8 & 1752 & 0.974958 \\
M2 & CART: gini, best splitter, depth 6, \texttt{min\_samples\_leaf=5}, seed 20261029 & 154 & 1348 & 0.750139 \\
M3 & Gaussian naive Bayes: training-set class priors, \texttt{var\_smoothing=1e-9} & 139 & 1412 & 0.785754 \\
\end{longtable}
}
\begin{multicols}{2}
\subsection*{T.2 Empirical reward law and feedback}
Let \(r_j(i)\) indicate correct classification of test record \(j\) by model \(i\). After committing \(A_s\), independently draw \(J_s\) uniformly with replacement from 1,797 records and set \(Y_s(i)=r_{J_s}(i)\). Conditional on this table, reward vectors are iid with allowed within-round dependence, exact means \(\mu_i=1797^{-1}\sum_jr_j(i)\), and unique optimum M1. \(K=4\) counts models, not replicates.

The learner sees its actions, clock and selected-model sorted (arm,reward) multiset, preserving zeros and multiplicity; reports omit originating rounds, item IDs, true labels and unselected predictions. It receives no means/channel parameters. Delay-zero feedback follows the current action; delay 128 from round \(s\) arrives at round \(s+128\)'s end.

Channels use decision-round delays. E0 reports immediately. E1 assigns delay 128 independently of reward with probability \(1-\mu_i\), otherwise 0. E2 delays errors by 128 and reports successes immediately. E1/E2 thus share per-arm delay marginals but differ in reward dependence. E3 uses E2 for suboptimal arms; each truly optimal-arm pull independently follows E2 with probability 0.05 and otherwise never reports. Only the evaluator knows the optimal set. This is a diagnostic oracle channel, not a deployable verification service: hiding optimal errors can strengthen completion, so E3 need not be harder. Every suboptimal arm admits a zero-allowance cutoff 128 in E1--E3, or 0 in E0.

\subsection*{T.3 Policies, comparison units, and uncertainty}
Policies are NC-lean, BHO, BRB (App. S), OPSE, returned-sample successive elimination (SE), returned-UCB and Uniform. BHO uses width 1 and, among eligible arms, selects the smallest pull count, then the largest historical index, then the smallest label. OPSE eliminates after completed scans of the active arms, using optimistic bounds for missing rewards. SE uses bounds from returned samples and, after completed scans, eliminates against the largest lower bound over all arms. Returned-UCB initially pulls each label once, then uses returned mean+\(\sqrt{2\log(\max\{2,t\})/n}\) for returned count \(n\), infinite at zero, with ascending-label ties.

For E0--E3, all 4×2×32×7=1,792 endpoints use \(T\in\{4096,16384\}\) and 32 seeds; App. T.5 adds E4 and E4b. Within seed, policies share iid record sequences and arm-specific channel uniforms; action coins have private streams. Seed-specific permutations hide model labels and horizons share prefixes. Seeds are Monte Carlo units; correlated horizons stay in the family.

For \(\Psi_T=\sum_i(\mu^*-\mu_i)N_i(T)\), Table T.2 normalizes the mean and seed SE by the exact uniform reference \(U_T=(T/4)\sum_i(\mu^*-\mu_i)\). Primary contrasts are NC-lean minus BHO/BRB/OPSE/SE across four channels×two horizons (32 coordinates); adding returned-UCB/Uniform gives the complete 48-coordinate extended family. Pointwise 95\% t intervals use df=31; Bonferroni-t uses \(t_{1-0.05/(2m),31}\), \(m=32,48\). These approximate normal-model intervals describe conditional Monte Carlo variation. The 28 descriptive channel contrasts compare E1 minus E2 and E2 minus E3 separately for each of seven policies and two horizons (2 × 7 × 2); each contrast retains all 32 paired seed differences.

\end{multicols}
{\small
\begin{longtable}{>{\raggedright\arraybackslash}p{43.44pt}>{\raggedright\arraybackslash}p{48.22pt}>{\raggedright\arraybackslash}p{48.22pt}>{\raggedright\arraybackslash}p{48.47pt}>{\raggedright\arraybackslash}p{48.22pt}>{\raggedright\arraybackslash}p{48.22pt}>{\raggedright\arraybackslash}p{58.81pt}>{\raggedright\arraybackslash}p{48.22pt}}
\multicolumn{8}{@{}p{\linewidth}@{}}{\textbf{Table T.2.} Normalized cumulative pseudo-regret, mean±seed SE across 32 paired seeds. Both quantities are divided by \(U_T\); smaller values are better. Values are rounded, with small nonzero SEs shown in scientific notation.}\\[3pt]
\toprule
\textbf{Channel/T} & \textbf{NC-lean} & \textbf{BHO} & \textbf{BRB} & \textbf{OPSE} & \textbf{SE} & \textbf{returned-UCB} & \textbf{Uniform} \\
\midrule
\endfirsthead
\multicolumn{8}{@{}l}{\footnotesize\itshape Continued from previous page}\\
\toprule
\textbf{Channel/T} & \textbf{NC-lean} & \textbf{BHO} & \textbf{BRB} & \textbf{OPSE} & \textbf{SE} & \textbf{returned-UCB} & \textbf{Uniform} \\
\midrule
\endhead
\midrule\multicolumn{8}{r@{}}{\footnotesize\itshape Continued on next page}\\
\endfoot
\bottomrule
\endlastfoot
E0/4,096 & 0.998±0.003 & 0.951±0.006 & 0.998±5e-05 & 1.000±0 & 1.000±0 & 0.269±0.004 & 0.998±0.003 \\
E0/16,384 & 0.754±0.008 & 0.391±0.003 & 1.000±1e-05 & 0.525±0.005 & 0.525±0.005 & 0.137±0.002 & 0.998±0.001 \\
E1/4,096 & 0.998±0.003 & 0.953±0.006 & 0.998±5e-05 & 1.000±0 & 1.000±0 & 0.275±0.004 & 0.998±0.003 \\
E1/16,384 & 0.754±0.008 & 0.395±0.003 & 1.000±1e-05 & 0.530±0.005 & 0.525±0.005 & 0.138±0.002 & 0.998±0.001 \\
E2/4,096 & 0.998±0.003 & 0.969±0.005 & 0.998±5e-05 & 1.000±0 & 1.000±0 & 0.342±0.007 & 0.998±0.003 \\
E2/16,384 & 0.774±0.012 & 0.407±0.003 & 1.000±1e-05 & 0.543±0.005 & 0.538±0.005 & 0.154±0.003 & 0.998±0.001 \\
E3/4,096 & 0.998±0.003 & 0.969±0.005 & 0.998±5e-05 & 1.000±0 & 1.000±0 & 0.079±0.002 & 0.998±0.003 \\
E3/16,384 & 0.774±0.012 & 0.407±0.003 & 1.000±1e-05 & 0.704±0.010 & 0.703±0.009 & 0.060±0.002 & 0.998±0.001 \\
\end{longtable}
}
\begin{multicols}{2}
\subsection*{T.4 Results}
NC-lean has higher mean regret than BHO in all eight cells, with positive primary Bonferroni-t intervals; at T=16,384 its normalized regret is 0.754--0.774 versus 0.391--0.407. Its means are below BRB's in all eight, with negative extended-family intervals in the four longer cells. Against OPSE/SE, short-horizon differences are slightly negative with zero-containing intervals; all longer-horizon differences are positive. In E0--E3, returned-UCB performs better despite lacking a reward-dependent-delay guarantee. In two-armed simulations with outcomes missing at random, Chen et al.~(2022) find that exploration-oriented algorithms that ignore missing outcomes keep assigning samples to the arm with more missing responses; in E3 that arm is the optimum, and returned-UCB's width, computed from returned counts, behaves the same way. All cells/zero contrasts remain, including 17 all-zero vectors among 28 specified channel contrasts.

\subsection*{T.5 Unconfirmed correct predictions (E4 and E4b)}
E4 keeps E2 for M0, M2 and M3. Each pull of the optimal model M1 reports an error at once; a correct prediction is confirmed at once with probability \(p\in\{0,0.05\}\) and otherwise never. E4b applies the same rule to M0 and M1. Confirmation coins form a new stream per model, generated per calendar round and shared across \(p\) and channels; all other inputs, the seven policies and the 32 paired seeds are those of T.3, now at \(T\in\{4096,16384,65536\}\), and the shorter runs are prefixes of the longer ones. This adds 2,688 endpoints, reported descriptively in Table T.3. These channels use evaluator-known model identities, as in E3.

With growing returned-sample counts, the returned averages of M1, and of M0 in E4b, converge to 0 at \(p=0\) and at \(p=0.05\) to \(146/221=0.660633\) (M1) and \(847/1877=0.451252\) (M0), below M0's 0.942682 in E4 and below M2 and M3 in E4b. Under returned-UCB at \(T=65536\), the pooled returned accuracies are 0.665 (M1, E4) and 0.662 and 0.450 (M1, M0, E4b) at \(p=0.05\). A policy that settles on M0 or M3 has per-round regret \(58/1797\) or \(340/1797\), that is 0.289 or 1.696 times the uniform per-round regret \(401/3594\).

By O.5, BHO's E4 trajectory equals that of an auxiliary channel in which M1 reports every outcome at once and the other models follow E2; in E4b the auxiliary channel lets M0 and M1 report at once. All 384 paired BHO endpoint comparisons match their auxiliary channels action by action, covering 11,010,048 action positions when shared prefixes are counted at each horizon. In these comparisons the BHO trajectory is independent of \(p\). The E2 models keep zero-allowance cutoff 128. In E4b, M0's failures report at once, so \(w_0(c)=0\) although its ordinary cutoff is infinite; Theorem 2 applies where Theorem 1 does not.

At \(T=65536\), BHO's normalized regret is 0.189 in all four cells, against 0.320 and 0.308 for returned-UCB in E4 (\(p=0\), 0.05) and 1.684 and 1.331 in E4b. Returned-UCB is lower in the four E4 cells with \(T\le16384\) and in E4b at \(p=0.05\), \(T=4096\); BHO is lower in the other seven cells. These are finite-horizon comparisons for one frozen classifier pool and simulated reporting rules.

\end{multicols}
{\small
\begin{longtable}{>{\raggedright\arraybackslash}p{56.42pt}>{\raggedright\arraybackslash}p{46.42pt}>{\raggedright\arraybackslash}p{46.42pt}>{\raggedright\arraybackslash}p{46.67pt}>{\raggedright\arraybackslash}p{46.42pt}>{\raggedright\arraybackslash}p{46.42pt}>{\raggedright\arraybackslash}p{56.61pt}>{\raggedright\arraybackslash}p{46.42pt}}
\multicolumn{8}{@{}p{\linewidth}@{}}{\textbf{Table T.3.} Channels E4 and E4b: normalized cumulative pseudo-regret, mean±seed SE across 32 paired seeds, rounded as in Table T.2. Rows give channel, \(p\) (.05 for 0.05) and \(T\). Descriptive; no simultaneous family is formed.}\\[3pt]
\toprule
\textbf{Channel/\(p\)/T} & \textbf{NC-lean} & \textbf{BHO} & \textbf{BRB} & \textbf{OPSE} & \textbf{SE} & \textbf{returned-UCB} & \textbf{Uniform} \\
\midrule
\endfirsthead
\multicolumn{8}{@{}l}{\footnotesize\itshape Continued from previous page}\\
\toprule
\textbf{Channel/\(p\)/T} & \textbf{NC-lean} & \textbf{BHO} & \textbf{BRB} & \textbf{OPSE} & \textbf{SE} & \textbf{returned-UCB} & \textbf{Uniform} \\
\midrule
\endhead
\midrule\multicolumn{8}{r@{}}{\footnotesize\itshape Continued on next page}\\
\endfoot
\bottomrule
\endlastfoot
E4/0/4,096 & 0.998±0.003 & 0.969±0.005 & 0.998±5e-05 & 1.000±0 & 1.019±0.006 & 0.521±0.008 & 0.998±0.003 \\
E4/0/16,384 & 0.774±0.012 & 0.407±0.003 & 1.000±1e-05 & 0.704±0.010 & 0.783±0.009 & 0.385±0.003 & 0.998±0.001 \\
E4/0/65,536 & 0.316±0.005 & 0.189±0.001 & 1.000±0 & 0.302±0.002 & 0.429±0.002 & 0.320±0.001 & 1.000±0.001 \\
E4/.05/4,096 & 0.998±0.003 & 0.969±0.005 & 0.998±5e-05 & 1.000±0 & 1.000±0 & 0.414±0.008 & 0.998±0.003 \\
E4/.05/16,384 & 0.774±0.012 & 0.407±0.003 & 1.000±1e-05 & 0.704±0.010 & 0.714±0.009 & 0.347±0.003 & 0.998±0.001 \\
E4/.05/65,536 & 0.316±0.005 & 0.189±0.001 & 1.000±0 & 0.302±0.002 & 0.404±0.002 & 0.308±0.001 & 1.000±0.001 \\
E4b/0/4,096 & 0.998±0.003 & 0.969±0.005 & 0.998±5e-05 & 1.000±0 & 1.078±0.005 & 1.301±0.015 & 0.998±0.003 \\
E4b/0/16,384 & 0.774±0.012 & 0.407±0.003 & 1.000±1e-05 & 1.000±0 & 1.578±0.006 & 1.604±0.005 & 0.998±0.001 \\
E4b/0/65,536 & 0.316±0.005 & 0.189±0.001 & 1.000±0 & 1.000±0 & 1.779±0.002 & 1.684±0.002 & 1.000±0.001 \\
E4b/.05/4,096 & 0.998±0.003 & 0.969±0.005 & 0.998±5e-05 & 1.000±0 & 1.000±0 & 0.542±0.020 & 0.998±0.003 \\
E4b/.05/16,384 & 0.774±0.012 & 0.407±0.003 & 1.000±1e-05 & 1.000±0 & 1.075±0.007 & 0.871±0.025 & 0.998±0.001 \\
E4b/.05/65,536 & 0.316±0.005 & 0.189±0.001 & 1.000±0 & 1.000±0 & 1.190±0.002 & 1.331±0.012 & 1.000±0.001 \\
\end{longtable}
}
\begin{multicols}{2}
\pdfbookmark[0]{Appendix U. Multiscale Delay Information Limits}{r060.nav.24}
\section*{Appendix U. Multiscale Delay Information Limits}
\subsection*{U.1 Full statement of Theorem 7}
Fix \(c=1/4\) and \(\alpha=3/4\). For either the ordinary cutoffs \(h_i=d_i(1-c\Delta_i)\) or the weighted cutoffs \(h_i=w_i(c)\), write
\[
\begin{gathered}
G=\max_{i\in\mathcal I}\Delta_i h_i,\qquad\\
A_T=\alpha^{-2}\log(eKT)\sum_{i\in\mathcal I}\Delta_i^{-1}.
\end{gathered}
\]
The next result concerns every learner in the observation model of Section 2.1.

\textbf{Theorem 7 (multiscale delay lower bound, full statement).} For every pair of integers \(m,H\ge1\), set \(K=2^m\) and \(T=HK\). There is a finite family \(\mathcal P_{m,H}\) of deterministic iid instances, each with one optimal arm, such that every possibly randomized learner \(\mathsf A\) satisfies
\begin{equation}
\begin{gathered}
\max_{P\in\mathcal P_{m,H}}R_T(\mathsf A,P)\\
\ge\frac{Hm}{4}=\frac{G\log K}{4\log2}.
\end{gathered}
\tag{U1}
\end{equation}
Every member has finite suboptimal cutoffs under both definitions, with \(G=H\) and \(\sum_{i\in\mathcal I}\Delta_i^{-1}=mK\). The lower bound still holds if the learner knows the full unlabeled family and the optimal mean. The optimal arm may never report; alternatively, all delays can be finite.

\subsection*{U.2 A fixed family of instances}
There is one optimal type with deterministic reward one and delay infinity. For each \(j=1,\ldots,m\), there are \(n_j=2^{m-j}\) arms of type \(j\), with
\begin{equation}
\begin{gathered}
Y_j=1-2^{-j},\qquad\\
\Delta_j=2^{-j},\qquad d_j=H2^j.
\end{gathered}
\tag{U2}
\end{equation}
Assign these types to the \(K\) arm labels in every possible way to obtain \(\mathcal P_{m,H}\). The count is \(1+\sum_{j=1}^m2^{m-j}=2^m=K\). Each fixed assignment is a constant joint reward-delay vector and hence an admissible iid law. The assignment stays fixed throughout the run.

For a type-\(j\) arm, the ordinary quantile at \(1-c\Delta_j\) is \(d_j\). Its weighted tail is
\[
b_j(h)=
\begin{cases}
\Delta_j,&h<d_j,\\
0,&h\ge d_j.
\end{cases}
\]
Since \(c<1\), its weighted cutoff also equals \(d_j\). Consequently both cutoff definitions give
\begin{equation}
\begin{gathered}
G=H,\qquad\\
\sum_{i\in\mathcal I}\Delta_i^{-1}\\
=\sum_{j=1}^m2^{m-j}2^j=mK.
\end{gathered}
\tag{U3}
\end{equation}
\subsection*{U.3 Earlier free revelations}
Draw one uniform assignment \(\Pi\) of the type multiset to the labels. This is a proof device: the instances being compared are the fixed members of \(\mathcal P_{m,H}\), not a law that changes types during the run.

Give the learner the unlabeled family and an additional oracle. At the end of absolute round \(d_j\), the oracle reveals all labels of type \(j\) and their complete reward-delay laws. These revelations are free and independent of the learner's actions. Any original learner can ignore the extra information, so it suffices to prove the lower bound for every learner in this larger class.

The ordinary report stream adds no information to the oracle history. A type-\(j\) arm selected in round \(s\ge1\) first reports at round \(s+d_j\ge1+d_j\), strictly after its type has been revealed. Once its law is known, its reports are determined by its previous selection times. Unrevealed arms have not reported at all, and the optimum never reports. Thus actions and earlier oracle revelations determine all ordinary reports, including their absence. The same argument would hold if reports also contained origin timestamps.

Let \(d_0=0\). During the decisions in phase
\begin{equation}
\mathcal T_j=\{d_{j-1}+1,\ldots,d_j\},
\tag{U4}
\end{equation}
the learner knows precisely the labels in types \(1,\ldots,j-1\). The revelation at the end of round \(d_j\) is unavailable to that round's decision. The remaining label set \(U_j\) has size
\begin{equation}
|U_j|=1+\sum_{\ell=j}^m2^{m-\ell}=2^{m-j+1}.
\tag{U5}
\end{equation}
Conditional on the learner's entire visible history, the remaining multiset is uniformly assigned to \(U_j\). To establish this for an arbitrary adaptive policy, first fix its private random seed, which is independent of \(\Pi\). Two assignments compatible with the same earlier revelations produce identical ordinary reports up to any phase-\(j\) decision. Induction over those decisions therefore gives identical past actions for the fixed seed. All compatible assignments have the same likelihood of the observed history, so their uniform prior remains uniform. Averaging over the private seed proves the assertion for randomized policies as well.

It follows that each unrevealed label has conditional expected gap
\begin{equation}
\begin{gathered}
\overline\Delta_j\\
=\frac{\sum_{\ell=j}^m2^{m-\ell}2^{-\ell}}{|U_j|}\\
\ge\frac{2^{m-j}2^{-j}}{2^{m-j+1}}\\
=2^{-j-1}.
\end{gathered}
\tag{U6}
\end{equation}
Every revealed arm has a type \(\ell<j\), hence a gap at least \(2^{-j+1}\). Therefore every action, whether it selects a revealed or an unrevealed label, has conditional expected gap at least \(2^{-j-1}\). This conclusion does not require the learner to sample uniformly.

\subsection*{U.4 Regret across scales}
For \(j\ge2\), the phase length is \(d_j-d_{j-1}=H2^{j-1}\), so (U6) gives expected phase regret at least \(H/4\). The first phase has length \(2H\) and contributes at least \(H/2\). The phases partition all \(T=d_m\) decisions. Taking conditional expectations and summing yields
\[
\mathbb E_{\Pi,\mathsf A}\Psi_T\ge mH/4.
\]
The expectation is an average over finitely many fixed assignments. For every learner, at least one assignment therefore has \(R_T\ge mH/4\). Together with (U3), this proves (U1). The phase bounds in fact sum to \((m+1)H/4\). Both cutoffs in U.2 equal \(d_j\) for every \(c\in[0,1)\), so this stronger delay lower bound holds throughout that range.

For the finite-delay version, replace the optimal type's delay by the deterministic value \(T+1\). None of its selected outcomes reports by round \(T\); every visible history and all regret calculations above are unchanged. The suboptimal cutoffs and \(G,A_T\) are also unchanged. Thus the same conclusion holds with deterministic finite delays on every arm.

\subsection*{U.5 Separating delay cost from the statistical term}
\textbf{Corollary U.1 (a uniform gap-weighted maximum needs a logarithmic factor).} There are no finite universal nonnegative constants \(C_s,C_d,C_0\) and learners satisfying
\begin{equation}
R_T\le\min\{T,C_sA_T+C_dG+C_0\}
\tag{U7}
\end{equation}
on every admissible instance. More generally, replacing \(G\) in (U7) by \(Gf(K)\) with nonnegative \(f(K)=o(\log K)\) is insufficient.

\textbf{Proof.} Take \(K=2^m\), \(H=K^2\) and \(T=K^3\). Then
\[
A_T=\frac{16}{9}mK(1+4m\log2),\qquad G=K^2,
\]
and hence
\begin{equation}
\begin{gathered}
\frac{A_T}{G}\\
=\frac{16}{9}\frac{m(1+4m\log2)}{2^m}\longrightarrow0,\\
\qquad\\
\frac{G\log K}{T}=\frac{\log K}{K}\longrightarrow0.
\end{gathered}
\tag{U8}
\end{equation}
The adverse labeling in Theorem 7 would force
\[
\begin{gathered}
\frac m4\\
\le C_s\frac{16}{9}\\
\frac{m(1+4m\log2)}{2^m}\\
+C_d+\frac{C_0}{4^m},
\end{gathered}
\]
which is impossible as \(m\) grows. For \(Gf(K)\), divide the corresponding inequality by \(m\); \(f(2^m)/m\to0\), and every term on the right tends to zero. The left side remains \(1/4\).

This is a growing-\(K\) obstruction. It does not exclude a constant depending on a fixed arm count, nor establish a matching upper bound in the original observation model.

\pdfbookmark[0]{Appendix V. Pending-Triggered Balancing}{r060.nav.25}
\section*{Appendix V. Pending-Triggered Balancing}
PTB uses the model and completed-round convention of Section 2. It keeps BHO's optimistic completion and historical indices, but invokes count priority only when the leader has substantial pending feedback.

\subsection*{V.1 Rule and guarantees}
Let \(\delta=(T+1)^{-2}\), \(\ell_T=\log(4KT/\delta)\), and \(L=\log(eKT)\). At completed time \(t\), write \(m_i(t),n_i(t),G_i(t)\) for arm \(i\)'s pull count, report count, and reported reward sum. Set \(P_i(t)=m_i(t)-n_i(t)\). At positive count, define
\begin{equation}
\begin{gathered}
\beta_i(t)=\sqrt{\frac{\ell_T}{2m_i(t)}},\qquad\\
u_i^+(t)=\frac{G_i(t)+P_i(t)}{m_i(t)},\qquad\\
\rho_i(t)=u_i^+(t)+\beta_i(t).
\end{gathered}
\tag{V1}
\end{equation}
Initialize \(H_i(0)=1\), and after processing all reports at each round's end update
\[
H_i(t)=\min\{H_i(t-1),1,\rho_i(t)\}.
\]
At zero count, retain \(H_i=1\), put \(\beta_i=\infty\), and do not evaluate \(\rho_i\). Let
\[
\begin{gathered}
M(t)=\max_iH_i(t),\qquad\\
\mathcal C_1(t)=\{i:H_i(t)+\beta_i(t)\ge M(t)\}.
\end{gathered}
\]
Choose the leader \(g\) by lexicographically minimizing \((m_i,-H_i,i)\) among arms with \(H_i=M\).

\begin{algorithm}[t]
\caption{Algorithm}
\textbf{Algorithm V.1 (pending-triggered balance, PTB).} For \(t=0,\ldots,T-1\):
\begin{enumerate}

\def\labelenumi{\arabic{enumi}.}

\setlength{\itemsep}{8pt}\setlength{\parskip}{0pt}

\item

  If \(m_g=0\), choose \(g\).

\item

  If \(m_g>0\) and \(P_g/m_g\le\beta_g\), choose \(g\).

\item

  Otherwise, choose the lexicographic minimum of \((m_i,-H_i,i)\) in \(\mathcal C_1(t)\).

\item

  Pull the chosen arm, increment its count, process every report arriving at this round's end, and update every positive-count historical index, including those of idle arms.

\end{enumerate}

\end{algorithm}
The test is repeated each round. Every chosen arm belongs to \(\mathcal C_1(t)\). PTB uses only \(K,T\) and observable aggregates, stores \(O(K)\) scalars, and takes \(O(K+r_t)\) work when \(r_t\) reports arrive.

Fix \(c\in[0,1)\) and set \(\alpha=1-c\). For each bad arm let
\[
\begin{gathered}
h_i=w_i(c)=\min\{h\in\mathbb N_0:\\
\mathbb E[(1-Y_i)\mathbf1\{D_i>h\}]\le c\Delta_i\}.
\end{gathered}
\]
Write \(W=\max_{i\in\mathcal I}h_i\), \(Z=\sum_{i\in\mathcal I}\Delta_i^{-1}\), and \(A_T=LZ/\alpha^2\), with empty sums and maxima zero.

\textbf{Theorem V.1 (general protection).} If every bad-arm cutoff is finite, PTB satisfies
\begin{equation}
R_T\le\min\left\{T,\, 18473A_T+\frac{9264\Phi_T(W)}{\alpha}\right\}.
\tag{V2}
\end{equation}
The same numerical threshold bounds \(\Psi_T\) with probability at least \(1-(T+1)^{-2}\). Multiple optima and arbitrary optimal-arm reporting laws are allowed. If \(\mathcal I=\varnothing\), regret is zero; if a required cutoff is infinite, only the trivial \(T\) bound is asserted.

\textbf{Theorem V.2 (bounded-delay deterministic rewards).} Fix finite \(K\), deterministic rewards, a unique optimal arm, and fixed positive suboptimal gaps. Suppose every arm's delay lies almost surely in \(\{0,\ldots,d_0\}\), for a fixed finite integer \(d_0\) unknown to PTB. Then, for each \(i\in\mathcal I\),
\begin{equation}
\begin{gathered}
\lim_{T\to\infty}\frac{\mathbb E N_i(T)}{\ell_T}\\
=\frac1{2\Delta_i^2},\qquad\\
\lim_{T\to\infty}\frac{R_T}{\ell_T}\\
=\frac12\sum_{i\in\mathcal I}\frac1{\Delta_i}.
\end{gathered}
\tag{V3}
\end{equation}
The instance is fixed while the known horizon varies.

\subsection*{V.2 Which selections require balance}
Use the per-arm iid-stack representation of Appendix A.1. On an event \(E\), require for every arm and every \(1\le m\le T\),
\[
\begin{gathered}
\left|\frac1m\sum_{k\le m}Y_{i,k}-\mu_i\right|\\
\le\sqrt{\frac{\ell_T}{2m}},
\end{gathered}
\]
and, for every bad arm,
\begin{equation}
\begin{gathered}
\sum_{k\le m}(1-Y_{i,k})\mathbf1\{D_{i,k}>h_i\}\\
\le mb_i(h_i)+\sqrt{m\ell_T/2}.
\end{gathered}
\tag{V4}
\end{equation}
Hoeffding's inequality and the finite union bound give \(\Pr(E^c)\le3KT e^{-\ell_T}\le\delta\). The bounds hold at adaptive prefix counts and require no independence between arms.

Work on \(E\). With \(\overline Y_i(t)\) the mean of all latent rewards pulled from arm \(i\), optimistic completion satisfies
\begin{equation}
\begin{gathered}
u_i^+(t)-\overline Y_i(t)\\
=\frac1{m_i(t)}\\
\sum_{\text{pending pulls of }i}(1-Y).
\end{gathered}
\tag{V5}
\end{equation}
Thus \(\rho_i\ge\mu_i\), \(H_i\ge\mu_i\), and \(M\ge\mu^*\). Also
\[
\rho_i\le\mu_i+2\beta_i+\frac{P_i}{m_i}.
\]
A positive-count bad arm selected by the leader branch passed the pending test, so
\begin{equation}
\begin{gathered}
\mu^*\le H_i\le\rho_i\le\mu_i+3\beta_i,\\
\qquad \Delta_i\le3\beta_i.
\end{gathered}
\tag{V6}
\end{equation}
Put \(\gamma_i=\alpha\Delta_i\) and \(Q_i=4608\ell_T/\gamma_i^2\). Call a bad selection cold if its pre-action count is below \(Q_i\), and mature otherwise. At maturity, \(\beta_i\le\gamma_i/96\), contradicting (V6). Zero-count choices are cold. Consequently every mature bad selection uses the full balanced branch.

For \(B_i(t,h)=m_i(t)-m_i((t-h)_+)\), the completion argument of Appendix A.2 gives, at every positive-count time,
\begin{equation}
\begin{gathered}
\rho_i(t)\le\\
\mu^*-\gamma_i+\frac{B_i(t,h_i)}{m_i(t)}+3\beta_i(t).
\end{gathered}
\tag{V7}
\end{equation}
This holds also when arm \(i\) is idle: old pending pulls have delay exceeding \(h_i\), and at most \(B_i(t,h_i)\) pulls are newer. Unit-width eligibility and maturity therefore imply
\begin{equation}
\frac{B_i(t,h_i)}{m_i(t)}\ge\frac{23\gamma_i}{24},
\qquad h_i\gamma_i\ge4416\ell_T.
\tag{V8}
\end{equation}
Each arm has at most \(Q_i+1\) cold selections, whose total regret is at most \((4608\ell_T/\alpha^2+1)Z\). If \(W\max_i\gamma_i<4416\ell_T\), all bad selections are cold. This includes \(W=0\); \(T=1\) is cold as well.

\subsection*{V.3 Shared calendar intervals under selective balance}
Apply Lemma A.1 with \(\ell=\ell_T\) and \(D=W\). The event in V.2 is the fixed-cutoff event of A.1, and (V7) is its completion bound (A6). Every PTB choice belongs to \(\mathcal C_1(t)\). By (V6), on \(E\) every mature suboptimal choice uses the full balanced branch and therefore has minimum count in that set. These are exactly the allocation properties of Lemma A.1.

Lemma A.1 therefore gives (V2) in expectation and with the stated probability, including clipping and the cold-only cases \(W=0\) and \(T=1\). An empty bad set has zero regret. This proves Theorem V.1 for each fixed \(c\). Because PTB is independent of \(c\), Appendix A.7 gives both bounds with the infimum of the right-hand side of (V2) over \(c\) with finite \(W\).

\subsection*{V.4 Bounded delays and the exact exploration coefficient}
On every supported delay path, at completed time \(t\) pending origins lie in \((t-d_0,t]\), so
\begin{equation}
\begin{gathered}
0\le P_i(t)\le\min\{m_i(t),d_0\},\qquad\\
P_i(t)^2\le d_0m_i(t).
\end{gathered}
\tag{V9}
\end{equation}
When \(\ell_T\ge2d_0\), every positive-count arm passes \(P_i/m_i\le\beta_i\). PTB then follows historical greedy throughout the run, including its zero-count choices, without knowing \(d_0\).

For deterministic rewards, \(G_i=\mu_i n_i\), so
\[
\rho_i=\mu_i+(1-\mu_i)P_i/m_i+\beta_i.
\]
Put \(a_i=(1-\mu_i)d_0\). Monotone counts and the historical update imply
\begin{equation}
\begin{gathered}
\min\{1,\mu_i+\beta_i(t)\}\le H_i(t)
\le\mu_i+\beta_i(t)+\frac{a_i}{m_i(t)}\\
\quad(m_i(t)>0).
\end{gathered}
\tag{V10}
\end{equation}
Every earlier score is at least \(\mu_i+\beta_i(t)\); clipping and initialization contribute the value one. The current score gives the upper bound. In particular \(H_*\ge\mu^*\).

Assume \(\ell_T\ge2d_0\). A bad arm selected at positive pre-count \(m\) is then a historical maximizer and satisfies
\[
\Delta_i m\le\sqrt{\ell_Tm/2}+a_i.
\]
Define
\begin{equation}
q_i=\left(
\frac{\sqrt{\ell_T/2}+
\sqrt{\ell_T/2+4\Delta_i a_i}}{2\Delta_i}
\right)^2,\quad B_i=\lfloor q_i\rfloor+1.
\tag{V11}
\end{equation}
Every selected pre-count, including zero, is at most \(q_i\). Thus
\begin{equation}
N_i(T)\le B_i\le
\frac{\ell_T}{2\Delta_i^2}+\frac{2a_i}{\Delta_i}+1.
\tag{V12}
\end{equation}
To obtain the second inequality, expand \(q_i\) with \(x=\sqrt{\ell_T/2}\) and use \(\sqrt{x^2+4\Delta_i a_i}\le x+2\Delta_i a_i/x\). Under \(\ell_T\ge2d_0\), this bounds every supported delay path and hence the expected count.

For the lower bound, let \(B=\sum_{i\in\mathcal I}B_i\) and \(s=T-B-1\). Suppose \(s>\max_{i\in\mathcal I}B_i\). There is an optimal action, and just before its last selection,
\[
\begin{gathered}
m_*=N_*(T)-1\ge s,\qquad m_i\le B_i<m_*\\
\quad(i\in\mathcal I).
\end{gathered}
\]
Uniqueness of the optimum is used here. If any bad arm were a historical maximizer, the greedy count tie rule would choose it instead. Hence \(H_i<M=H_*\) for every bad arm. Each has positive count and \(H_i<1\), since otherwise its index would be one and it would be a maximizer. Equation (V10) is therefore unclipped on its lower side, and
\[
\begin{gathered}
\mu_i+\beta_i\le H_i<H_*\\
\le\mu^*+\sqrt{\ell_T/(2s)}+a_*/s.
\end{gathered}
\]
With \(\epsilon_T=\sqrt{\ell_T/(2s)}+a_*/s\), this gives
\begin{equation}
\begin{aligned}
\frac{\ell_T}{2(\Delta_i+\epsilon_T)^2}&<N_i(T)\\
&\le\frac{\ell_T}{2\Delta_i^2}+\frac{2a_i}{\Delta_i}+1.
\end{aligned}
\tag{V13}
\end{equation}
For a fixed instance, \(B=O(\ell_T)\), \(s\sim T\), \(\epsilon_T\to0\), and both \(\ell_T\ge2d_0\) and \(s>\max_i B_i\) eventually hold. The deterministic bounds (V13) squeeze every delay path and its expectation to the count limit in (V3). Summing over the fixed finite bad-arm set proves the regret limit.

At \(d_0=0\), PTB equals historical greedy at every horizon. Appendix N's coefficient \((1+\theta)^2/(2\Delta_i^2)\) then makes its asymptotic regret coefficient one quarter of unit-width BHO's. Theorem V.2 supplies the bounded-positive-delay extension through (V9)--(V13).

\end{multicols}
\end{document}